\documentclass{article} 
\usepackage{metafiles/iclr2026_conference}
\usepackage{times}
\usepackage{hyperref}
\usepackage{url}

\usepackage[utf8]{inputenc}
\usepackage[T1]{fontenc}
\usepackage{hyperref}
\usepackage{url}
\usepackage{booktabs}
\usepackage{amsfonts}
\usepackage{nicefrac}
\usepackage{microtype}
\usepackage{xcolor}
\usepackage{enumitem}

\usepackage{graphicx}
\usepackage{amsmath} 
\usepackage{multirow}
\usepackage[normalem]{ulem}
\usepackage{xspace}
\usepackage{wrapfig}

\def\network{RTS\xspace}
\newcommand{\Lagr}{\mathcal{L}}

\title{Radiant Triangle Soup with Soft Connectivity Forces for 3D Reconstruction and Novel View Synthesis}

\author{
  Nathaniel Burgdorfer \\
  Department of Computer Science\\
  Stevens Institute of Technology\\
  Hoboken, NJ 07030 \\
  \texttt{nburgdor@stevens.edu} \\
  \And
  Philippos Mordohai \\
  Department of Computer Science\\
  Stevens Institute of Technology\\
  Hoboken, NJ 07030 \\
  \texttt{pmordoha@stevens.edu} \\
}

\iclrfinalcopy 

\begin{document}

\maketitle

\begin{figure}[ht]
    \vspace{-8mm}
    \centering
    \includegraphics[width=0.85\linewidth]{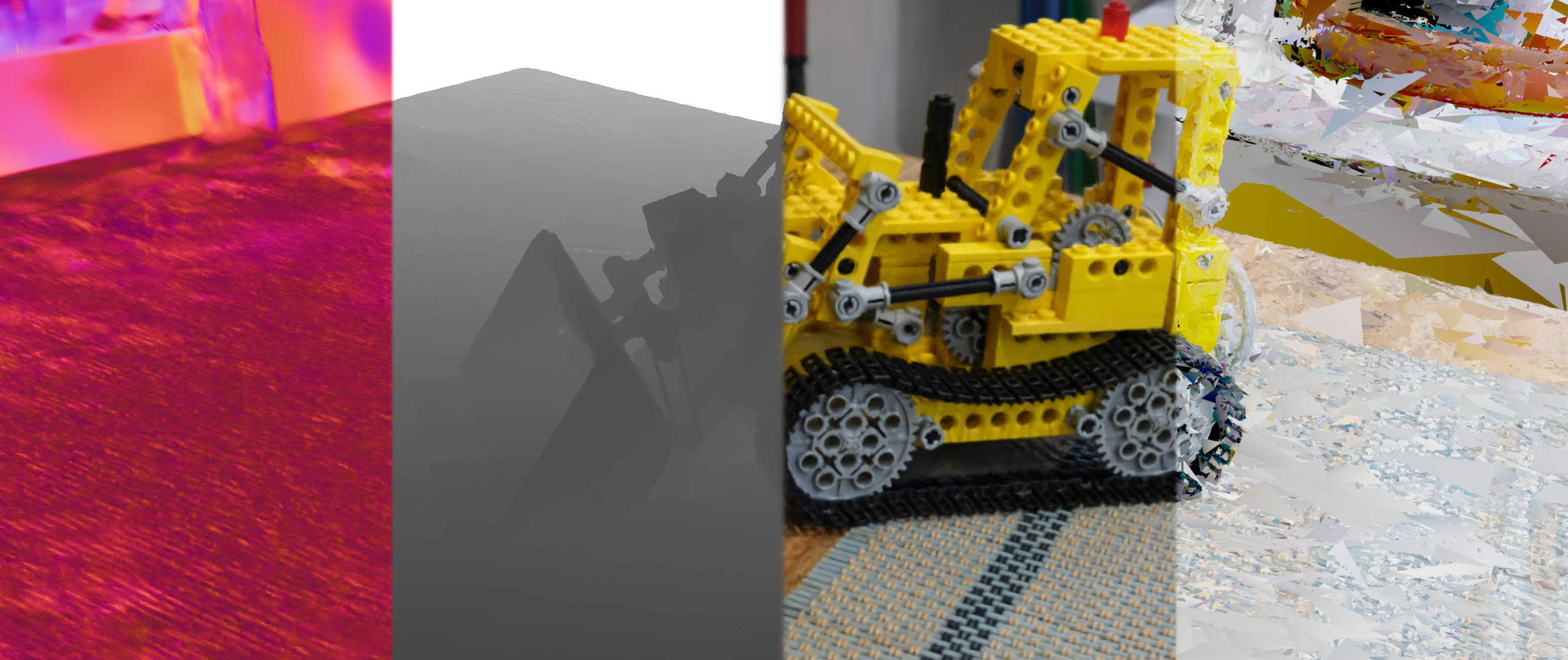}
    \caption{\small
    Optimizing Radiant Triangle Soup (\network{}) produces high quality 3D models from captured images. Above, from left to right, are the rendered normal map, rendered depth map, rendered image, and the direct rasterization of the triangle primitives, as if they were opaque.
    }
    \label{fig:tagline}
\end{figure}

\begin{abstract}
    We introduce an inference-time scene optimization algorithm utilizing triangle soup, a collection of disconnected translucent triangle primitives, as the representation for the geometry and appearance of a scene. Unlike full-rank Gaussian kernels, triangles are a natural, locally-flat proxy for surfaces that can be connected to achieve highly complex geometry. When coupled with per-vertex Spherical Harmonics (SH), triangles provide a rich visual representation without incurring an expensive increase in primitives. We leverage our new representation to incorporate optimization objectives and enforce spatial regularization directly on the underlying primitives. The main differentiator of our approach is the definition and enforcement of soft connectivity forces between triangles during optimization, encouraging explicit, but soft, surface continuity in 3D. Experiments on representative 3D reconstruction and novel view synthesis datasets show improvements in geometric accuracy compared to current state-of-the-art algorithms without sacrificing visual fidelity.
\end{abstract}

\section{Introduction}
\label{sec:intro}

Gaussian Splatting (GS) methods are effective at Novel View Synthesis (NVS). However 3D Gaussian Splatting (3DGS) \citep{kerbl_2023_3d} fails to accurately model the geometry of scene surfaces. 3D Gaussians are by design smooth, unbounded volumetric primitives, which are inherently ill-suited for representing flat surfaces and sharp boundaries. Optimization for novel view synthesis involves alpha-blending several overlapping Gaussians, none of which have to be located on the underlying surfaces. As a result, optimizing a representation of 3D Gaussian kernels for image synthesis commonly causes floaters or blurry artifacts in the scene.

Several authors \citep{huang_2024_2d,guedon_2024_sugar,chen_2024_pgsr,dai_2024_high} have proposed flattening the kernels, with either strict 2D Gaussians \citep{huang_2024_2d} or flattened 3D Gaussians via loss regularization \citep{guedon_2024_sugar,chen_2024_pgsr}, to better model thin surfaces. These methods have shown that modifying the underlying primitives and enforcing optimization objectives on rendered geometry leads to improvements in geometric accuracy. While flat primitive help with surface alignment, diffuse ellipsoids still fail to accurately model depth discontinuities (see Fig.~\ref{fig:dtu_comp}).

Triangles are the fundamental primitives in computer graphics because they are necessarily planar and convex. They can approximate any surface as piecewise planar at any tolerance level, can directly model sharp discontinuities, and can be rendered rapidly without approximations.

We present a new methodology for 3D reconstruction, which we named Radiant Triangle Soup (\network{}) that enables gradient-based optimization of a Radiance Field (RF) using translucent triangle primitives as the scene representation (see Fig.~\ref{fig:tagline}). We provide a complete framework, including differentiable mechanisms for rasterization, as well as non-differentiable mechanisms for initialization, pruning, and densification. Our experiments show \network{} achieves competitive results in terms of both appearance and geometry (see Fig.~\ref{fig:dtu_qual}).

Conceptually, \network{} provides a feature that is not supported by any other GS scene representation: an avenue for explicit information sharing among 3D among primitives. Conventional Gaussian kernels interact via alpha-blending when rendered onto common pixels. Back-propagating from the loss at each pixel to the primitives is the only means of coordination across Gaussians. Using multiple images from various viewpoints gives rise to several indirect constraints on the primitives, but direct constraints are currently absent.

Conversely, \network{} is the first framework to enable direct information exchange among neighboring primitives. We formulate connectivity losses between neighboring triangle edges, allowing for a more direct and effective coordination of, and constraint on, primitive behavior throughout the optimization process. Even though using triangles as primitives alone achieves better geometric accuracy compared to Gaussian surfels, encouraging connectivity during optimization leads to more accurate surfaces with less floating artifacts (see Table~\ref{tab:ablation}).
    
Color expressivity on a per-primitive basis is another dimension where the \network{} representation excels. It is common practice to use a single color parametrization for each primitive. However, this may require large numbers of overlapping primitives in order to reconstruct details in the surfaces of the scene. Using triangles as the representation allows for each vertex to encode a separate color (in the form of Spherical Harmonics), leading to more expressive primitives through bilinear interpolation.

To summarize, the work presented here:

\begin{itemize}[leftmargin=*, itemsep=0pt, topsep=0pt]
    \item Develops a new scene representation through alpha-blending of triangle primitives.
    \item Introduces explicit 3D forces between primitives to encourage soft connectivity.
    \item Increases primitive expressivity via multi-color encoding.
\end{itemize}

\section{Related Work}
\label{sec:related}

We begin this section with Gaussian splatting formulations that favor surfaces and continue with methods that rely on non-Gaussian primitives. Surveys on other aspects of Gaussian splatting, omitted due to space limitations, include \citep{bao_2025_3d,dalal_2024_gaussian,luo_2024_review,wu_2024_recent}. The seminal work of \citet{kerbl_2023_3d} on 3D Gaussian Splatting  introduced an explicit alternative to NeRF \citep{mildenhall_2020_nerfa} that is able to achieve high-quality rendering at much higher speed. 3DGS relies on interleaved differentiable optimization and non-differentiable density control of the explicit representation. The optimization process decreases view synthesis errors for one of the training images at each iteration and density control guides the placement of primitives.

2D Gaussian Splatting (2DGS) \citep{huang_2024_2d} modified 3DGS to prioritize the reconstruction of surfaces, rather than volumetric material. This was accomplished by collapsing the 3D GS to 2D discs by setting the minimum eigenvalue of the Gaussian to 0. The authors of SuGaR \citep{guedon_2024_sugar} and PGSR \citep{chen_2024_pgsr} devised similar techniques for aligning 2D Gaussian with the surfaces via regularization and a multi-view loss, respectively. Gaussian Surfels \citep{dai_2024_high} rely on monocular surface normal estimates and a normal-depth consistency loss. Gaussian Opacity Fields (GOF) \citep{yu_2024_gaussian} extract surfaces as the zero-level set of 3D Gaussians. 3D-Half-Gaussian Splatting (3D-HGS) \citep{li_2024_3d} enables the representation of perfectly planar surfaces and hard edges by attaching a splitting plane to each Gaussian, aligning it to the local surface, and setting the density of one of the half-spaces to 0.

\begin{figure}[t]
\footnotesize
    \centering
    \begin{tabular}{c}
        \includegraphics[width=0.90\textwidth]{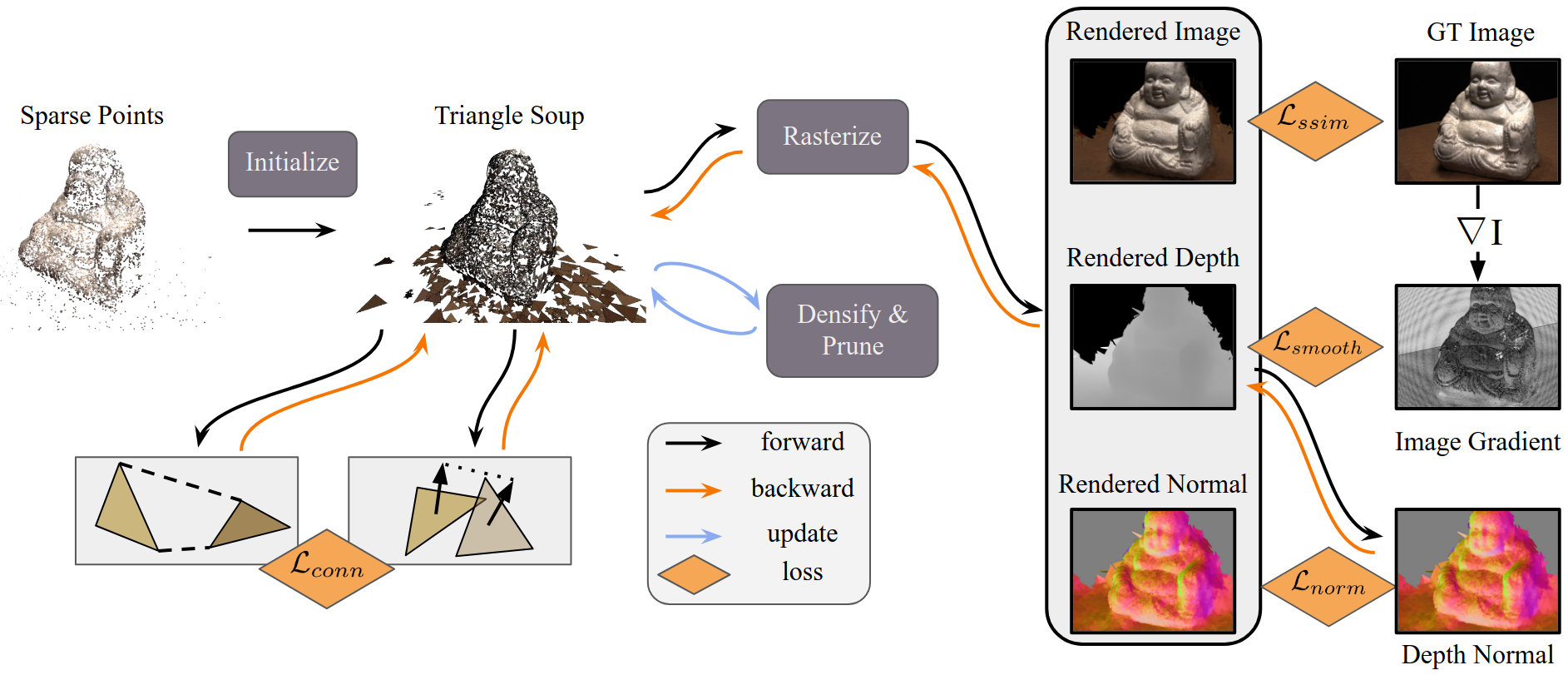}
    \end{tabular}
    \caption{We begin with a set of sparse points. From these points we initialize a set of triangles and compute nearest neighbor connections for each triangle edge. During optimization, we render an image, depth map, and normal map for each view. We compute both 2D loss (over all output renderings), as well as 3D loss (directly over the primitive connections). Adaptive densification is non-differentiable and is performed at set intervals throughout the optimization. Triangle edge neighbors are recomputed following adaptive densification.}
    \label{fig:pipeline}
    \vspace{-5mm}
\end{figure}

Representations based on NeRF and 3DGS have found great success, but have also inspired researchers to seek alternatives. NeuRBF \citep{chen_2023_neurbf} utilizes Radial Basis Functions (RBFs) to overcome limitations of NeRF due to the global nature of its MLP and features. GES \citep{hamdi_2024_ges} is based on an explicit representation which replaces the Gaussian kernel with a Generalized Exponential Function, overcoming the low-pass effect of the Gaussian and thus requiring  fewer primitives to represent the scene. Similar approaches based on smooth, non-Gaussian kernels include DARB-Splatting \citep{arunan_2025_darb}, SolidGS \citep{shen_2024_solidgs}, Beyond Gaussians \citep{chen_2024_gaussians} and Deformable Beta Splatting \citep{liu_2025_deformable}.

Besides the methods that force their Gaussians to be planar \citep{huang_2024_2d,guedon_2024_sugar,chen_2024_pgsr}, but still diffuse, there are others that represent the scene with collections of planar primitives. \citet{zanjani_2025_planar} initially use Gaussian splats to model the scene and then merge them into 3D planes, which are abundant in indoor scenes. PlanarSplatting \citep{tan_2025_planarsplatting} is also designed for indoor scenes using planes, initialized via monocular depth estimation, as the only primitives. TRIPS \citep{franke_2024_trips} is based on the principle that point primitives can be rasterized into an image pyramid from which the appropriate layer can be selected according to the size of the projected point. Holes can be filled by a small network yielding accurate, crisp renderings. Triangle Splatting \citep{held_2025_triangle} uses triangles with diffuse edges as primitives, but does not support any mechanism for them to interact directly with each other.

Non-planar primitives were introduced by BG-Triangle \citep{wu_2025_bg} which uses B{\'e}zier Gaussian triangles that are effective near boundaries, but comes at the cost of operating on complex, non-planar primitives that are hard to render. Quadratic Gaussian Splatting \citep{zhang_2025_quadratic} uses deformable quadratic surfaces as primitives and geodesic, instead of Euclidean, distance-based density distributions that adapt to the curvature of the primitives.

Another class of methods rely on volumetric primitives. LinPrim \citep{vonluetzow_2025_linprim} is based on linear solid primitives with triangular faces and performs volumetric rendering by computing the entrance and exit points of each ray through the primitives. 3D Convex Splatting (3DCS) \citep{held_2025_3d} was inspired by the limitation of GS that requires very large numbers of primitives to model hard edges and flat surfaces due to the diffuseness of the Gaussians. As the name suggests, the scene is represented by a set of polyhedral convexes which undergo volumetric rendering, pruning and splitting operations. The concepts of smoothness and sharpness introduced by 3DCS have inspired our diffuseness (see Section~\ref{sec:method}). Radiant Foam \citep{govindarajan_2025_radiant} enables modeling light transport phenomena, like reflection and refraction, by tessellating the space into Voronoi cells and iteratively optimizing the positions of the Voronoi vertices. All methods in this paragraph rely on different forms of volumetric elements with planar faces. Computing the intersections of each primitive with a ray requires multiple ray-triangle intersections, and is thus more expensive than \network{}.

A recent trend in the literature has been joint optimization of two representations: one for synthesizing novel views and one that is more faithful to the surfaces \citep{choi_2024_meshgs,jiang_2025_halogs}. MILo \citep{guedon_2025_milo} maintains Gaussians that are alpha-blended for view synthesis and a watertight mesh without texture that captures the geometry of the scene.

A few common themes emerge by analyzing the above methods. Most advocate the use of bounded primitives to enhance their ability to represent sharp edges and flat surfaces with small numbers of primitives. Our primitives are bounded but have diffuse boundaries to facilitate optimization. No other method, however, has a mechanism for direct inter-primitive communication like RTS.

\section{Method}
\label{sec:method}

\begin{wrapfigure}{r}{0.38\textwidth}
    \vspace{-20mm}
    \begin{center}
        \includegraphics[width=0.3\textwidth]{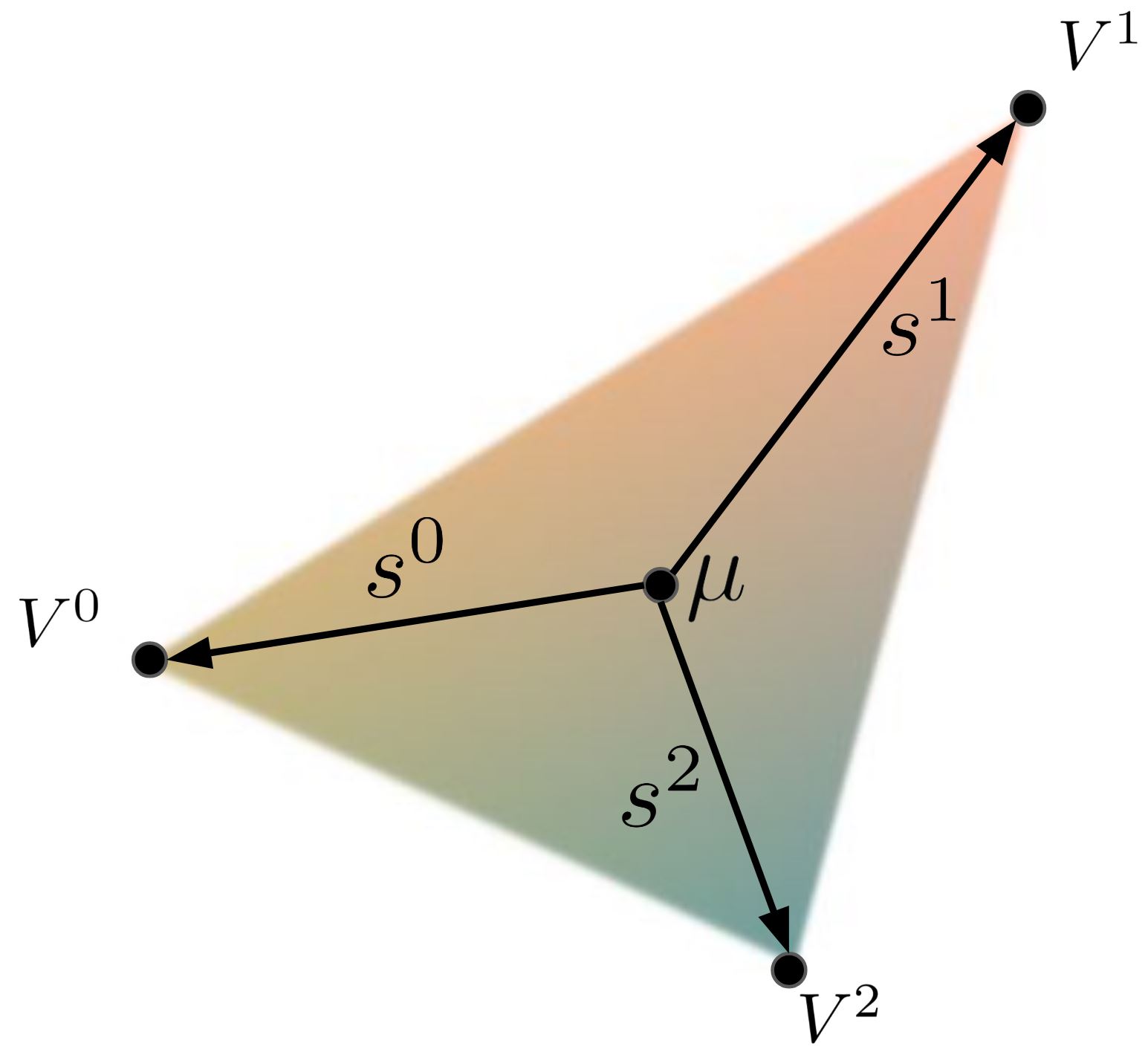}
    \end{center}
    \caption{Triangle parameterization. Each triangle is parameterized by the incenter $\mu$, and three scale values $\begin{bmatrix}s^0 & s^1 & s^2\end{bmatrix}$. These parameters, along with the rotation matrix $R$, define the coordinates of each vertex $V^j$.}
    \label{fig:param}
    \vspace{-1mm}
\end{wrapfigure}

Given a set of images together with camera poses and a set of sparse points $\mathcal{S}$ computed via Structure-From-Motion \citep{schoenberger_2016_structure}, we construct an explicit scene representation using triangle primitives, the parameterization and initiliazation of which we discuss in Sections~\ref{subsec:param} and~\ref{subsec:init}, respectively. The triangles are endowed with diffuse boundaries, similar to 3DCS \citep{held_2025_3d} (Section~\ref{subsec:diffuse}). To model the surfaces in the scene, we render into each camera the image, depth, and normal map of the triangles through alpha-blending (see Section~\ref{subsec:rasterization}). The triangle parameters are directly updated via back-propagation after computing losses between the rendering of the scene and the ground truth image from each view. We also include 2D losses on the rendered depth and normal maps (see Section~\ref{sec:optimization}). In our representation, primitives maintain \textit{soft} connectivity with their neighbors, discussed in Section~\ref{subsec:connectivity}, for which we compute additional loss directly over the connection orientations (see Section~\ref{sec:optimization}). Similar to previous works \citep{kerbl_2023_3d, huang_2024_2d, guedon_2024_sugar, chen_2024_pgsr, yu_2024_gaussian}, we develop a strategy for adaptive density control in order to facilitate the addition and removal of primitives in the scene (see Section~\ref{subsec:densification}). We show an overview of our optimization algorithm in Fig.~\ref{fig:pipeline}.

\subsection{Parameterization}
\label{subsec:param}
From the set $\mathcal{S}$ of sparse points, we first create an initial set of triangles $\mathcal{T}$. The triangle primitives $t_n \in \mathcal{T}$ in our scene representation are parametrized with,

\begin{equation}
    t_n = \{\mu, \Delta, s, R, \alpha, \sigma\}
\end{equation}
where $\mu$ is the incenter of the triangle, $\Delta = \begin{bmatrix}\delta^0 & \delta^1 & \delta^2\end{bmatrix}$ are the per-vertex Spherical Harmonics, $s = \begin{bmatrix}s^0 & s^1 & s^2\end{bmatrix}$ are the vertex scales, $R$ is the $3 \times 3$ rotation matrix, $\alpha$ is the opacity, and $\sigma$ is the diffuse scalar, discussed in Section~\ref{subsec:diffuse}.

\subsection{Initialization}
\label{subsec:init}
The coordinates of each sparse SfM point are used as the initialization for the incenter, and the point colors as the initialization for the zero-component of the Spherical Harmonics for all three vertices, which are then optimized separately. Similar to previous works, the scale for our primitives is computed based on the average distance to the three nearest neighboring points. Each triangle starts as equilateral, using the scale to parameterize the distance of each vertex from the incenter of the triangle $\mu$ along the bisectors of the angles (see Fig.~\ref{fig:param}). The diffuse scalar $\sigma$ is also set as a function of the average distance to the three nearest neighboring points. The parameterization of the initial rotation matrix for each triangle is defined such that the $\overrightarrow{V^2V^0}$ edge and the triangle normal $\widehat{n}$ are used as the $x-$ and $z-$ axis, respectively. We then initialize each triangle by applying a random rotation to this matrix.

\subsection{Diffuse Primitive Boundaries}
\label{subsec:diffuse}
\begin{figure}[t]
\footnotesize
    \centering
    \begin{tabular}{ccc}
        \includegraphics[width=0.30\textwidth]{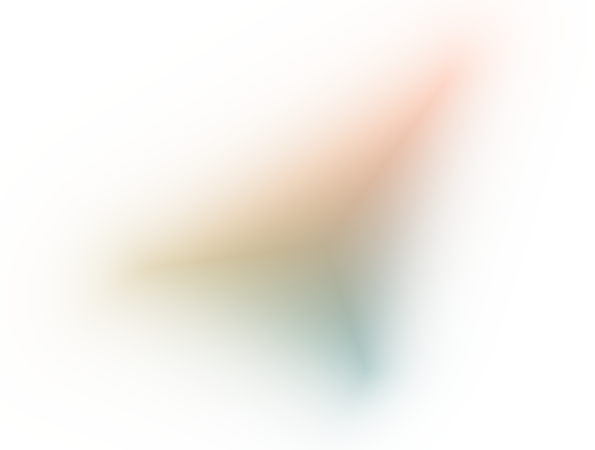} &
        \includegraphics[width=0.30\textwidth]{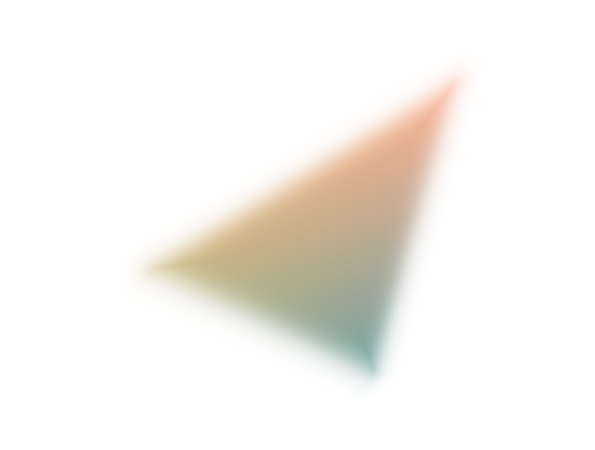} &
        \includegraphics[width=0.30\textwidth]{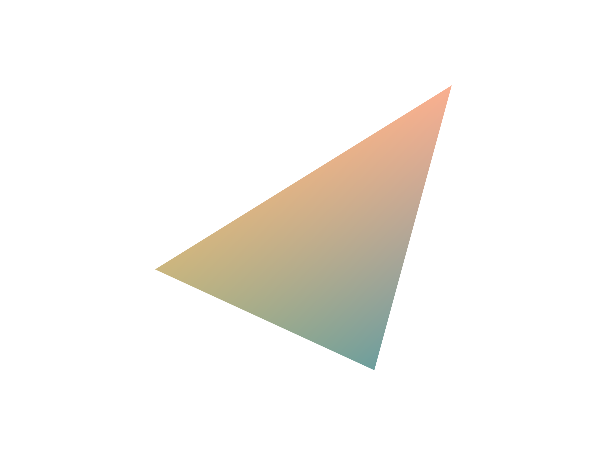} \\
        & $\longrightarrow$ increasing $\sigma$ &
    \end{tabular}
    \caption{Controlling triangle diffuseness with $\sigma$. Small values of $\sigma$ \textbf{(left)} result in diffuse triangles with extended range of influence. Large values of $\sigma$ \textbf{(right)} result in sharp triangles with no blurring across edges.}
    \label{fig:sigma}
    \vspace{-2mm}
\end{figure}

Alpha-blending is the primary driving force of optimization for all splatting frameworks. Blending across primitives facilitates optimization via gradient descent and encourages, among other behaviors, movement in primitive position and orientation. It is challenging in practice to optimize a scene with fully opaque primitives that do not smoothly blend with one another. It is therefore desirable to make the triangle primitives diffuse near the edges. Taking inspiration from 3DCS \citep{held_2025_3d}, we parameterize the response from each triangle as a function of the signed distance from the ray-triangle intersection to the nearest edge,
\begin{equation}
 w_{\sigma} = \frac{1}{1+e^{(\sigma l)}}
\end{equation}
where $w_{\sigma}$ is the diffuse weight, $l$ is the signed distance between the intersection and the nearest triangle edge in the plane of the triangle, and the scalar $\sigma$ is an optimizable parameter of each triangle that controls the level of diffuseness. As $\sigma$ increases, the boundary of the triangle becomes less diffuse, creating sharper primitive renderings (see Fig.~\ref{fig:sigma}). For the signed distance, $l<0$ occurs when the intersection point lies outside of the triangle boundary. This formulation slightly deviates from that of 3DCS, as the signed distance is computed directly, compared to their smooth approximation.

\subsection{Rasterization}
\label{subsec:rasterization}
Throughout this paper, we use barycentric coordinates, $\lambda=[\lambda^0, \lambda^1, \lambda^2]^T$, to rasterize triangles. Please see Section~\ref{sec:background} for more details. With the barycentric coordinates of the ray-triangle intersections and the diffuse weight computed, we interpolate the color for the current pixel w.r.t. the colors of each vertex weighted by the barycentric coordinates, $c_n = \begin{bmatrix} c^0 & c^1 & c^2 \end{bmatrix} \lambda$, where $c^0$, $c^1$, and $c^2 \in \mathbb{R}^3$ are computed from the SH components of each vertex, and $\lambda$ are the barycentric coordinates of the intersection point.

We aggregate the contribution to the current output pixel, $i$, from each intersected primitive, $c_i = \sum w_n c_n$, where $w_n = \alpha w_{\sigma} T$, and $T = \Pi_{j=1}^{i-1} (1 - \alpha_j)$ is the transmittance for the current primitive \citep{kerbl_2023_3d}.

Previous works \citep{huang_2024_2d, chen_2024_pgsr, yu_2024_gaussian} provide two methods for rendering per-pixel depth; (1)~computing the average weighted intersection depth of all traversed primitives (mean depth), and (2)~using only the depth of the primitive that causes the transmittance $T$ to exceed $0.5$ (median depth). In our work, we use median depth. Using the mean depth encourages the formation of many translucent layers of primitives. Rendering median depth removes this blending and helps guide the formation of surfaces. We directly use $d$ from Eq.~\ref{eq:depth} for the depth of each intersection.

Additionally, we compute the surface normal, $\widehat{n}_n$, for each triangle as the cross-product between two edges.
Unlike the rendered depth, to render per-pixel surface normals, we follow previous work \citep{huang_2024_2d, chen_2024_pgsr} and alpha-blend all the surface normals of all intersected primitives. The intuition is that through a normal consistency loss (see Section~\ref{sec:optimization}), the blended normals must align with the normals computed from the median depth map. This encourages all the intersected triangles to align (and eventually collapse onto) the median surface. Thus, we render per-pixel surface normals through alpha-blending, $\overrightarrow{n_i} = \sum w_n \widehat{n}_n$.

\begin{wrapfigure}{r}{0.35\textwidth}
    \vspace{-15mm}
    \footnotesize
    \centering
    \begin{tabular}{c}
        \includegraphics[width=0.34\textwidth]{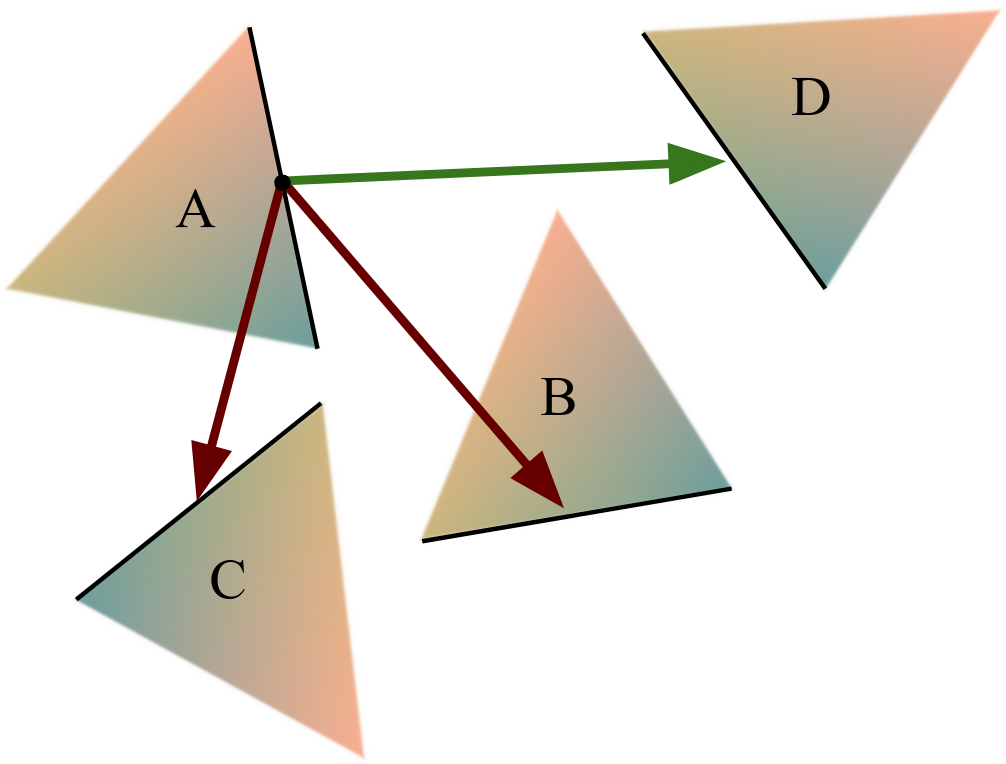}
    \end{tabular}
    \caption{Examples of edge association for connecting nearby triangles. For simplicity, it is assumed the triangles are \textit{co-planar} in this figure.}
    \label{fig:connect}
    \vspace{-5mm}
\end{wrapfigure}

\subsection{Connectivity}
\label{subsec:connectivity}
To enable soft connectivity among primitives and encourage flat, connected surfaces, we add a connectivity term to the optimization objective. For each triangle edge, we assign, and periodically update, a connection to the nearest neighboring triangle edge. 
Taking inspiration from the energy functions introduced in 3D scene flow estimation \citep{vogel_2015_3d}, the connectivity term of the loss increases according to the distance between their vertices and the inner product between their normals, discussed in Section~\ref{sec:optimization}.

Naively connecting with the nearest edge without considering the relative orientations may cause connections that
would require large changes in rotation to either triangle, leading to undesirable behaviors during optimization.
To prevent this behavior, connections are only established if the triangle edges are "facing" each other.
In practice, we use the inner product between the unit vectors orthogonal to the triangle edges (in the plane of each triangle) as the criterion for establishing connections.
We provide an example in Fig.~\ref{fig:connect} where a connection to the highlighted edge of triangle D is valid, while the other two connections would cause large rotations.

\subsection{Adaptive Density Control}
\label{subsec:densification}

\begin{wrapfigure}{r}{0.30\textwidth}
    \vspace{-10mm}
    \centering
    \begin{tabular}{c}
        \includegraphics[width=0.28\textwidth]{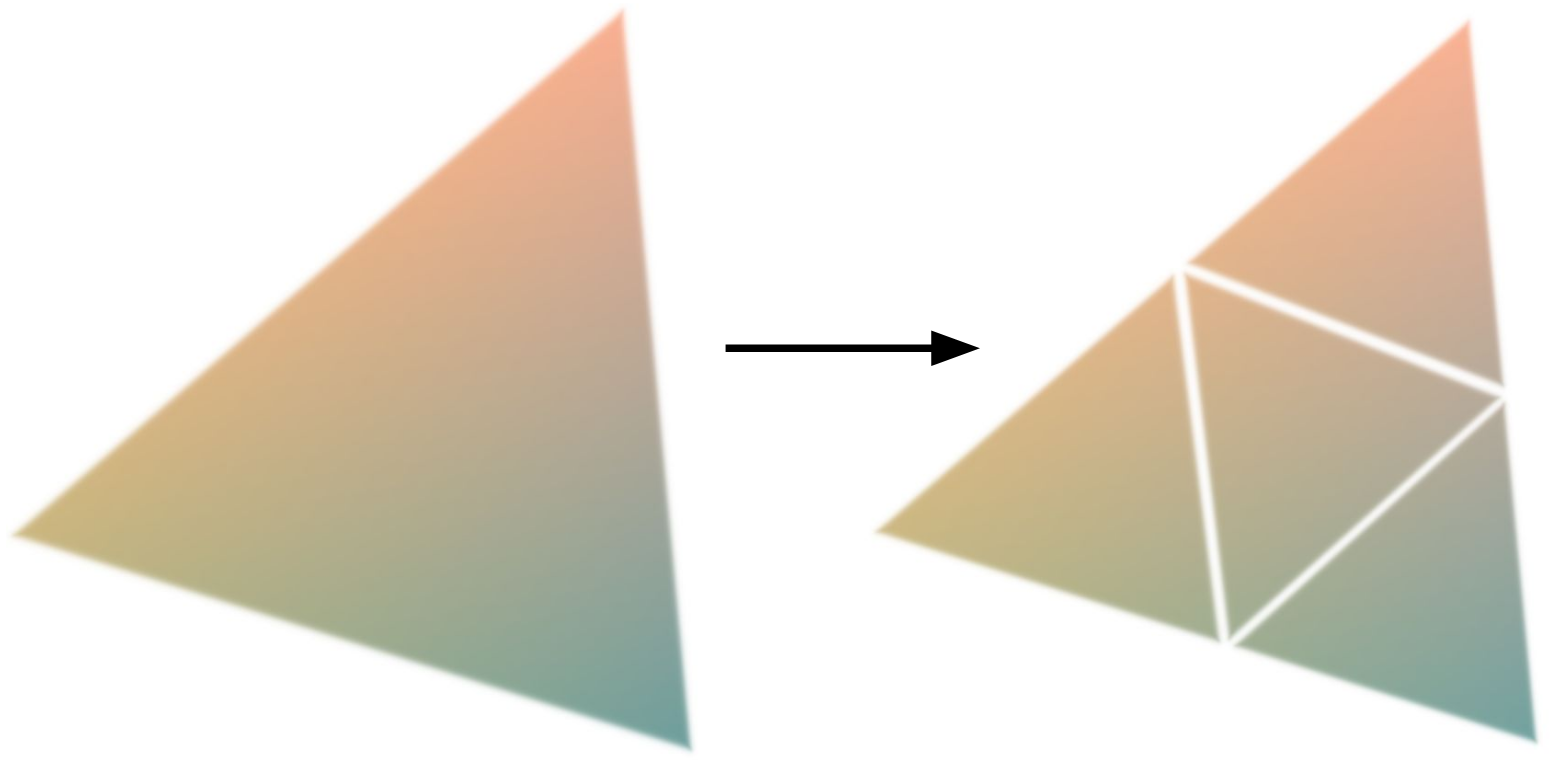}
    \end{tabular}
    \caption{Triangles are \textit{split} with interpolated colors per-vertex.}
    \label{fig:densify}
    \vspace{-5mm}
\end{wrapfigure}

We perform adaptive density control through the process of cloning, splitting and pruning triangles. Previous works \citep{kerbl_2023_3d, huang_2024_2d, guedon_2024_sugar, chen_2024_pgsr} perform the cloning and splitting procedures by duplicating primitives conditioned on scale and position gradients. In order to properly split large triangles, we must split them into four sub-triangles. During early iterations in the optimization, when triangles are split, the sub-triangles move independently to better align with surfaces in the scene. During the later stages when most of the surfaces have formed, the triangles split to enable more detailed surface color representation, remaining relatively attached via the connectivity forces. See Fig.~\ref{fig:densify} for a visualization of the splitting procedure. 

Similar to previous work, we directly clone small primitives selected for densification. Much like 2DGS \citep{huang_2024_2d}, our method does not directly rely on the gradient of the projected 2D primitive center. Instead of computing an approximation via projecting 3D gradients into the camera plane \citep{huang_2024_2d}, we directly condition densification on the magnitude of the incenter gradient $\nabla \mu$ in 3D.

To remove uninformative triangles, we prune primitives that meet the following criteria: (i) triangles that are transparent ($\alpha < 0.05$), (ii) triangles with one or fewer edge connections, (iii) triangles that do not intersect the camera frustum of at least three views after each epoch.

\section{Optimization}
\label{sec:optimization}
Our objective function comprises terms computed in 2D (on the images plane), and in 3D (on the triangles).

\begin{table}[b]
\caption{Chamfer distance evaluation on scenes from DTU \citep{aanaes_2016_large}. Following previous literature, we average the accuracy and completeness Chamfer distances on the widely used evaluation set. Chamfer distances are measured in millimeters. The best results are in boldface and the second best are underlined.}
\label{tab:dtu_geo}
\centering
\resizebox{\linewidth}{!}{%
\begin{tabular}{l|ccccccccccccccc|c}
\hline
    \multirow{2}{*}{Method} &
    \multicolumn{16}{c}{DTU} \\ \cline{2-17}
        &
        24 &
        37 &
        40 &
        55 &
        63 &
        65 &
        69 &
        83 &
        97 &
        105 &
        106 &
        110 &
        114 &
        118 &
        122 &
        Mean (mm)$\downarrow$ \\ \hline\hline
    3DGS \citep{kerbl_2023_3d} &
        2.14 &
        1.53 &
        2.08 &
        1.68 &
        3.49 &
        2.21 &
        1.43 &
        2.07 &
        2.22 &
        1.75 &
        1.79 &
        2.55 &
        1.53 &
        1.52 &
        1.50 &
        1.96 \\ 
    SuGaR \citep{guedon_2024_sugar} &
        1.47 &
        1.33 &
        1.13 &
        0.61 &
        2.25 &
        1.71 &
        1.15 &
        1.63 &
        1.62 &
        1.07 &
        0.79 &
        2.45 &
        0.98 &
        0.88 &
        0.79 &
        1.33 \\ 
    2DGS \citep{huang_2024_2d} &
        0.48 &
        0.91 &
        0.39 &
        0.39 &
        1.01 &
        0.83 &
        0.81 &
        1.36 &
        1.27 &
        0.76 &
        0.70 &
        1.40 &
        0.40 &
        0.76 &
        0.52 &
        0.80 \\ 
    GOF \citep{yu_2024_gaussian} &
        0.50 &
        0.82 &
        \textbf{0.37} &
        \underline{0.37} &
        1.12 &
        \underline{0.74} &
        0.73 &
        1.18 &
        1.29 &
        0.68 &
        0.77 &
        0.90 &
        0.42 &
        0.66 &
        0.49 &
        0.74 \\
    Gaussian Surfels \citep{dai_2024_high} &
        0.66 &
        0.93 &
        0.54 &
        0.41 &
        1.06 &
        1.14 &
        0.85 &
        1.29 &
        1.53 &
        0.79 &
        0.82 &
        1.58 &
        0.45 &
        0.66 &
        0.53 &
        0.88 \\
    PGSR \citep{chen_2024_pgsr} &
        \textbf{0.36} &
        \textbf{0.57} &
        \underline{0.38} &
        \textbf{0.33} &
        \underline{0.78} &
        \textbf{0.58}&
        \textbf{0.50} &
        \underline{1.08} &
        \textbf{0.63} &
        \underline{0.59} &
        \textbf{0.46} &
        \textbf{0.54} &
        \textbf{0.30} &
        \textbf{0.38} &
        \textbf{0.34} &
        \textbf{0.52} \\
    TriangleSplatting \citep{held_2025_triangle} & 
        0.98 &
        1.07 &
        1.07 &
        0.51 &
        1.67 &
        1.44 &
        1.17 &
        1.32 &
        1.75 &
        0.98 &
        0.96 &
        1.11 &
        0.56 &
        0.93 &
        0.72 &
        1.06 \\
    \textbf{\network{}} &
        \underline{0.42} &
        \underline{0.61} &
        0.74 &
        0.39 &
        \textbf{0.53} &
        0.86 &
        \underline{0.70} &
        \textbf{0.84} &
        \underline{0.72} &
        \textbf{0.37} &
        \underline{0.68} &
        \underline{0.87} &
        \underline{0.34} &
        \underline{0.58} &
        \underline{0.44} &
        \underline{0.61} \\ \hline
\end{tabular}%
}
\end{table}

\newcommand\heightdtuqual{0.17\textwidth}
\begin{figure}[t]
\footnotesize
    \centering
\resizebox{\linewidth}{!}{%
    \begin{tabular}{ccccc}
        \includegraphics[height=\heightdtuqual]{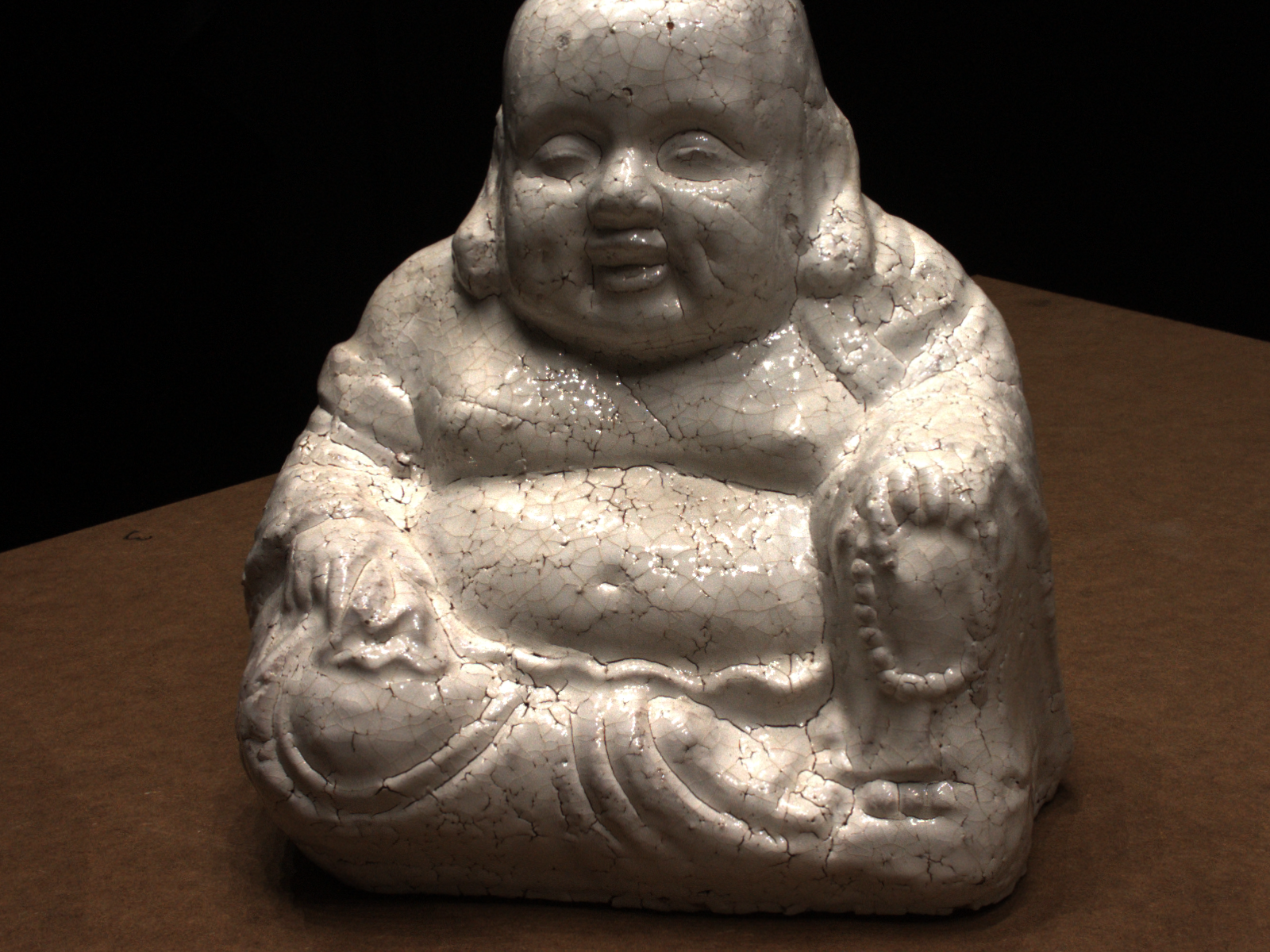} &
        \includegraphics[height=\heightdtuqual]{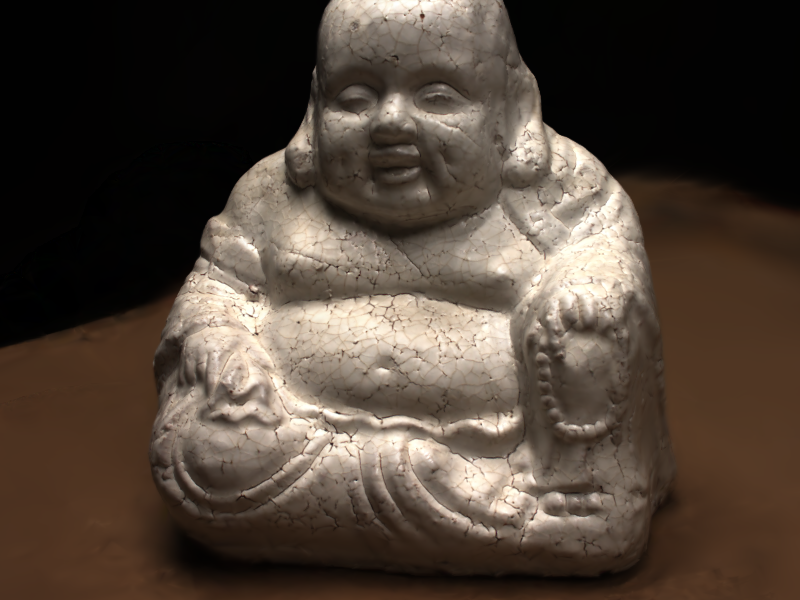} &
        \includegraphics[height=\heightdtuqual]{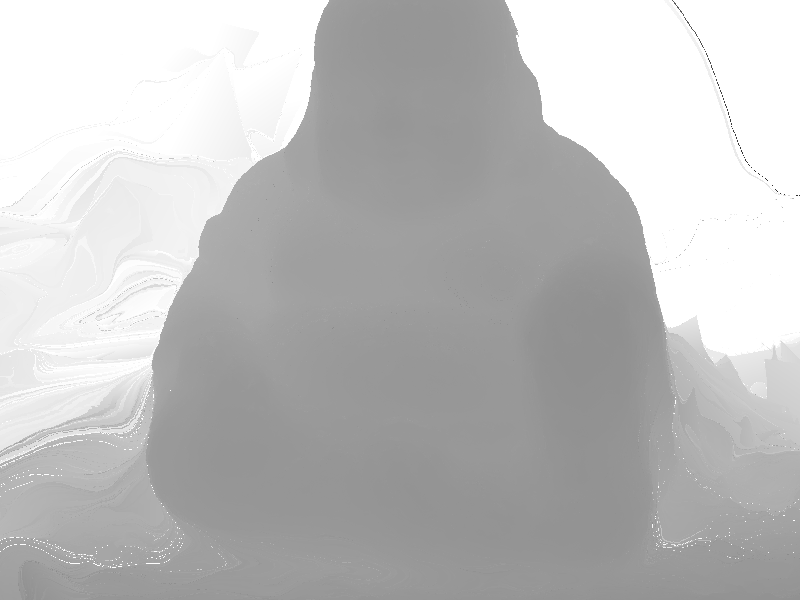} &
        \includegraphics[height=\heightdtuqual]{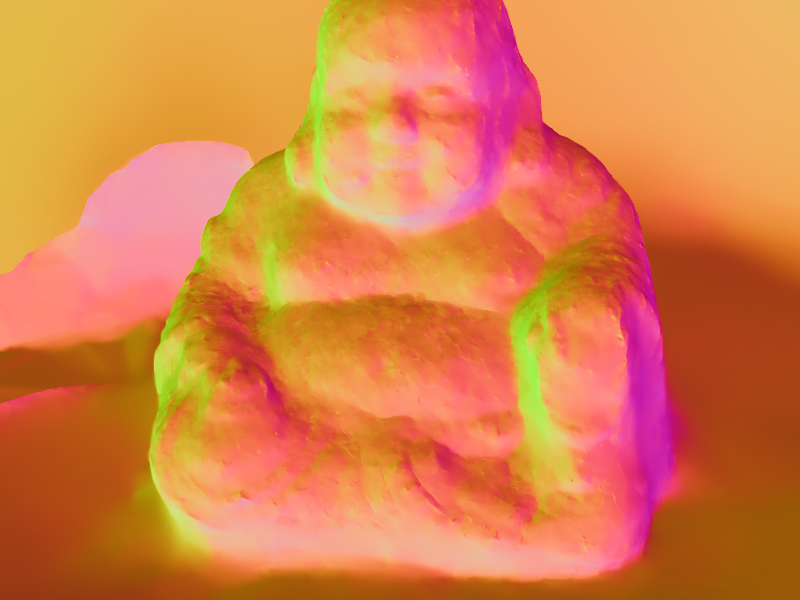} &
        \includegraphics[height=\heightdtuqual]{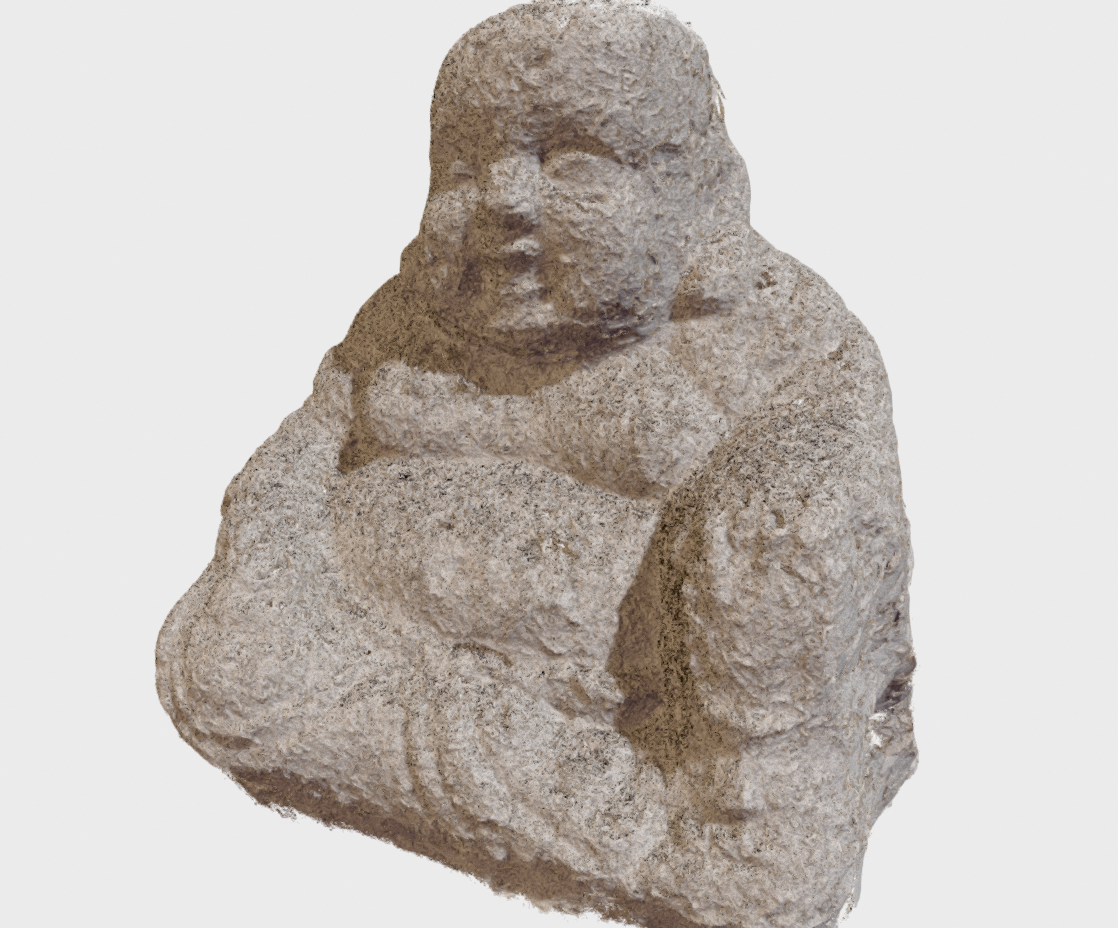} \\
        \includegraphics[height=\heightdtuqual]{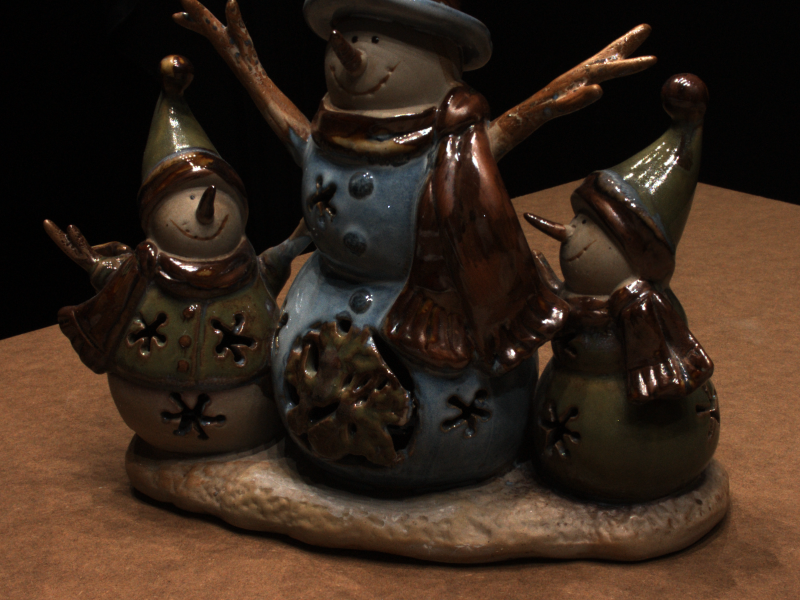} &
        \includegraphics[height=\heightdtuqual]{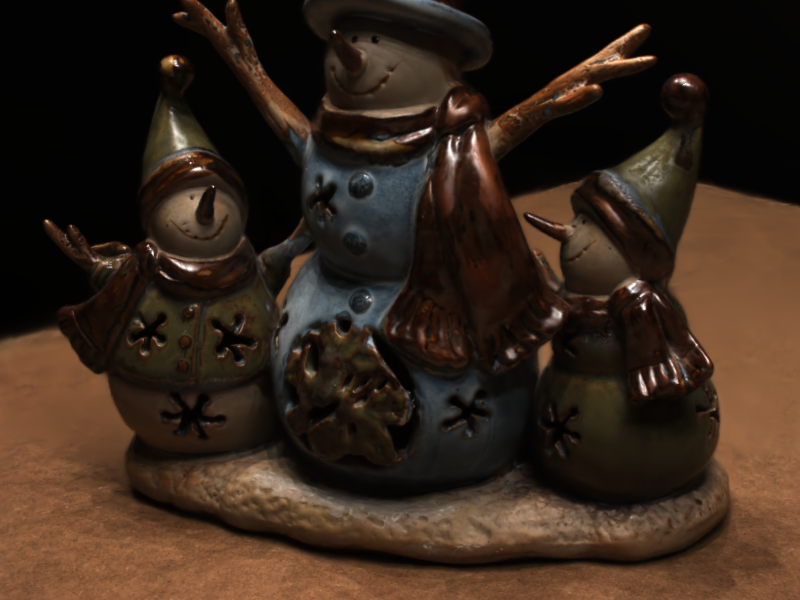} &
        \includegraphics[height=\heightdtuqual]{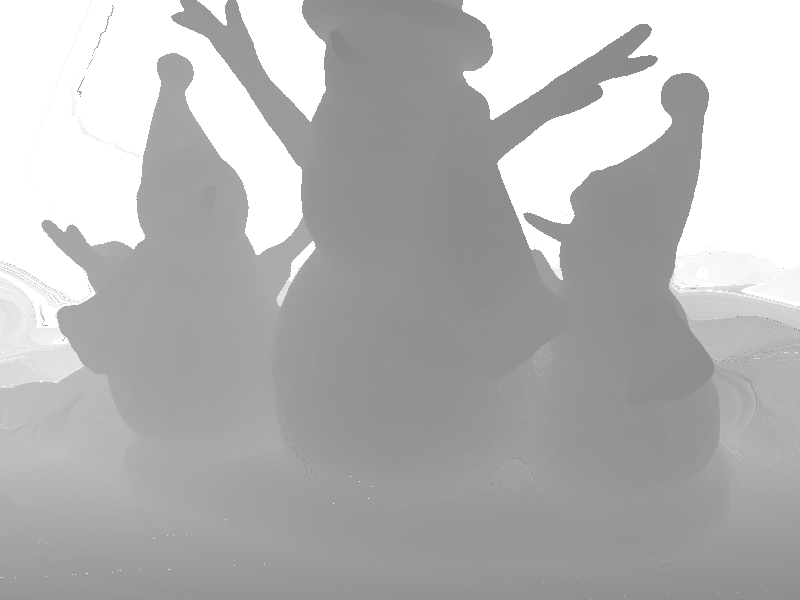} &
        \includegraphics[height=\heightdtuqual]{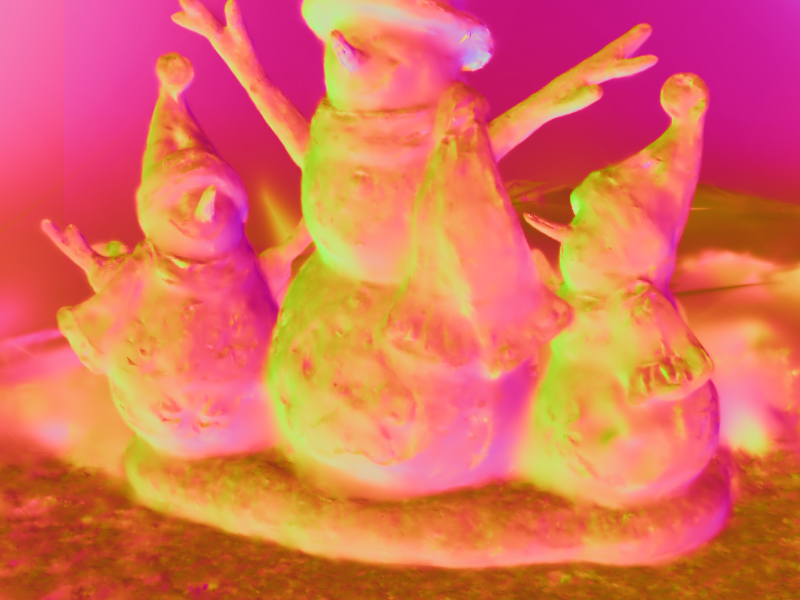} &
        \includegraphics[height=\heightdtuqual]{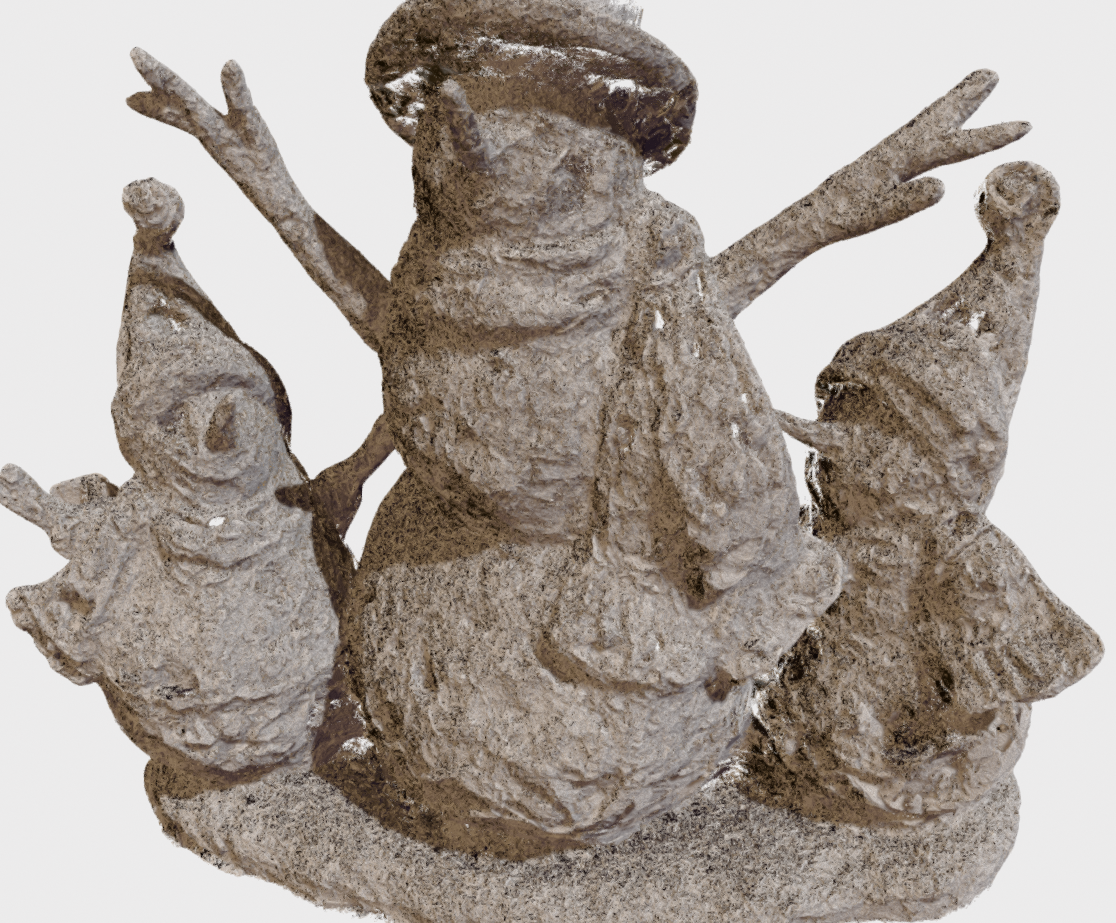} \\
        \includegraphics[height=\heightdtuqual]{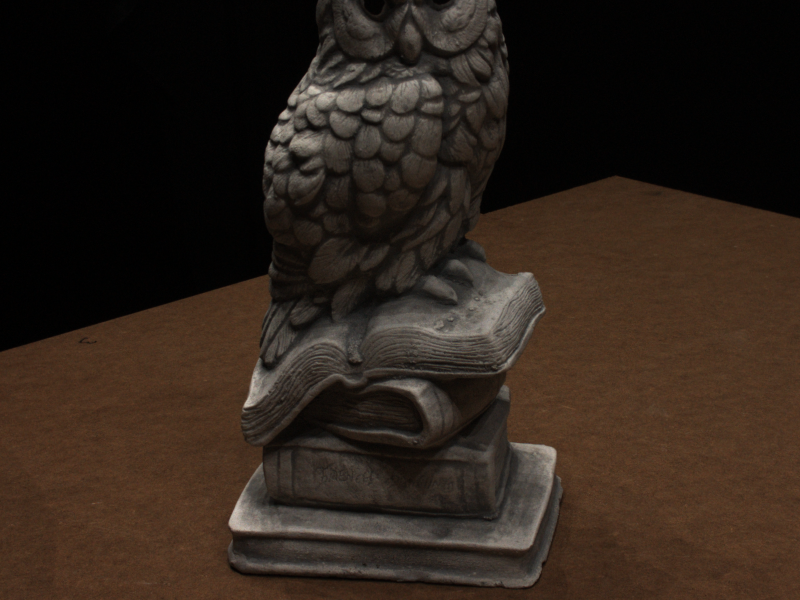} &
        \includegraphics[height=\heightdtuqual]{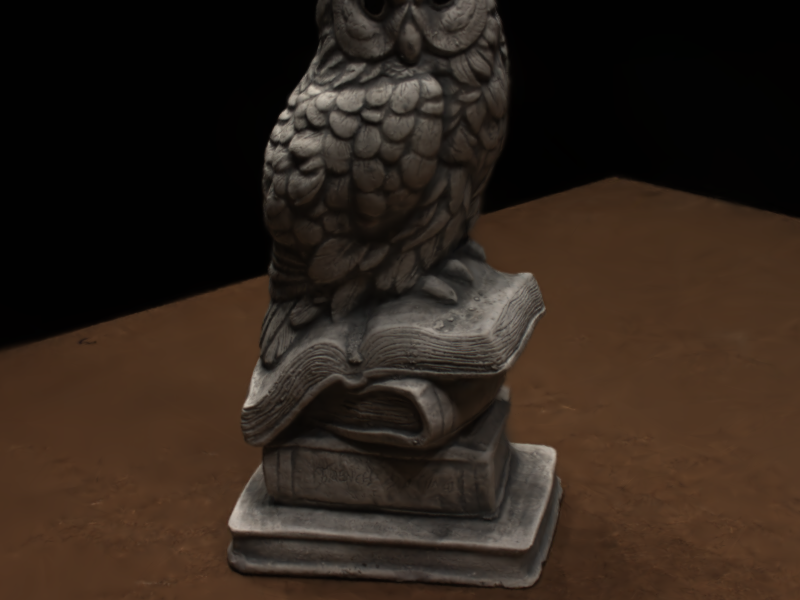} &
        \includegraphics[height=\heightdtuqual]{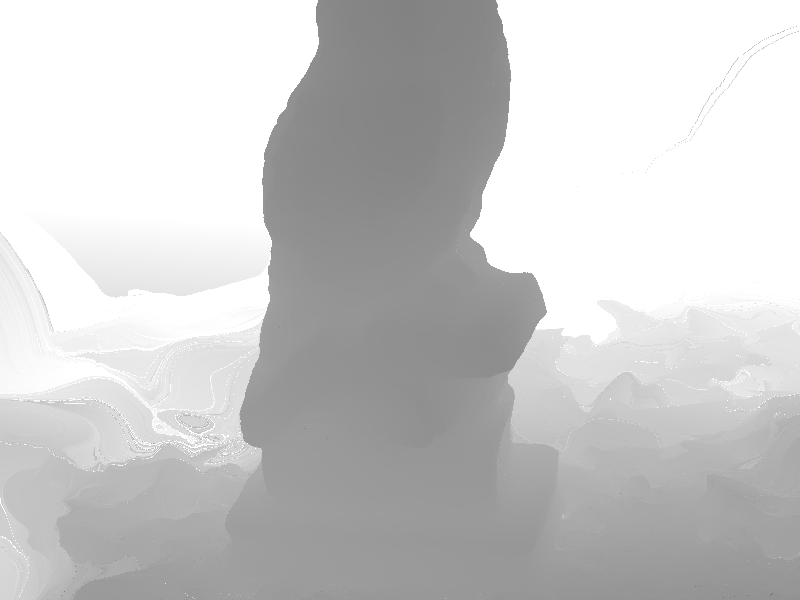} &
        \includegraphics[height=\heightdtuqual]{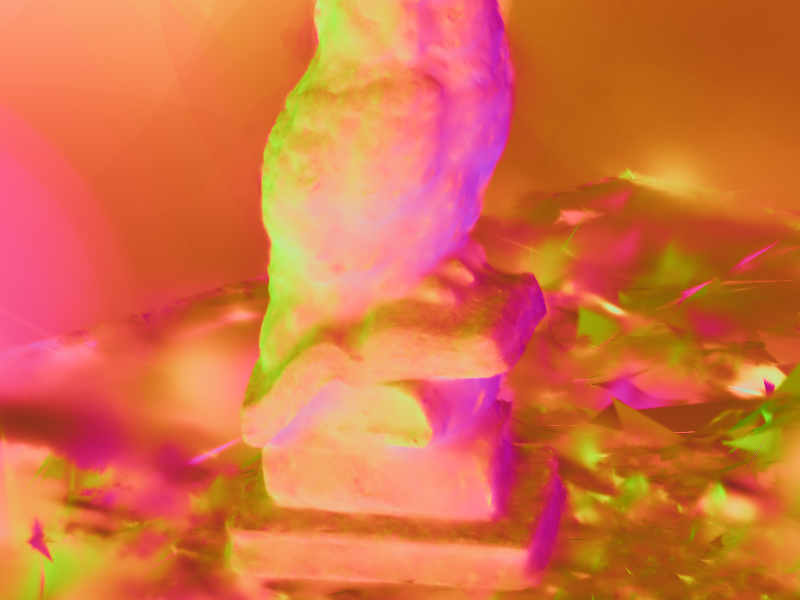} &
        \includegraphics[height=\heightdtuqual]{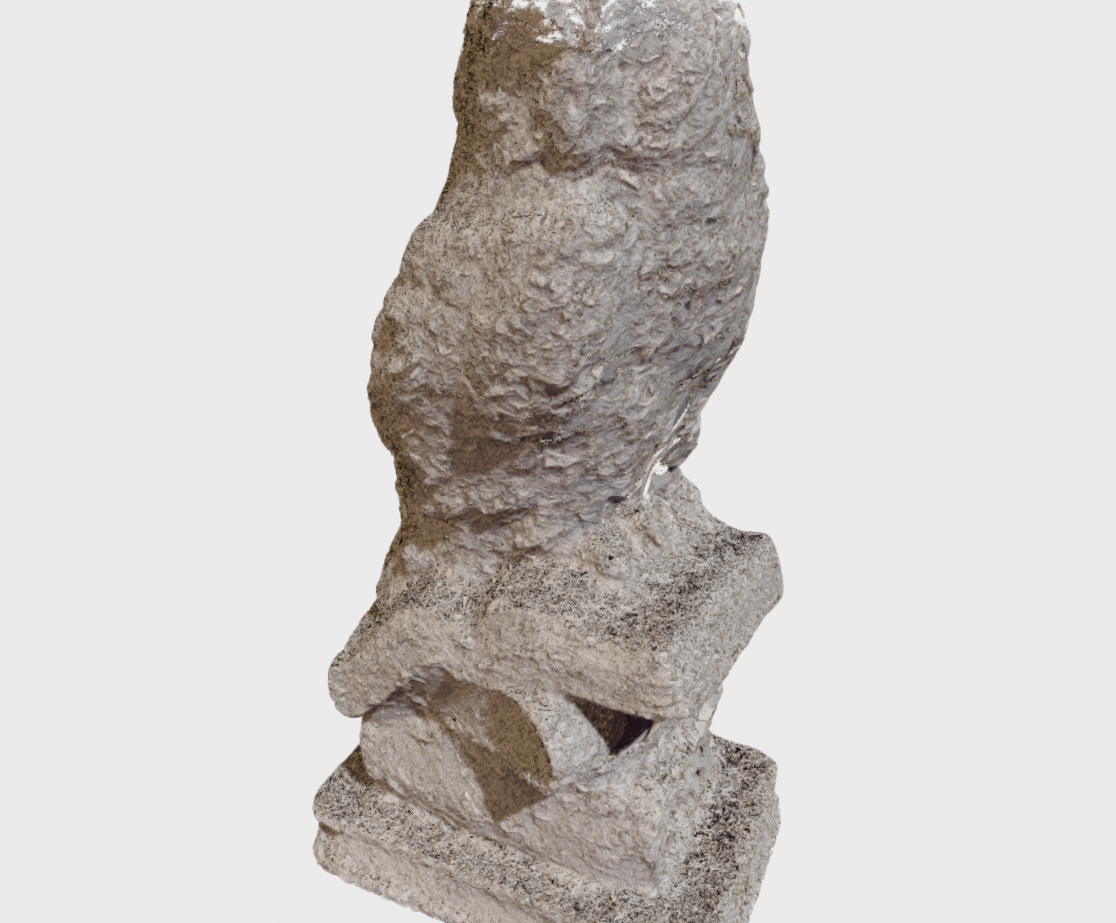} \\
        Ground Truth & Rendered Image & Rendered Depth & Rendered Normal & Point Cloud \\
    \end{tabular}
    }
    \caption{Qualitative results on the DTU dataset \citep{aanaes_2016_large}. \network{} estimates high-quality geometry, maintaining thin structures, such as the carrot noses and branches (middle row) and the book edges (bottom row), while effectively modeling smooth surfaces, such as the Buddha (top row).}
    \label{fig:dtu_qual}
    \vspace{-4mm}
\end{figure}

\subsection{Rendering Losses}
Following previous work \citep{kerbl_2023_3d, huang_2024_2d}, we compute the SSIM between the rendered and input images, $\Lagr_{ssim}$, and apply a normal consistency loss \citep{huang_2024_2d}, $\Lagr_{norm}$, to help locally align the triangles with the rendered surface. Please see Section~\ref{sec:sup_loss} the supplement.

From the unsupervised depth estimation literature \citep{chang_2022_rca, godard_2017_unsupervised, mahjourian_2018_unsupervised}, we adopt a smoothness term on the rendered depth map conditioned on the gradient of the input image. This penalizes large gradients in the rendered depth map where we have small gradients in the input image,
\begin{equation}
    \Lagr_{smooth} = \frac{1}{N} \sum_{i,j} ||\partial_x D_{i,j}||  e^{||\partial_x I_{i,j}||} + ||\partial_y D_{i,j}||  e^{||\partial_y I_{i,j}||}
\end{equation}

\subsection{Scene Loss}
To encourage connectivity between primitives, we penalize the mean of the $L2$ distance between the connected vertex pairs of neighboring triangle edges,
\begin{equation}
    \Lagr_{conn} = \sum_{a \in \Omega} \frac{1}{2} (||V_a^1 - V_b^1||_2 + ||V_a^2 - V_b^2||_2) + (1 - \widehat{n}_a^T \widehat{n}_b)
\end{equation}
where $V_a^j$ and $V_b^j$ are the $j^{th}$ vertex pair of connected edges $a$ and $b$, respectively. $\widehat{n}_a$ and $\widehat{n}_b$ are the normals of the connecting triangles and $\Omega$ is the set of all triangles that intersect the current camera frustum. Applying this loss to invisible triangles without rendering losses leads to over-smoothing.

The final objective is a weighted summation of all terms:
\begin{equation}
    \Lagr = \omega_0 \Lagr_{ssim} + \omega_1 \Lagr_{norm} + \omega_2 \Lagr_{smooth} + \omega_3 \Lagr_{conn}
\end{equation}

\section{Experiments}
\label{sec:experiments}

\newcommand\heightdtucomp{0.20\textwidth}
\begin{figure}[t]
\footnotesize
\centering
\resizebox{0.75\linewidth}{!}{%
    \begin{tabular}{ccc}
        \includegraphics[height=\heightdtucomp]{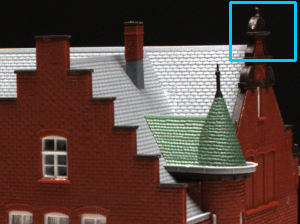} &
        \includegraphics[height=\heightdtucomp]{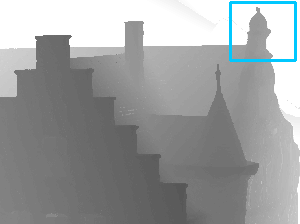} &
        \includegraphics[height=\heightdtucomp]{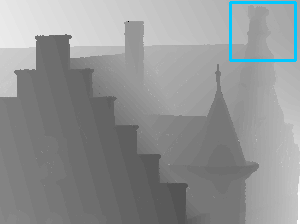} \\
        \includegraphics[height=\heightdtucomp]{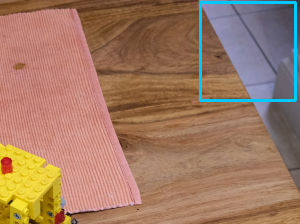} &
        \includegraphics[height=\heightdtucomp]{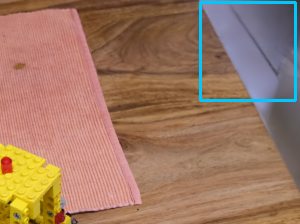} &
        \includegraphics[height=\heightdtucomp]{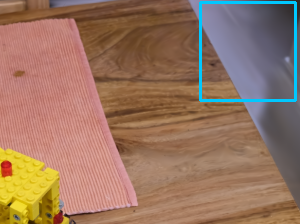} \\
        Ground Truth & \textbf{\network{}} & 2DGS \citep{huang_2024_2d} \\
    \end{tabular}
    }
    \caption{Qualitative comparison between \network{} and 2DGS \citep{huang_2024_2d} on; \textbf{Top} - \textit{scan024} from the DTU dataset \citep{aanaes_2016_large}, \textbf{Bottom} - \textit{kitchen} from the mip-NeRF 360 dataset \citep{barron_2022_mip}. \network{} is substantially more precise at estimating the geometry at discontinuities and rendering fine details, shown in the areas marked by the blue rectangles.}
    \label{fig:dtu_comp}
\end{figure}

\subsection{Implementation Details}
\label{sec:implementation}
We implement the majority of our \network{} framework in Python using PyTorch \citep{paszke_2019_pytorch}. For rasterization, we develop custom CUDA kernels for both the forward and backward pass. We run all our experiments on a single NVIDIA RTX A6000. For our loss weights, we choose $\omega = \begin{bmatrix} 1.0 & 0.05 & 50.0 & 1000.0\end{bmatrix}$ empirically.

Following state-of-the-art Multi-View Stereo methods \citep{yao_2018_mvsnet, yang_2022_non, mi_2022_generalized}, we directly generate a 3D point cloud for geometric evaluation from the rendered depth maps \textit{without performing any TSDF fusion}. To generate each point cloud, we use simple heuristic filtering on each depth map, similar to the post-processing presented in GBiNet \citep{mi_2022_generalized}. For each depth map, we measure the reprojection error of the depth values at every pixel using neighboring views and filter pixels based on this error. All depth estimates with a low reprojection error are back-projected to 3D points, forming the combined point cloud.

\begin{table}[b]
\caption{Novel View Synthesis on all scenes from the mip-NeRF 360 dataset \citep{barron_2022_mip}.}
\label{tab:mipnerf360}
\centering
\resizebox{0.75\linewidth}{!}{%
\begin{tabular}{l|ccc|ccc}
\hline
    \multirow{2}{*}{Method} &
    \multicolumn{3}{c|}{Outdoor} &
    \multicolumn{3}{c}{Indoor} \\ \cline{2-7} &
        PSNR$\uparrow$ & SSIM$\uparrow$ & LPIPS$\downarrow$ & 
        PSNR$\uparrow$ & SSIM$\uparrow$ & LPIPS$\downarrow$ \\ \hline\hline
    NeRF \citep{mildenhall_2020_nerfa} &
        21.46 & 0.458 & 0.515 & 
        26.84 & 0.790 & 0.370 \\ 
    Deep Blending \citep{hedman_2018_deep} &
        21.54 & 0.524 & 0.364 & 
        26.40 & 0.844 & 0.261 \\
    Instant NGP \citep{mueller_2022_instant} &
        22.90 & 0.566 & 0.371 & 
        29.15 & 0.880 & 0.216 \\
    MipNeRF360 \citep{barron_2022_mip} &
        24.47 & 0.691 & 0.283 & 
        \textbf{31.72} & 0.917 & 0.180 \\ \hline
    SuGaR \citep{guedon_2024_sugar} &
        22.93 & 0.629 & 0.356 & 
        29.43 & 0.906 & 0.225 \\
    3DGS \citep{kerbl_2023_3d} &
        \underline{24.64} & 0.731 & 0.234 & 
        30.41 & 0.920 & 0.189 \\    
    2DGS \citep{huang_2024_2d} &
        24.34 & 0.717 & 0.246 & 
        30.40 & 0.916 & 0.195 \\ 
    GOF \citep{yu_2024_gaussian} &
        \textbf{24.82} & \underline{0.750} & \textbf{0.202} & 
        30.79 & 0.924 & 0.184 \\ 
    PGSR \citep{chen_2024_pgsr} &
        24.76 & \textbf{0.752} & \underline{0.203} & 
        30.36 & \textbf{0.934} & \underline{0.147} \\ 
    3DCS \citep{held_2025_3d} &
        24.07 & 0.700 & 0.238 & 
        \underline{31.33} & 0.927 & 0.166 \\
    TriangleSplatting \citep{held_2025_triangle} &
        24.27 & 0.722 & 0.217 &
        30.80 & \underline{0.928} & 0.160 \\
    \textbf{\network{}} &
        21.41 & 0.657 & 0.349 & 
        30.28 & 0.921 & \textbf{0.130} \\ \hline
\end{tabular}%
}
\end{table}

\subsection{Evaluation}
We test our framework on the DTU dataset \citep{aanaes_2016_large}, an indoor dataset that contains images of 124 scenes taken from a camera mounted on an industrial robot arm. All scenes share the same camera trajectories, with ground-truth point clouds captured via structured light.

We evaluate our new approach on the DTU dataset and record the Chamfer distance in Table~\ref{tab:dtu_geo}. We show competitive results alongside the leading state-of-the-art planar GS methods. Across the test set, \network{} is the most geometrically accurate in several scenes and second most overall. While PGSR demonstrates impressive reconstruction results, the algorithm utilizes a full suite of multi-view objective functions that significantly improve the geometric reconstruction quality. We provide qualitative results on the DTU dataset in Fig.~\ref{fig:dtu_qual}, showing visualizations of the rendered images, depth maps, normal maps, and final point clouds for three scenes. In Fig.~\ref{fig:dtu_comp} (\textit{top}), we provide a comparison of depth map renderings between \network{} and 2DGS~\citep{huang_2024_2d}. \network{} is able to reconstruct fine details on the surfaces of objects that are typically blurred with Gaussian representations.

\begin{table}[t]
\centering
\caption{Ablation study on the contribution of the 3D loss terms using the entire DTU evaluation set. Here, $\lambda_c$ is the loss weight for the connectivity term, $L_{conn}$, and $\lambda_s$ is the loss weight for the depth smoothness term $L_{smooth}$.  The best results are in boldface and the worst are underlined.}
\label{tab:ablation}
\resizebox{0.85\linewidth}{!}{%
\begin{tabular}{l|cccccc}
\hline
    Method &
    Acc.(mm) $\downarrow$ &
    Comp.(mm) $\downarrow$ &
    SSIM $\uparrow$ &
    PSNR $\uparrow$ &
    LPIPS $\downarrow$ &
    Primitives (K) $\downarrow$ \\ \hline\hline
    w/o $\Lagr_{smooth}$ \& $\Lagr_{conn}$ &
        0.66 &
        0.68 &
        0.912 &
        30.58 &
        0.227 &
        \textbf{218} \\
    w/o $\Lagr_{smooth}$ &
        0.65 &
        \underline{0.71} &
        \textbf{0.921} &
        \textbf{32.13} &
        \textbf{0.212} &
        244 \\
    w/o $\Lagr_{conn}$ &
        \underline{0.67} &
        0.69 &
        \underline{0.908} &
        \underline{30.15} &
        0.231 &
        224 \\
    full ($\lambda_c=10.0$, $\lambda_s=0.8$) &
        0.64 &
        0.70 &
        0.918 &
        31.68 &
        0.213 &
        249 \\ 
    full ($\lambda_c=300.0$, $\lambda_s=20.0$) &
        0.61 &
        0.63 &
        0.910 &
        30.87 &
        \underline{0.232} &
        244 \\
    full ($\lambda_c=1000.0$, $\lambda_s=50.0$) &
        \textbf{0.59} &
        \textbf{0.62} &
        0.909 &
        30.57 &
        \underline{0.232} &
        \underline{297} \\ \hline
    \hline
\end{tabular}
}
\end{table}

We show additional results on the mip-NeRF 360 dataset \citep{barron_2022_mip}. Following the protocol specified by \citet{barron_2022_mip}, we separate the images in each scene, taking every eighth image as a test image and training on the remaining. As standard evaluation, we report PSNR, SSIM, and LPIPS \citep{zhang_2018_unreasonable} metrics. We show results on all indoor and outdoor scenes in Table~\ref{tab:mipnerf360}, as well as per-scene results in Table~\ref{tab:mipnerf360_per_scene}. For indoor scenes, \network{} shows results on par with the state-of-the-art methods, having the leading LPIPS score, the fourth overall SSIM score, and a highly competitive PSNR score among all listed methods. On two outdoor scenes (treehill and flowers), \network{} is limited in reconstructing extremely distant background foliage, impacting the overall metrics (please see Fig.~\ref{fig:mipnerf_360_outdoor} for qualitative results on these scenes). In Fig.~\ref{fig:dtu_comp} (\textit{bottom}), we provide a comparison of novel view synthesis between \network{} and 2DGS. \network{} is able to render fine textures in low visibility regions in scenes as opposed to blurring with Gaussian representations.

\subsection{Ablations}
\label{sec:ablations}
We show an ablation evaluating the contributions of proposed loss terms in Table~\ref{tab:ablation}.
Removing the soft connectivity leads to a decrease in overall accuracy of the output models, while removing the depth supervision negatively affects the completeness. The two supervision signals complement each other, and we show that increasing the loss weights allows for tuning the framework for either better novel view synthesis or geometry. Additionally, the results corresponding to the changes in magnitude of the loss weights demonstrates the stability of \network{} to changes in hyper-parameters.

To evaluate geometry, all previous planar GS algorithms utilize a GT foreground-background segmentation mask when generating the final models. Using this mask to generate the final models for evaluation removes the effects of floaters and inaccurate estimation near the surfaces being evaluated. To portray a more grounded evaluation of geometry, we provide an ablation study in Table~\ref{tab:dtu_geo_ablation} in which we compute Chamfer distance of models without the use of any GT masks.

\section{Limitations \& Conclusions}
\label{sec:conclusion}

In this work, we introduce a new scene representation, namely Radiant Triangle Soup (RTS). To the best of our knowledge, we are the first to introduce explicit 3D forces between primitives in a splatting framework, helping to coordinate the positioning of primitives to directly form surfaces. Modifying the weights of these forces allows for tuning between 3D reconstruction quality and novel-view synthesis quality.
The main limitation of our current algorithm is its inability to extract watertight meshes. Furthermore, due to the periodic nearest-neighbors search, there is a minor increase in run-time proportional to the number of primitives. Please see Section~\ref{sec:sup_implementation} for more details.

The introduction of Triangle Soup as the underlying representation for Radiance Fields is amenable to future work in surface optimization. We plan to extend \network{} with modified primitive connectivity strategies and perform optimization over watertight meshes.

\clearpage
\newpage

\bibliographystyle{metafiles/iclr2026_conference.bst}
\bibliography{
    bib/macros,
    bib/refs
    }

\clearpage
\newpage

\setcounter{figure}{0}
\setcounter{table}{0}
\setcounter{section}{0}
\setcounter{equation}{0}
\renewcommand\thefigure{S.\arabic{figure}}  
\renewcommand\thetable{S.\arabic{table}}
\renewcommand\thesection{S.\arabic{section}}
\renewcommand\theequation{S.\arabic{equation}}

\section*{\Large Supplemental Material}

Here we include additional material on background geometry used for computing barycentric coordinates, further implementation details, and additional results.

\section{Background}
\label{sec:background}

In this section, we introduce the necessary background geometry used to parameterize a ray-triangle intersection in barycentric coordinates. The \textit{barycentric coordinates} of a point on a triangle $\lambda=[\lambda^0, \lambda^1, \lambda^2]^T$, s.t. $\sum_{j=1}^3 \lambda^j = 1$, provide a means for expressing the coordinates of the point as a linear combination of the coordinates of the three vertices, $[V^0, V^1, V^2]$. This parameterization is important for graphics applications to be able to efficiently rasterize triangles onto screen space and interpolate the color of a pixel from each vertex.

In rendering, we compute the ray-triangle intersections using,
\begin{equation}
    P = C + (\widehat{r}  d)
\end{equation}
where $C \in \mathbb{R}^3$ is the camera center, $\widehat{r} \in \mathbb{R}^3$ is the unit vector for the ray through pixel $i$, and $d \in \mathbb{R}$ is the depth along the ray from the camera center to the intersection point, computed as follows,

\begin{equation}
    d = \frac{\widehat{n} \cdot \overrightarrow{CB}}{\widehat{n} \cdot \widehat{r}}
\label{eq:depth}
\end{equation}
where $B \in \mathbb{R}^3$ is the barycenter of the triangle and $\widehat{n} \in \mathbb{R}^3$ is its normal.

We compute the barycentric coordinates for point $P$ using the triple products between the normal, a triangle edge, and the vector from each vertex to the point,
\begin{equation}
    \lambda = \frac{1}{\widehat{n} \cdot (\overrightarrow{V_0V_1} \times \overrightarrow{V_0V_2})} \begin{bmatrix} \widehat{n} \cdot (\overrightarrow{V_1V_2} \times \overrightarrow{V_1P}) \\ \widehat{n} \cdot (\overrightarrow{V_2V_0} \times \overrightarrow{V_2P}) \\ \widehat{n} \cdot (\overrightarrow{V_0V_1} \times \overrightarrow{V_0P}) \end{bmatrix}
\end{equation}

Intuitively, the contribution of each vertex is proportional to the area of the sub-triangle formed by the intersection point $P$ and the other two vertices of the triangle. This weight becomes larger as $P$ approaches the vertex.

\section{Implementation Details}
\label{sec:sup_implementation}

In this section, we describe our experimental setting and optimization parameters in detail. To begin optimization, similar to previous work \citep{huang_2024_2d, chen_2024_pgsr}, all geometric supervision is disabled, with optimization only being guided initially by the SSIM loss $\Lagr_{ssim}$. We enable the normal consistency loss $\Lagr_{norm}$ at iteration $7,000$ and enable both the smoothness loss $\Lagr_{smooth}$ and connectivity loss $\Lagr_{conn}$ at iteration $10,000$, both chosen empirically. All triangles start with an initial opacity ($\alpha$) value set to $0.1$, with opacity for all primitives being reset every $3,000$ iterations. We run optimization on the scenes from the DTU and Mip-NeRF 360 datasets for $25,000$ and $30,000$ iterations, respectively. Densification and pruning is run every $250$ iterations starting after iteration $2,000$. The maximum incenter gradient threshold for densification is set to $7.5 e^{-5}$ in all experiments.

The learning rates for each respective parameter are set as follows:
\begin{itemize}[leftmargin=*, itemsep=0pt, topsep=0pt]
    \item Spherical Harmonics ($\Delta$): $2.5e^{-3}$
    \item Opacity ($\alpha$): $5e^{-2}$
    \item Incenter ($\mu$): $\begin{bmatrix} 1.5e^{-4}, & 2e^{-6} \end{bmatrix}$
    \item Rotation ($R$): $1e^{-3}$
    \item Scale ($s$): $4e^{-3}$
    \item Diffuse Scalar ($\sigma$): $1e^{-3}$
\end{itemize}
where the incenter learning rate follows an exponential decay scheduler starting with $1.5e^{-4}$ and ending with $2e^{-6}$ conditioned on the total number of iterations. We also tested a linear decay scheduler leading to similar results.

\paragraph{Edge Connectivity Overhead}
To compute neighboring edge connections, we construct a single KD-Tree containing the midpoints of all triangle edges which is then queried for each edge once. The connected edge indices are the only structure that is stored during optimization. Since the parameters of the primitives are modified during optimization, we need to recompute the KD-Tree and connected edge indices every 250 iterations (aligned with Adaptive Density Control), which ultimately leads to a minimal overhead for moderate size scenes. To be concrete, building the KD-Tree and computing the neighboring indices takes on average 4 seconds for around 300,000 primitives, which is more than the average number of primitives needed to reconstruct the scenes for the DTU dataset. Since this operation only happens every 250 iterations starting after iteration 10,000 (when the connectivity loss is activated), the overhead of this operation only adds roughly 5 minutes to the optimization process, which is about a $6\%$ increase in runtime. Reconstructing larger scenes, or more precisely, scenes that require more primitives, will naturally demand a larger overhead.

\paragraph{Run-time}
The run-times for our approach are roughly $1.5$ hours on DTU and $4.5$ hours on Mip-NeRF 360, with scenes from DTU and scenes from Mip-NeRF 360 having on average $249,941$ and $1,275,985$ primitives, respectively, using $0.5$ resolution for the DTU dataset and $0.25$ resolution for the Mip-NeRF 360 dataset.

\newcommand\quartlwa{0.17\textwidth}
\begin{figure}[t]
\footnotesize
    \centering
\resizebox{\linewidth}{!}{%
    \begin{tabular}{ccccc}
        \includegraphics[height=\quartlwa]{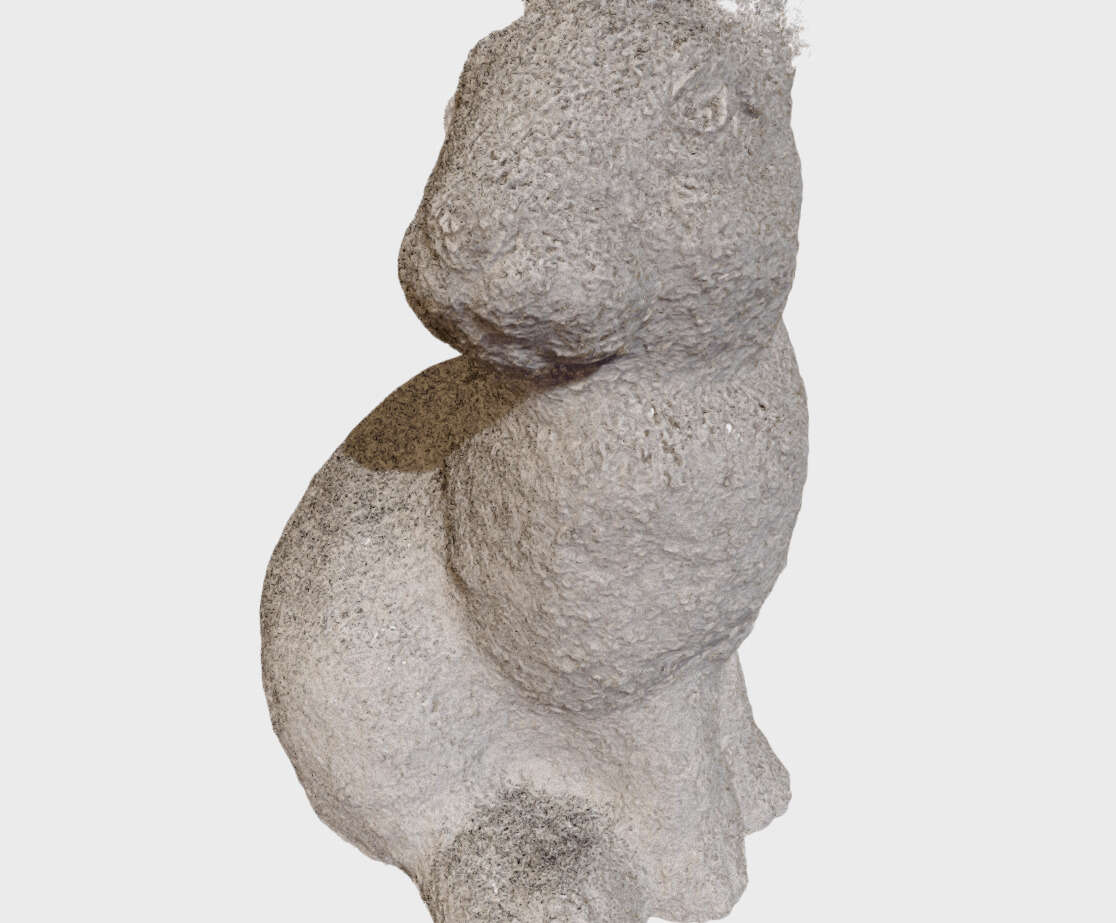} &
        \includegraphics[height=\quartlwa]{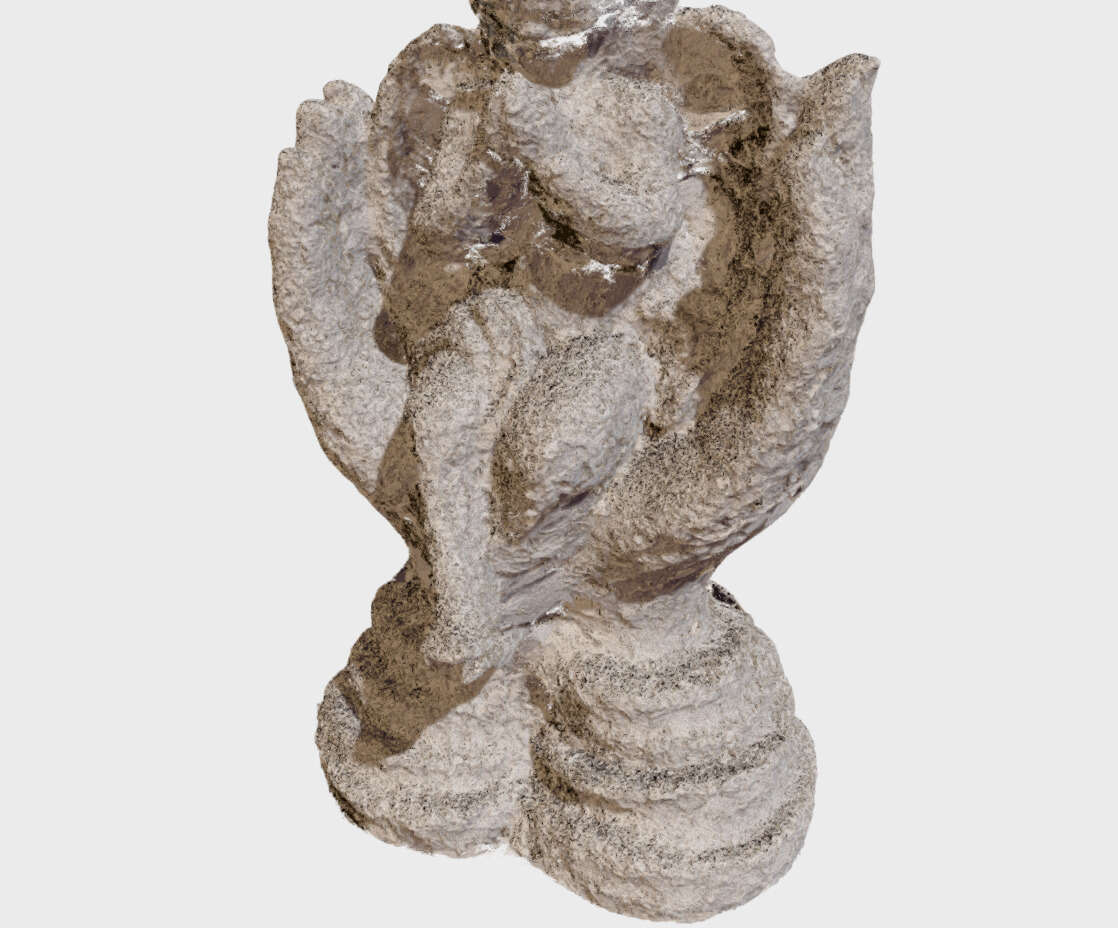} &
        \includegraphics[height=\quartlwa]{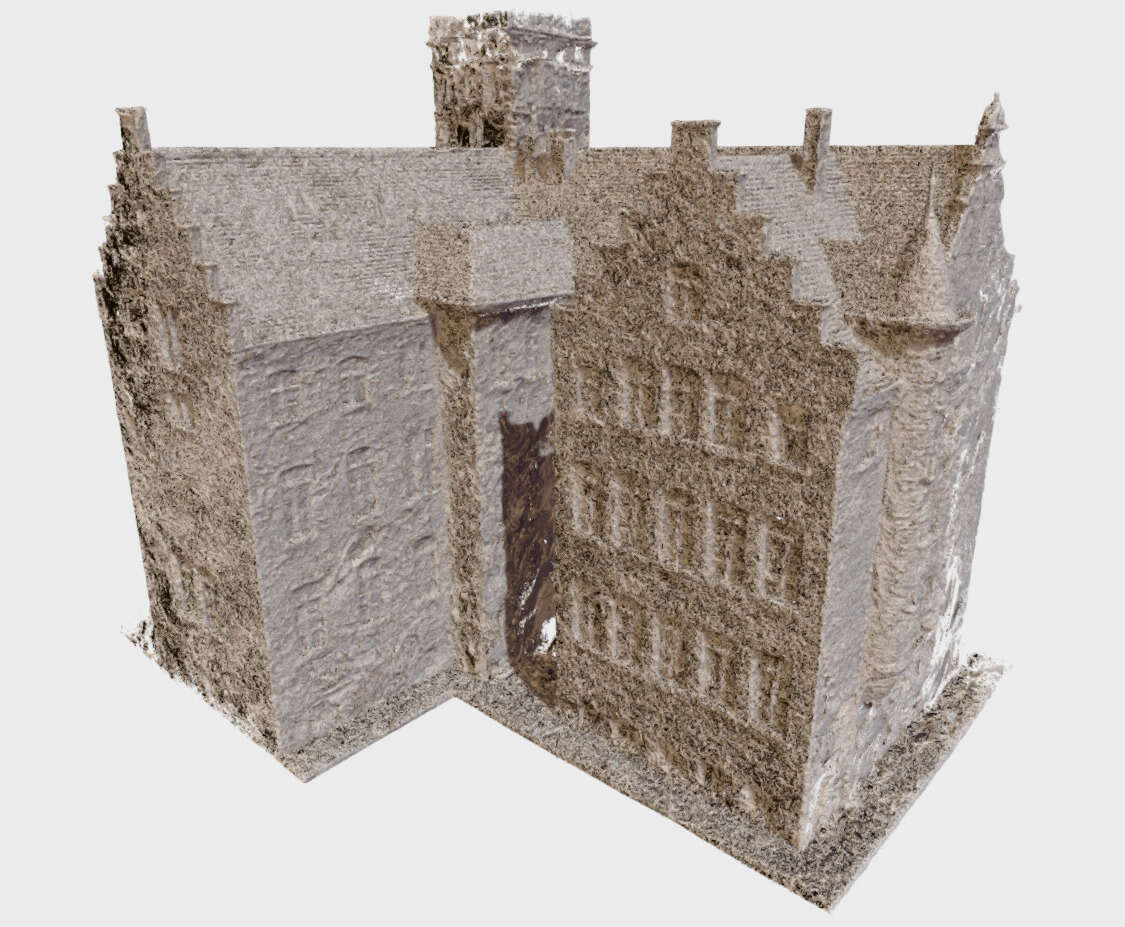} &
        \includegraphics[height=\quartlwa]{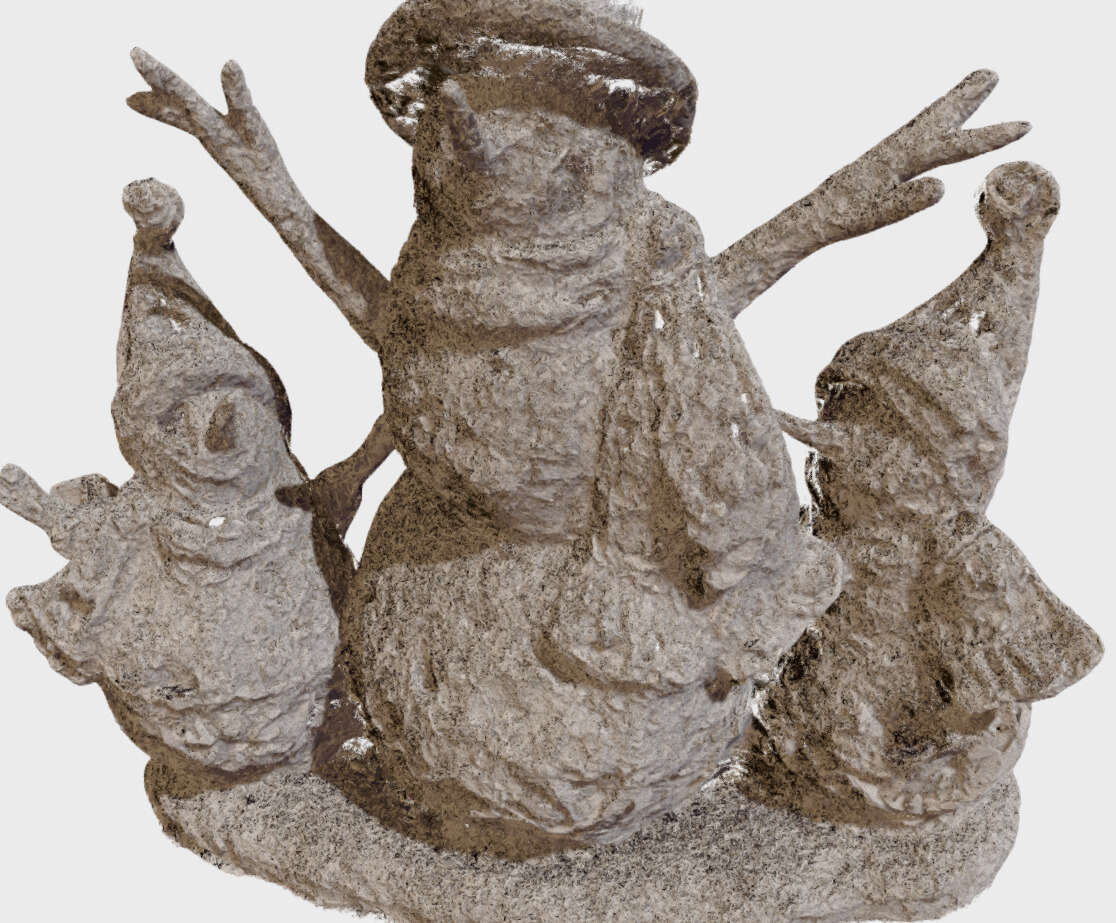} &
        \includegraphics[height=\quartlwa]{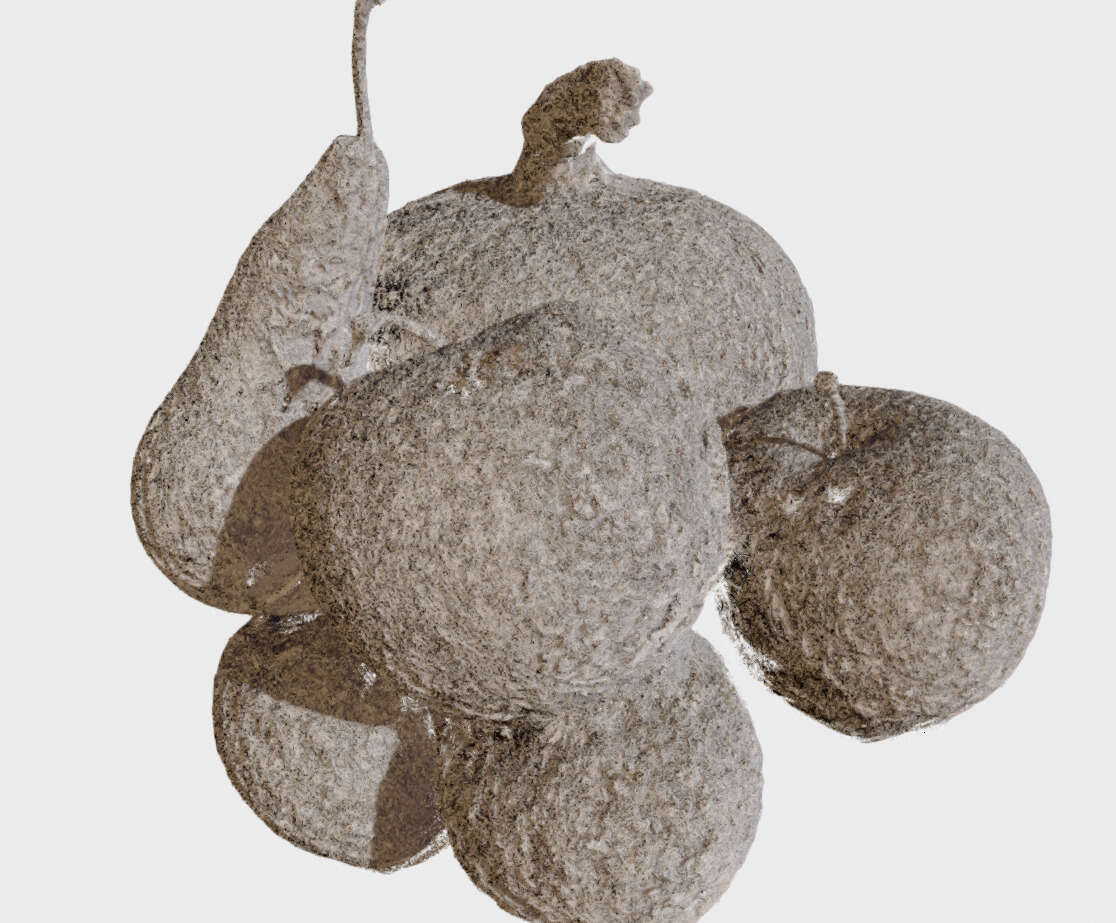} \\
        \includegraphics[height=\quartlwa]{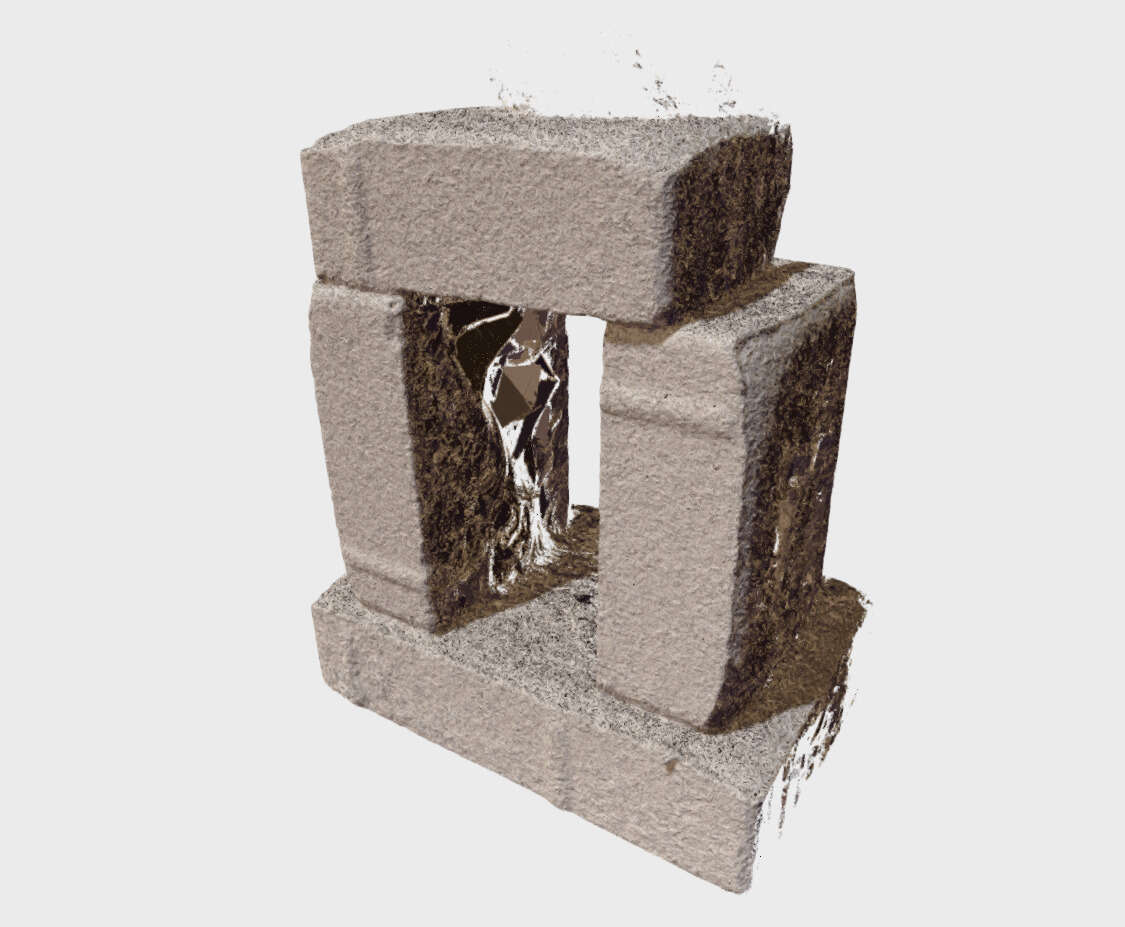} &
        \includegraphics[height=\quartlwa]{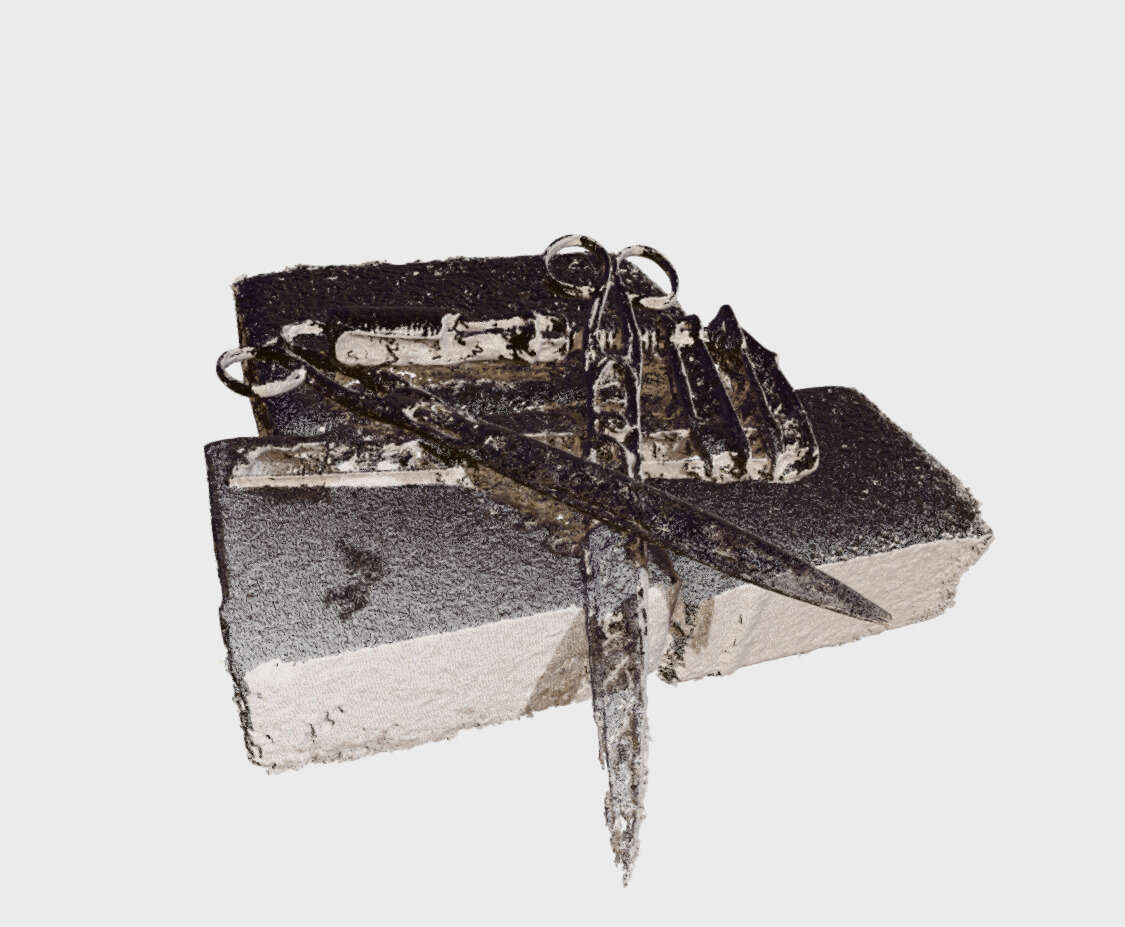} &
        \includegraphics[height=\quartlwa]{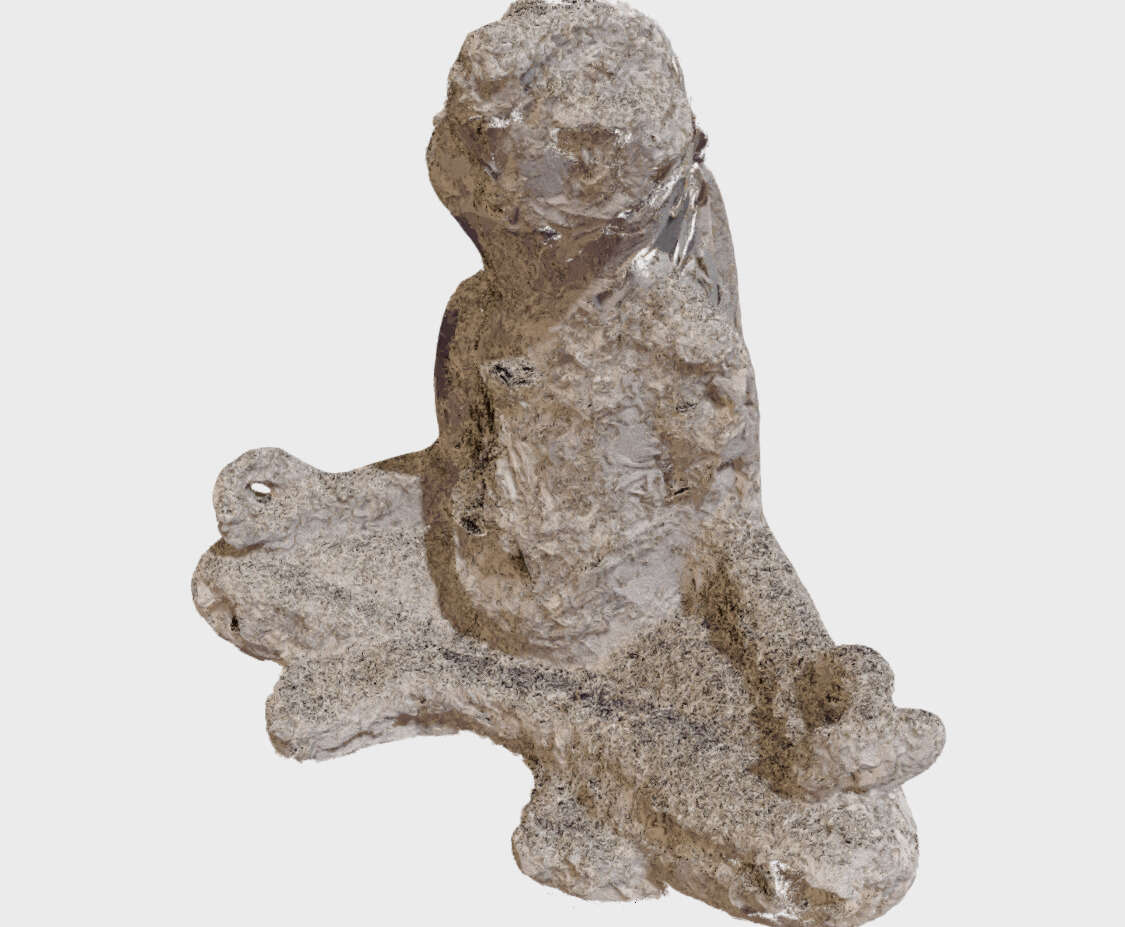} &
        \includegraphics[height=\quartlwa]{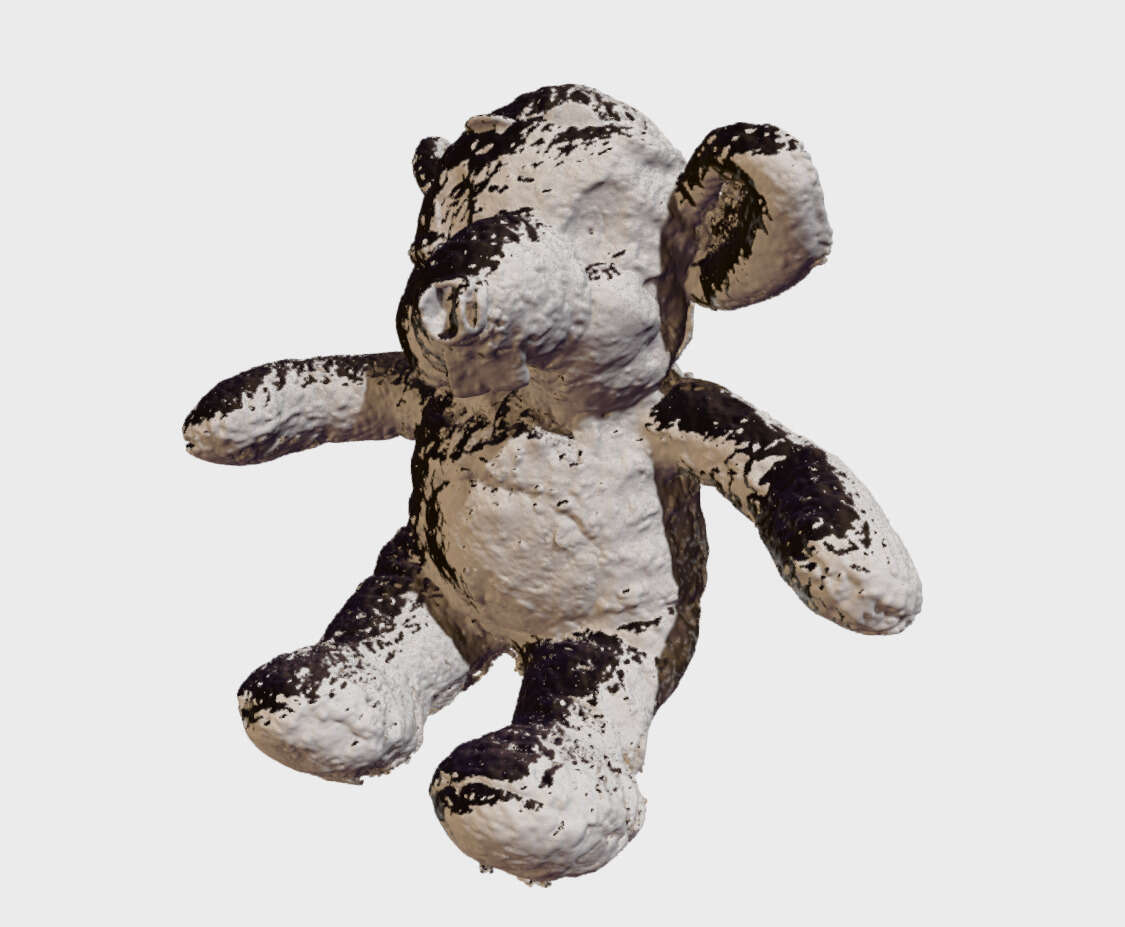} &
        \includegraphics[height=\quartlwa]{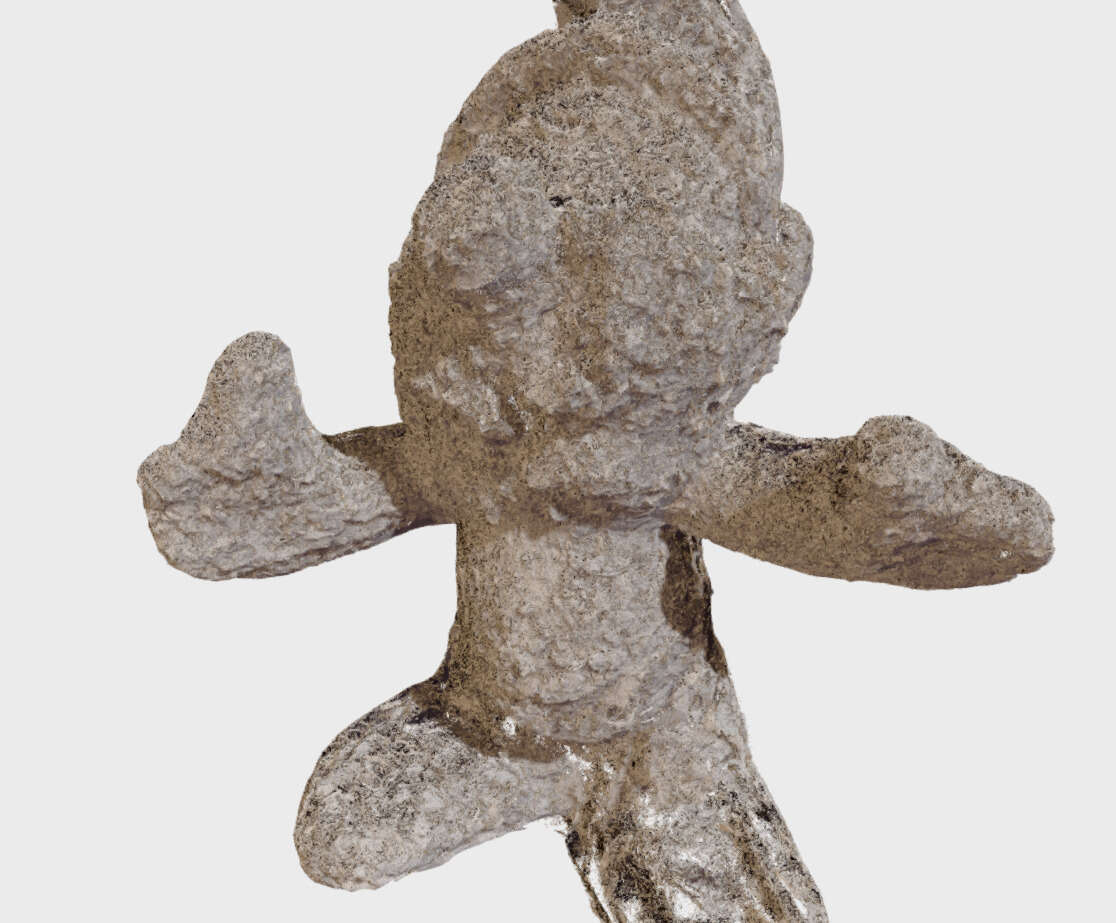} \\
        \includegraphics[height=\quartlwa]{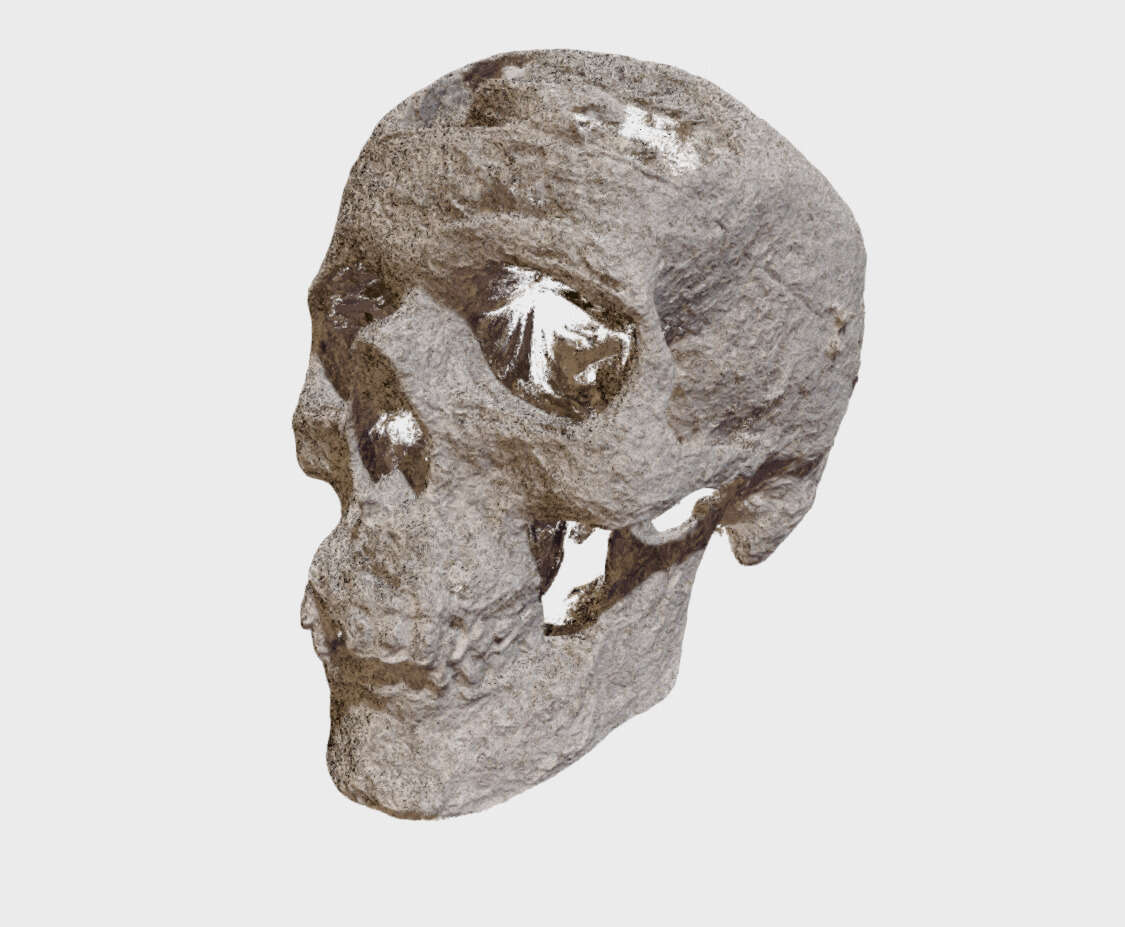} &
        \includegraphics[height=\quartlwa]{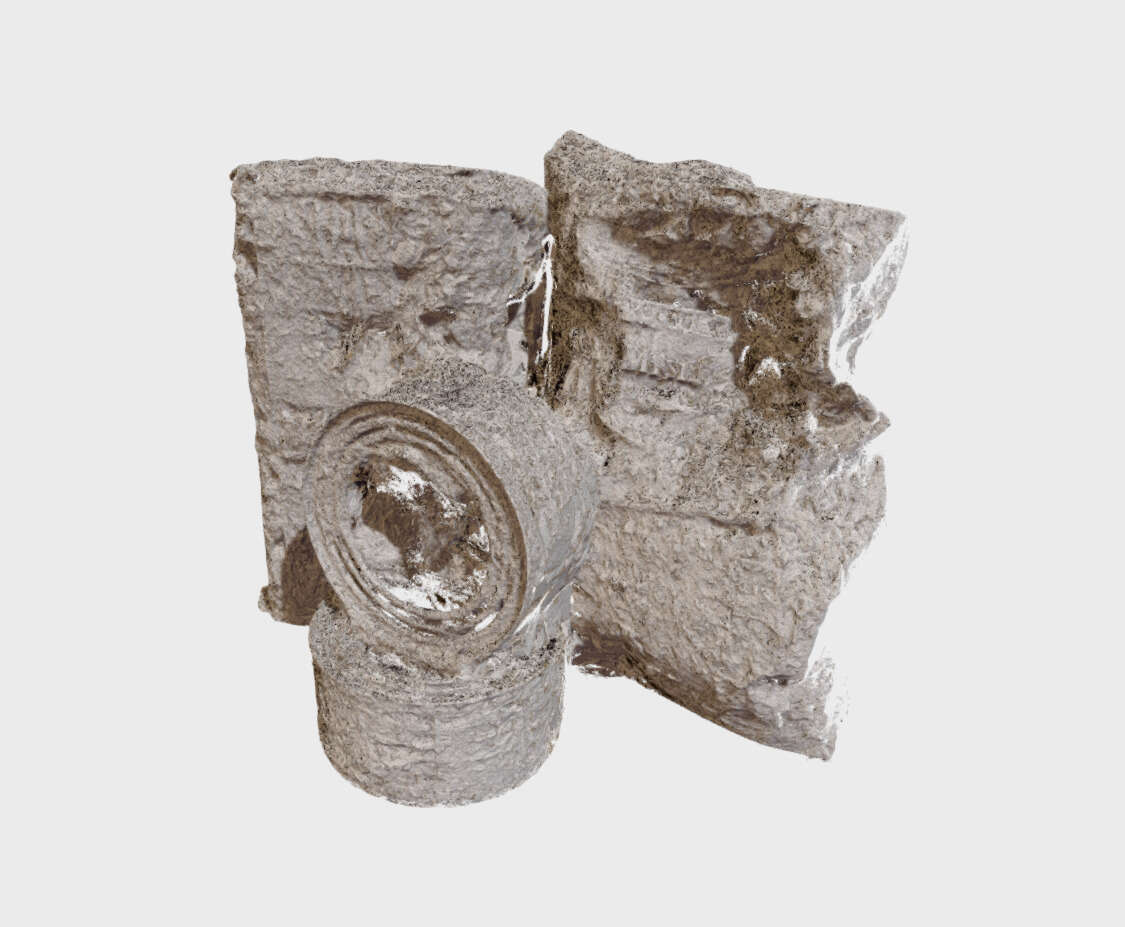} &
        \includegraphics[height=\quartlwa]{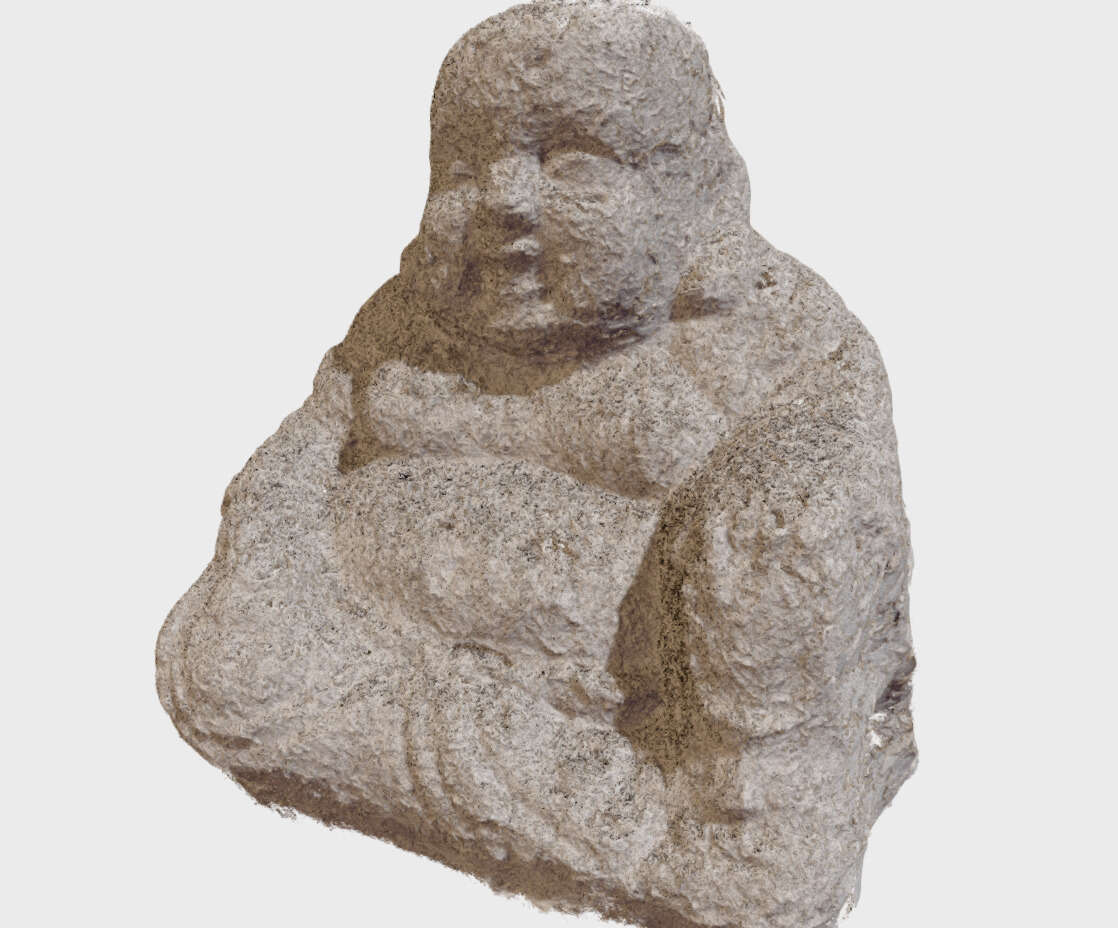} &
        \includegraphics[height=\quartlwa]{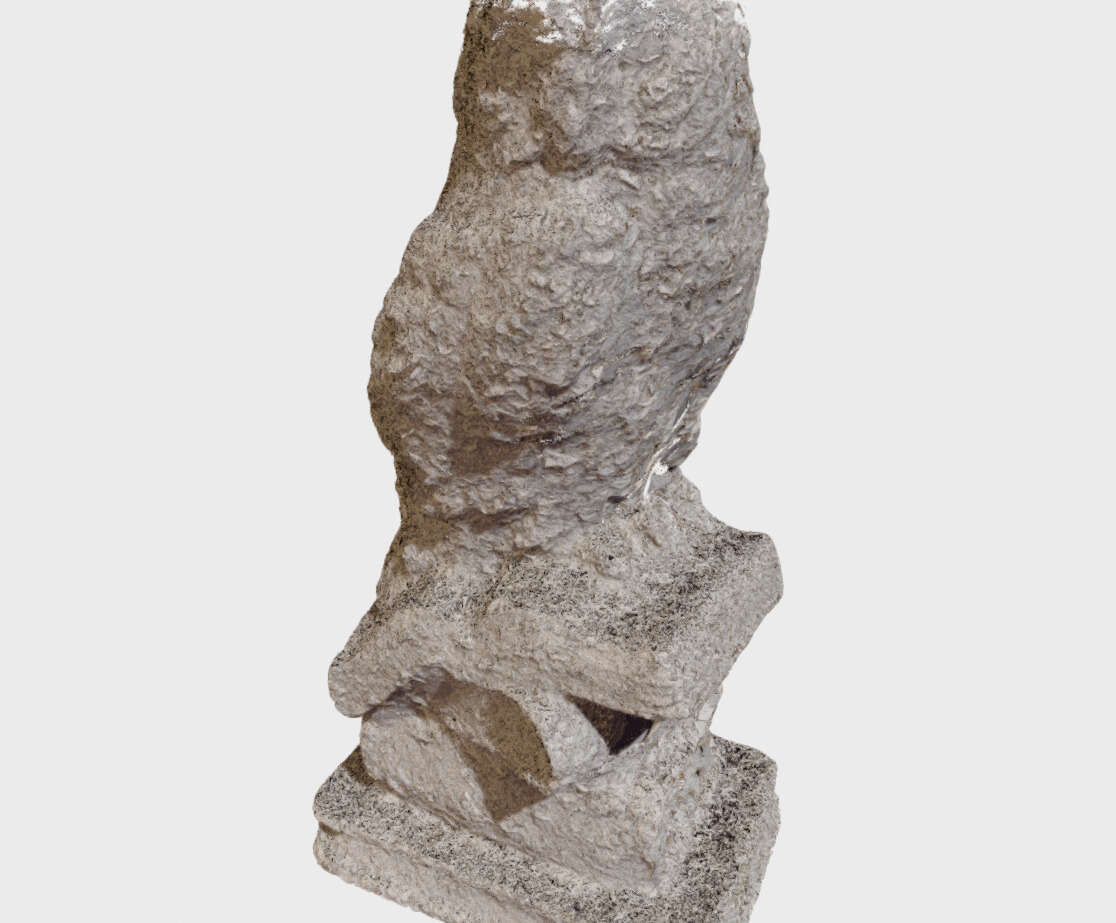} &
        \includegraphics[height=\quartlwa]{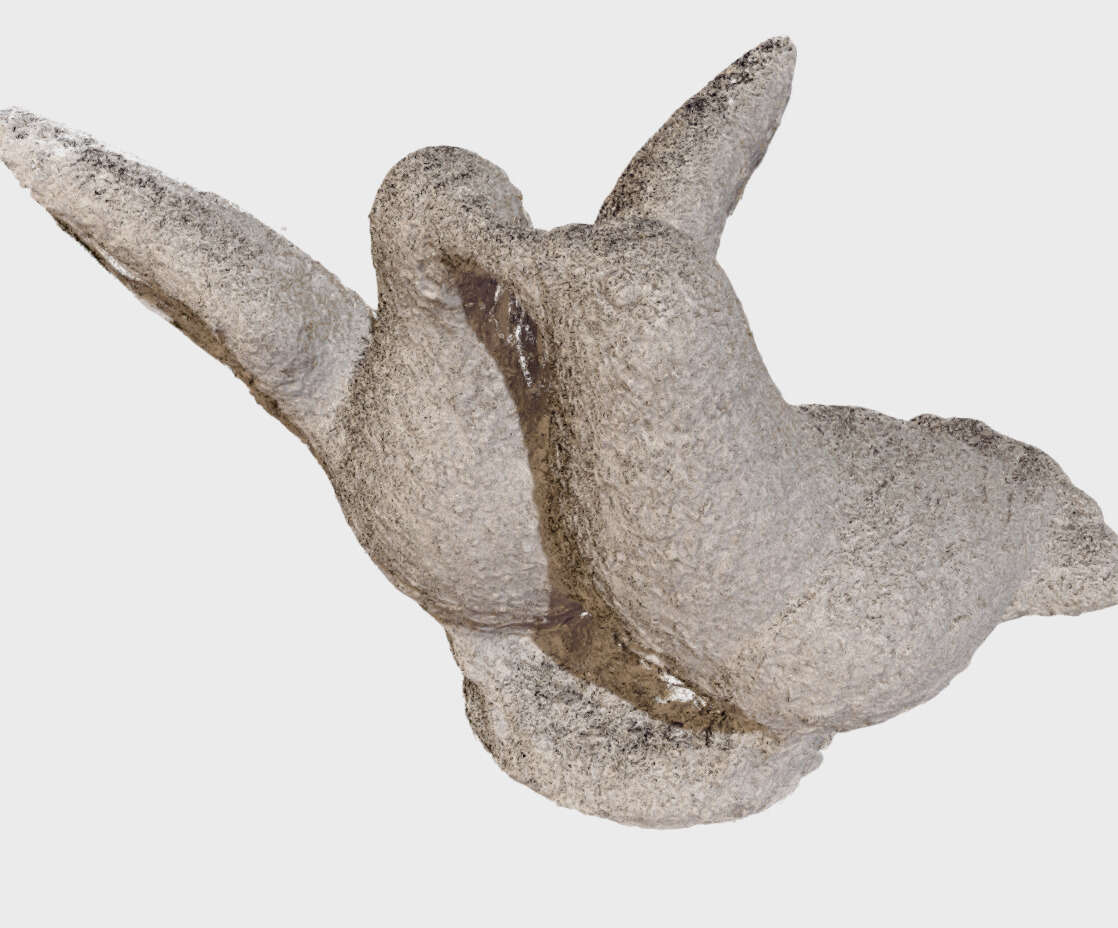} \\
    \end{tabular}
    }
    \caption{Point cloud visualization for all scenes from the evaluation set from the DTU dataset\citep{aanaes_2016_large}.}
    \label{fig:dtu_points}
\end{figure}

\section{Loss Functions}
\label{sec:sup_loss}
For completeness, we define the loss terms used in this paper that were introduced in previous work \citep{kerbl_2023_3d, huang_2024_2d}. The SSIM loss is computed as follows:
\begin{equation}
    \Lagr_{ssim} = \frac{1}{N} \sum_{i,j} (1 - \gamma) L_1 + \gamma L_{D\text{-}SSIM}
\end{equation}

The normal consistency loss is computed as follows:
\begin{equation}
    \Lagr_{norm} = \frac{1}{N} \sum_{i,j} (1 - \widehat{n}^T \widehat{n}_d)
\end{equation}
This encourages alignment between the rendered normal $\widehat{n}$ with the normal computed from the rendered depth map $\widehat{n}_d$,

\section{Additional Evaluations}
\label{sec:additional_eval}

\begin{table}[t]
\caption{Novel View Synthesis per-scene on all scenes from the mip-NeRF 360 dataset \citep{barron_2022_mip}.}
\label{tab:mipnerf360_per_scene}
\centering
\resizebox{0.35\linewidth}{!}{%
\begin{tabular}{l|ccc}
\hline
    Scene & SSIM & PSNR & LPIPS \\ \hline\hline
    room & 0.919 & 31.25 & 0.156 \\
    counter & 0.900 & 28.27 & 0.139 \\
    kitchen & 0.936 & 30.69 & 0.083 \\
    bonsai & 0.930 & 30.93 & 0.142 \\ \hline
    bicycle & 0.637 & 20.31 & 0.365 \\
    flowers & 0.617 & 19.99 & 0.351 \\
    garden & 0.742 & 24.89 & 0.250 \\
    stump & 0.658 & 23.25 & 0.410 \\
    treehill & 0.631 & 18.64 & 0.371 \\ \hline
\end{tabular}%
}
\end{table}

In Table~\ref{tab:mipnerf360_per_scene}, we show the per-scene novel view synthesis evaluation for \network{} on the mip-NeRF 360 dataset \citep{barron_2022_mip} using the standard metrics of PSNR, SSIM, and LPIPS. As \citet{vonluetzow_2025_linprim} note, using primitives with explicit boundaries can begin to introduce hard edges in regions with poor visibility while smoother primitives degrade more gracefully. While the diffuse boundaries of our triangles help prevent much of this behavior, its effects are noticeable in some of the reconstructed images, especially in outdoor scenes with distant background foliage. See Fig.~\ref{fig:mipnerf_360_outdoor}

As mentioned in Section~\ref{sec:ablations}, we compare the Chamfer distances of \network{} and two competitive GS algorithms in Table~\ref{tab:dtu_geo_ablation} with and without the use of the GT segmentation masks. While the distances increase for all methods, this experiment demonstrates how \network{} is more effective at floater removal and background modeling.

We provide additional renderings of novel views on challenging indoor and outdoor scenes with fine details and non-Lambertian surfaces. We show qualitative geometric reconstruction results on all scenes from the DTU dataset in Fig.~\ref{fig:dtu_points} and novel view synthesis results on all indoor and outdoor scenes from the Mip-NeRF 360 datasets in Fig.~\ref{fig:mipnerf_360_indoor} and Fig.~\ref{fig:mipnerf_360_outdoor}, respectively.

\begin{table}[t]
\caption{Chamfer distance evaluation on scenes from DTU \citep{aanaes_2016_large}. Following previous literature, we average the accuracy and completeness Chamfer distances on the widely used evaluation set. Chamfer distances are measured in millimeters. \textbf{Top}: Chamfer distance evaluation using the GT mask. \textbf{Bottom}: Chamfer distance evaluation \textit{without} using the GT mask.}
\label{tab:dtu_geo_ablation}
\centering
\resizebox{\linewidth}{!}{%
\begin{tabular}{l|ccccccccccccccc|c}
\hline
    \multirow{2}{*}{Method} &
    \multicolumn{16}{c}{DTU} \\ \cline{2-17}
        &
        24 &
        37 &
        40 &
        55 &
        63 &
        65 &
        69 &
        83 &
        97 &
        105 &
        106 &
        110 &
        114 &
        118 &
        122 &
        Mean (mm)$\downarrow$ \\ \hline\hline
    2DGS \citep{huang_2024_2d} &
        0.48 &
        0.91 &
        0.39 &
        0.39 &
        1.01 &
        0.83 &
        0.81 &
        1.36 &
        1.27 &
        0.76 &
        0.70 &
        1.40 &
        0.40 &
        0.76 &
        0.52 &
        0.80 \\ 
    PGSR \citep{chen_2024_pgsr} &
        \textbf{0.36} &
        \textbf{0.57} &
        0.38 &
        \textbf{0.33} &
        {0.78} &
        \textbf{0.58}&
        \textbf{0.50} &
        1.08 &
        \textbf{0.63} &
        0.59 &
        \textbf{0.46} &
        \textbf{0.54} &
        \textbf{0.30} &
        \textbf{0.38} &
        \textbf{0.34} &
        \textbf{0.52} \\
    \textbf{\network{}} &
        0.42 &
        0.61 &
        0.74 &
        0.39 &
        \textbf{0.53} &
        0.86 &
        0.70 &
        \textbf{0.84} &
        0.72 &
        \textbf{0.37} &
        0.68 &
        0.87 &
        0.34 &
        0.58 &
        0.44 &
        0.61 \\ \hline
    2DGS \citep{huang_2024_2d} [no mask] &
         1.22 &
         1.69 &
         \textbf{0.88} &
         \textbf{0.43} &
         0.96 &
         0.77 &
         0.85 &
         1.23 &
         1.87 &
         1.69 &
         0.91 &
         2.01 &
         0.82 &
         0.82 &
         1.07 &
         1.15 \\ 
    PGSR \citep{chen_2024_pgsr} [no mask] &
         1.10 &
         1.59 &
         0.97 &
         0.45 &
         1.98 &
         \textbf{0.64} &
         \textbf{0.59} &
         1.70 &
         1.69 &
         1.57 &
         \textbf{0.74} &
         \textbf{0.63} &
         0.44 &
         \textbf{0.71} &
         \textbf{0.74} &
         1.04 \\
    \textbf{\network{}} [no mask] &
         \textbf{0.52} &
         \textbf{0.92} &
         0.94 &
         0.55 &
         \textbf{0.91} &
         1.01 &
         0.74 &
         \textbf{1.09} &
         \textbf{1.05} &
         \textbf{0.67} &
         0.88 &
         1.07 &
         \textbf{0.42} &
         0.79 &
         0.80 &
         \textbf{0.82} \\ \hline
\end{tabular}%
}
\end{table}

\newcommand\heightmipindoor{0.20\textwidth}
\begin{figure}[t]
\footnotesize
    \centering
\resizebox{\linewidth}{!}{%
    \begin{tabular}{ccc}
        \includegraphics[height=\heightmipindoor]{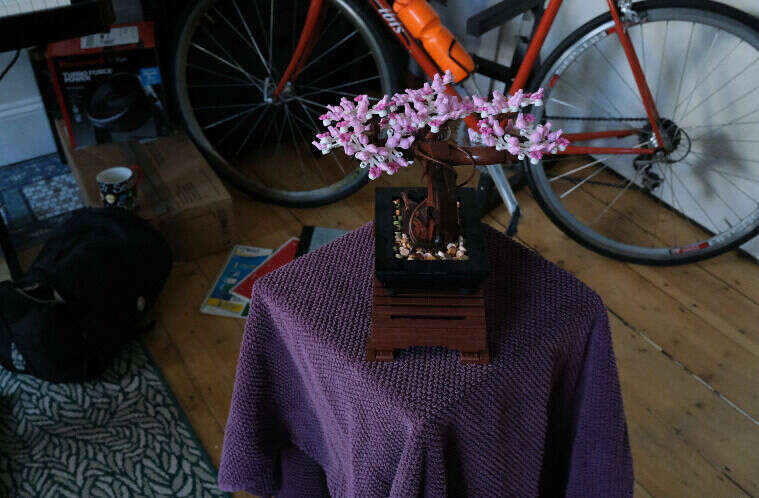} &
        \includegraphics[height=\heightmipindoor]{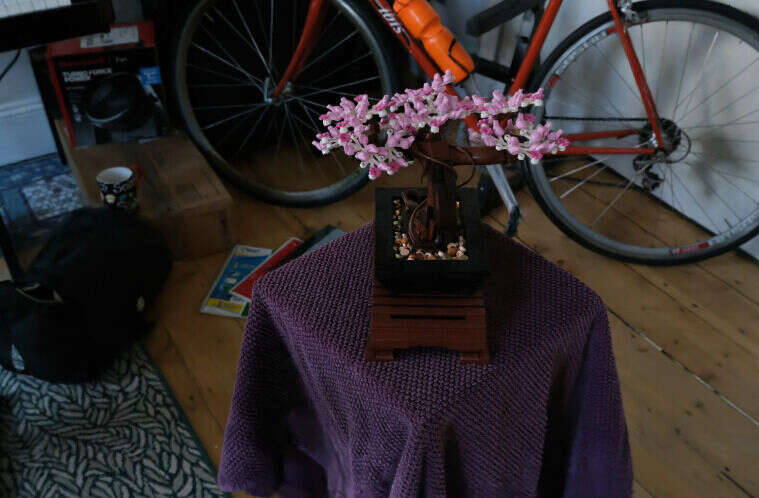} &
        \includegraphics[height=\heightmipindoor]{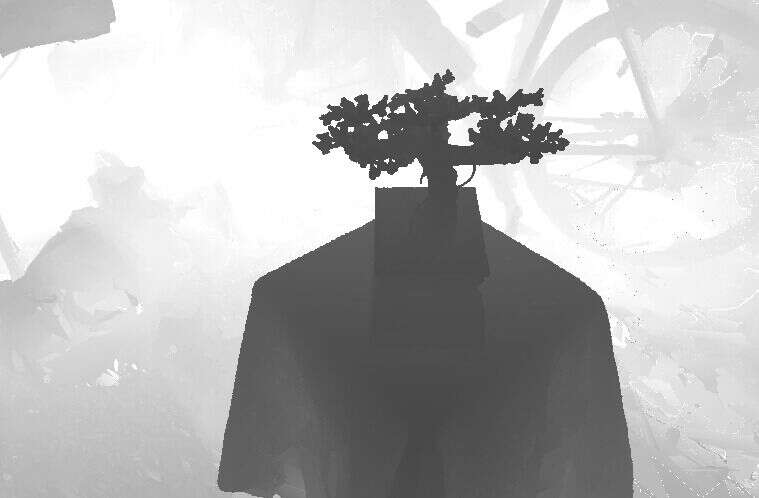} \\
        \includegraphics[height=\heightmipindoor]{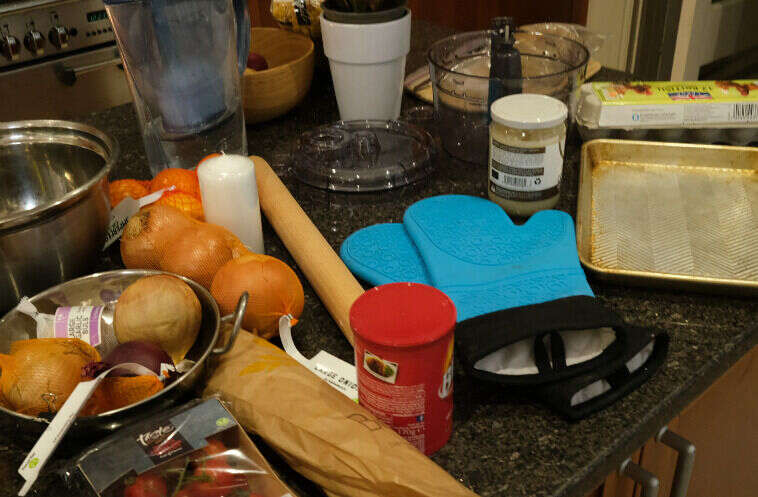} &
        \includegraphics[height=\heightmipindoor]{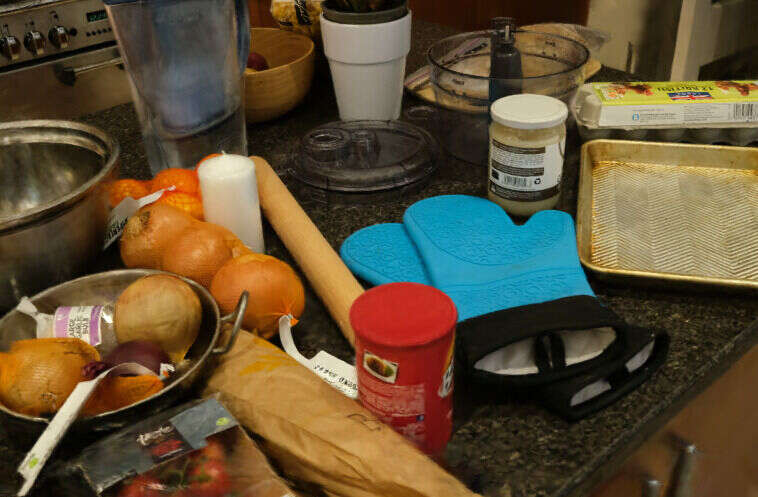} &
        \includegraphics[height=\heightmipindoor]{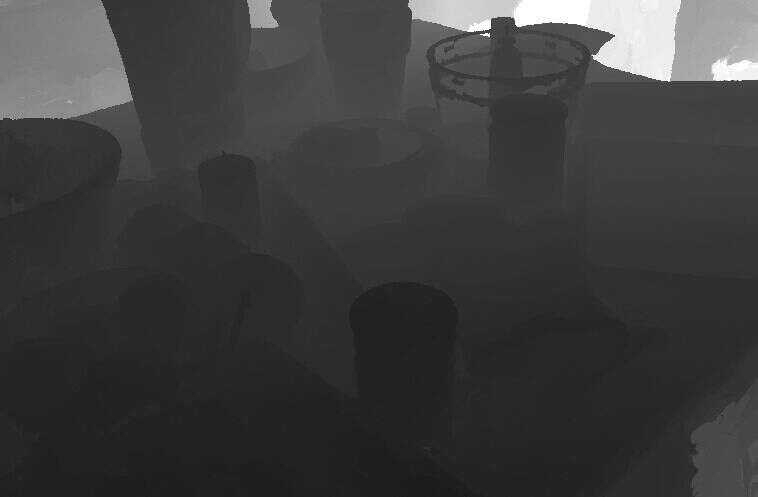} \\
        \includegraphics[height=\heightmipindoor]{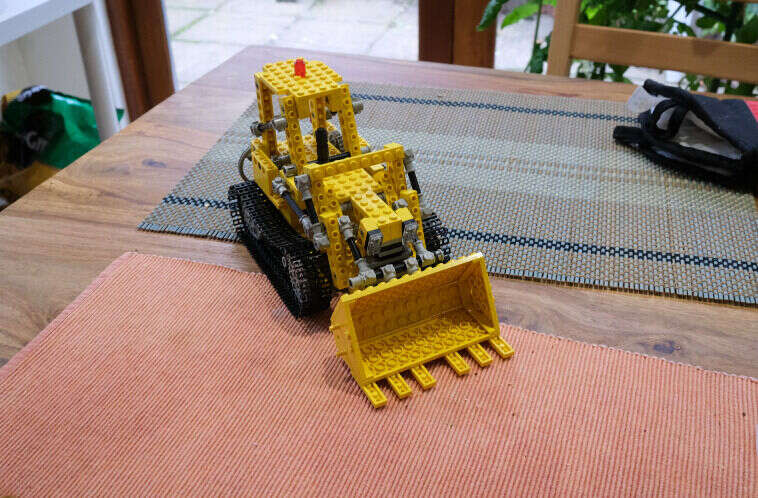} &
        \includegraphics[height=\heightmipindoor]{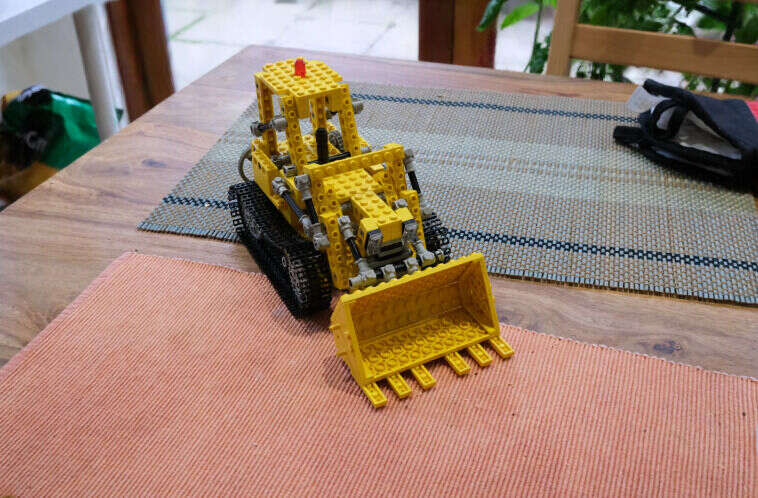} &
        \includegraphics[height=\heightmipindoor]{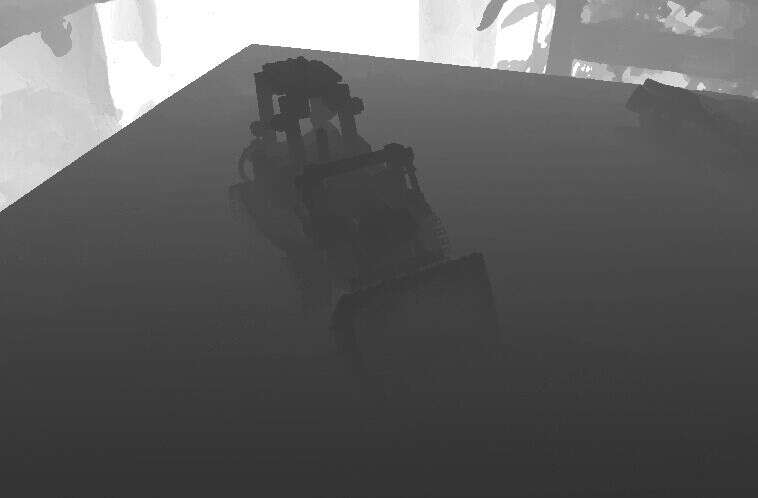} \\
        \includegraphics[height=\heightmipindoor]{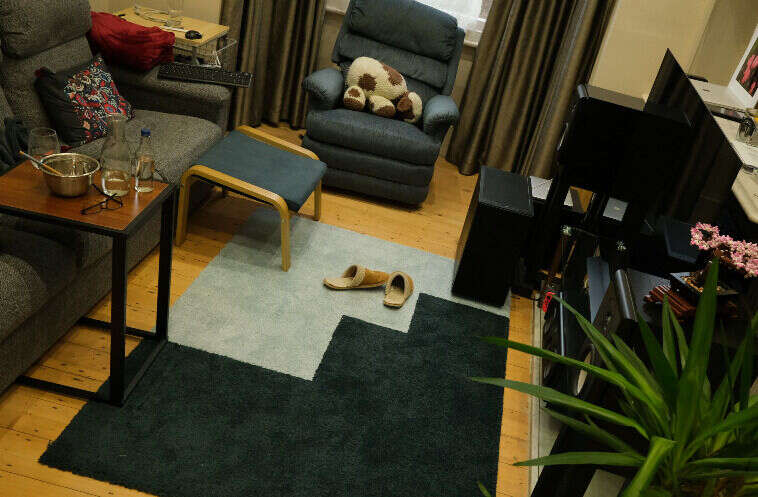} &
        \includegraphics[height=\heightmipindoor]{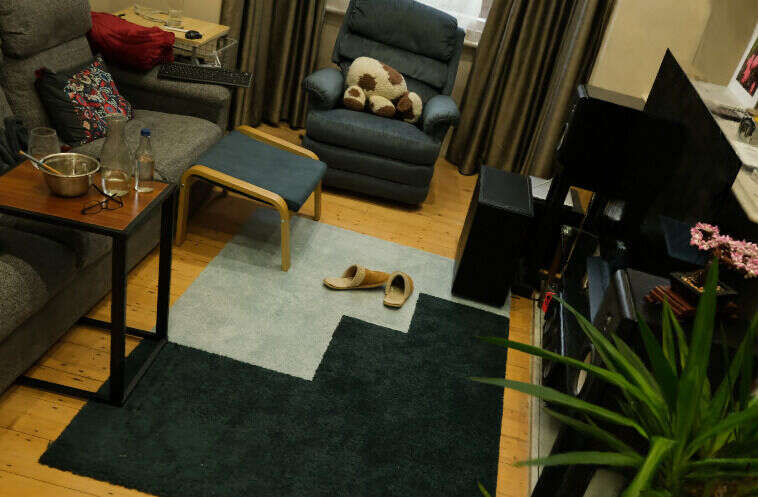} &
        \includegraphics[height=\heightmipindoor]{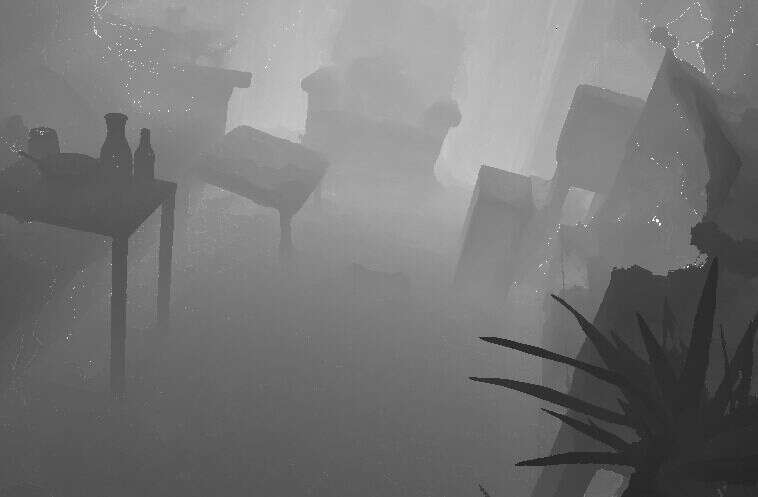} \\        
        (a) GT Image & (b) Rendered Image & (c) Rendered Depth \\
    \end{tabular}
    }
    \caption{Visualizations of all indoor scenes, \textit{(top-down) [bonsai, counter, kitchen, room]}, from the mip-NeRF 360 dataset \citep{barron_2022_mip}.}
    \label{fig:mipnerf_360_indoor}
\end{figure}

\newcommand\heightmipoutdoor{0.20\textwidth}
\begin{figure}[t]
\footnotesize
    \centering
\resizebox{\linewidth}{!}{%
    \begin{tabular}{ccc}
        \includegraphics[height=\heightmipoutdoor]{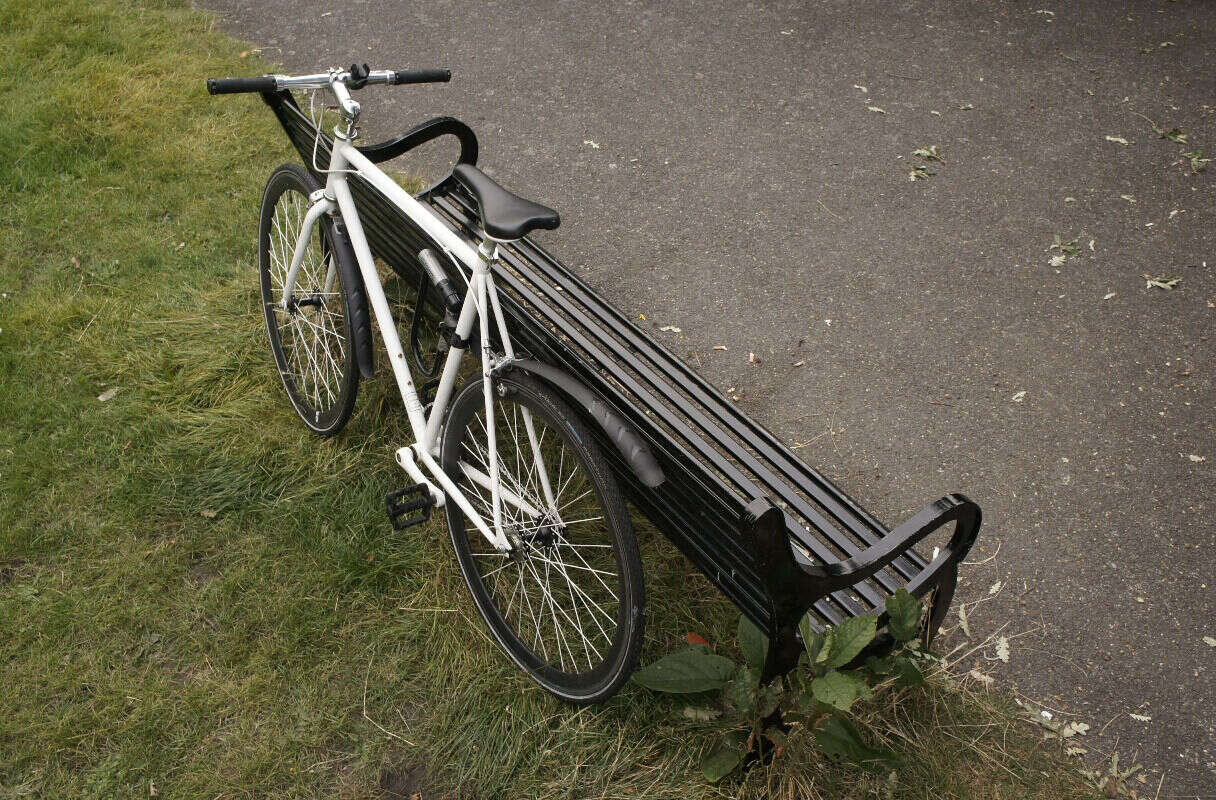} &
        \includegraphics[height=\heightmipoutdoor]{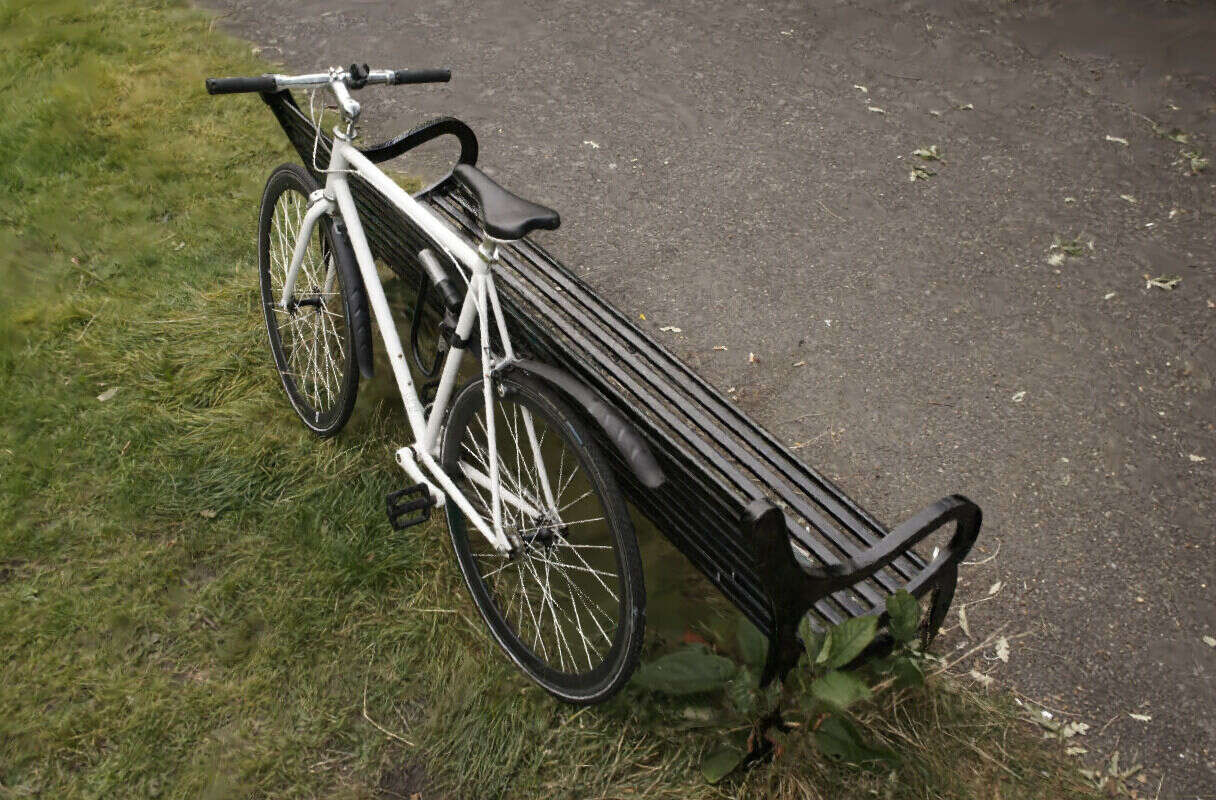} &
        \includegraphics[height=\heightmipoutdoor]{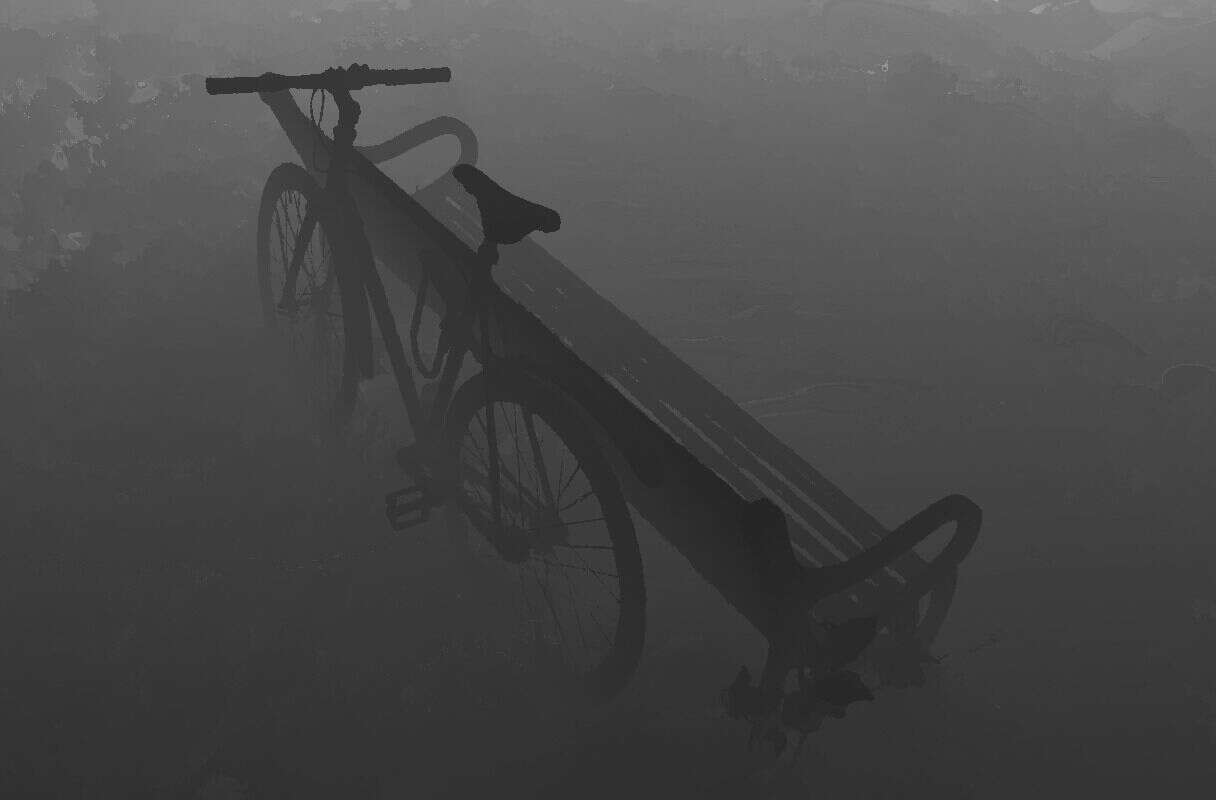} \\
        \includegraphics[height=\heightmipoutdoor]{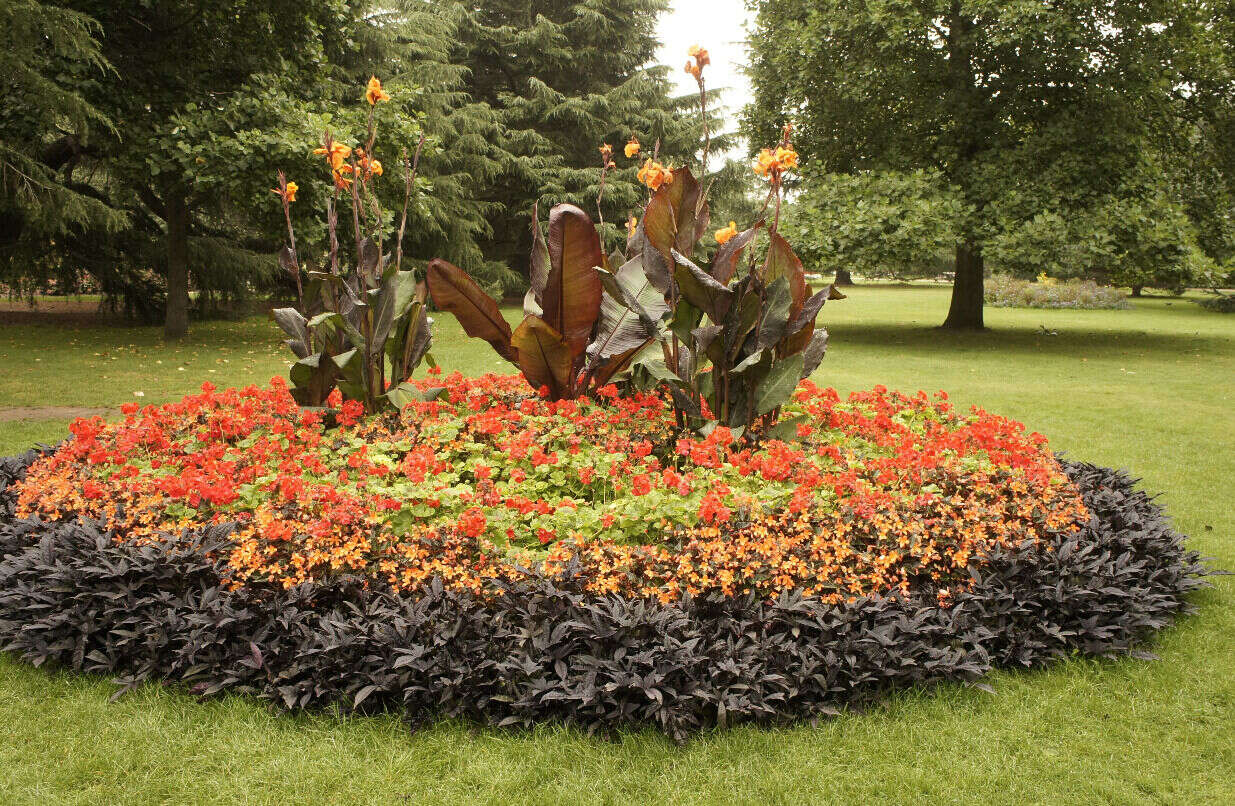} &
        \includegraphics[height=\heightmipoutdoor]{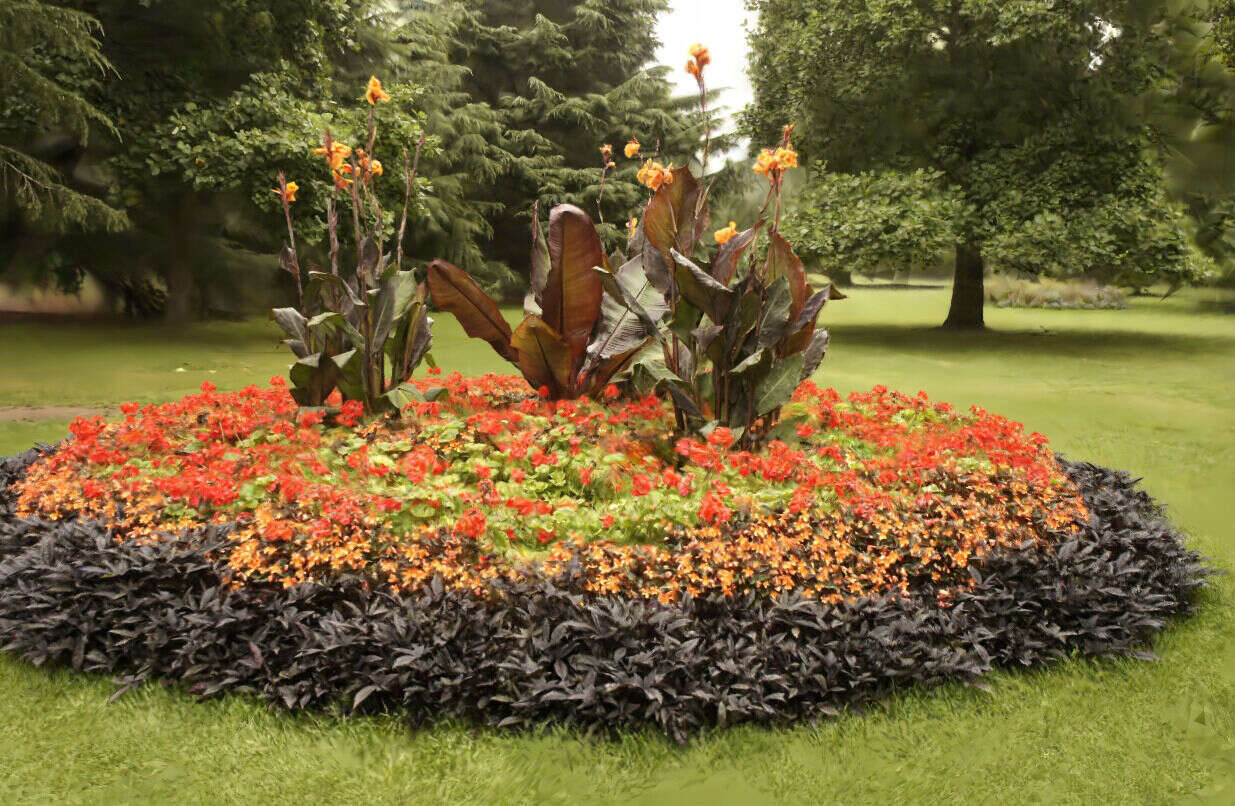} &
        \includegraphics[height=\heightmipoutdoor]{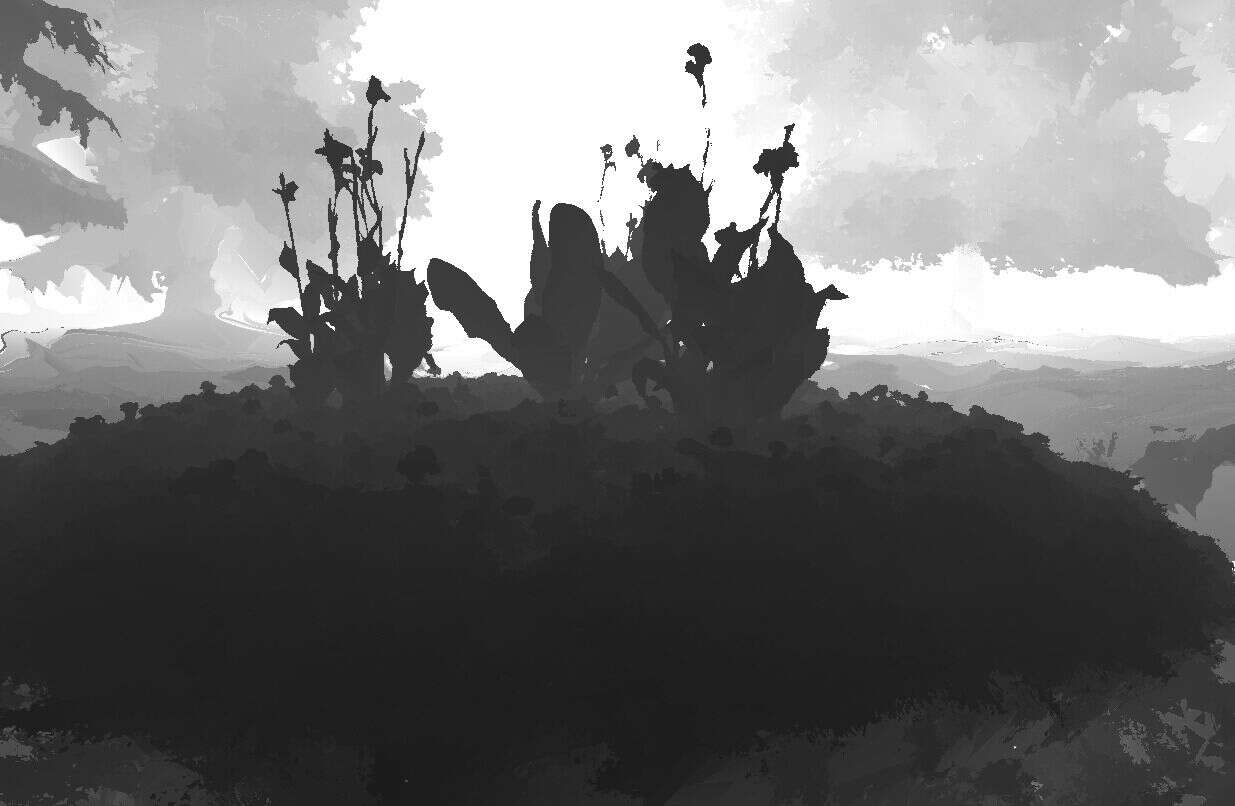} \\
        \includegraphics[height=\heightmipoutdoor]{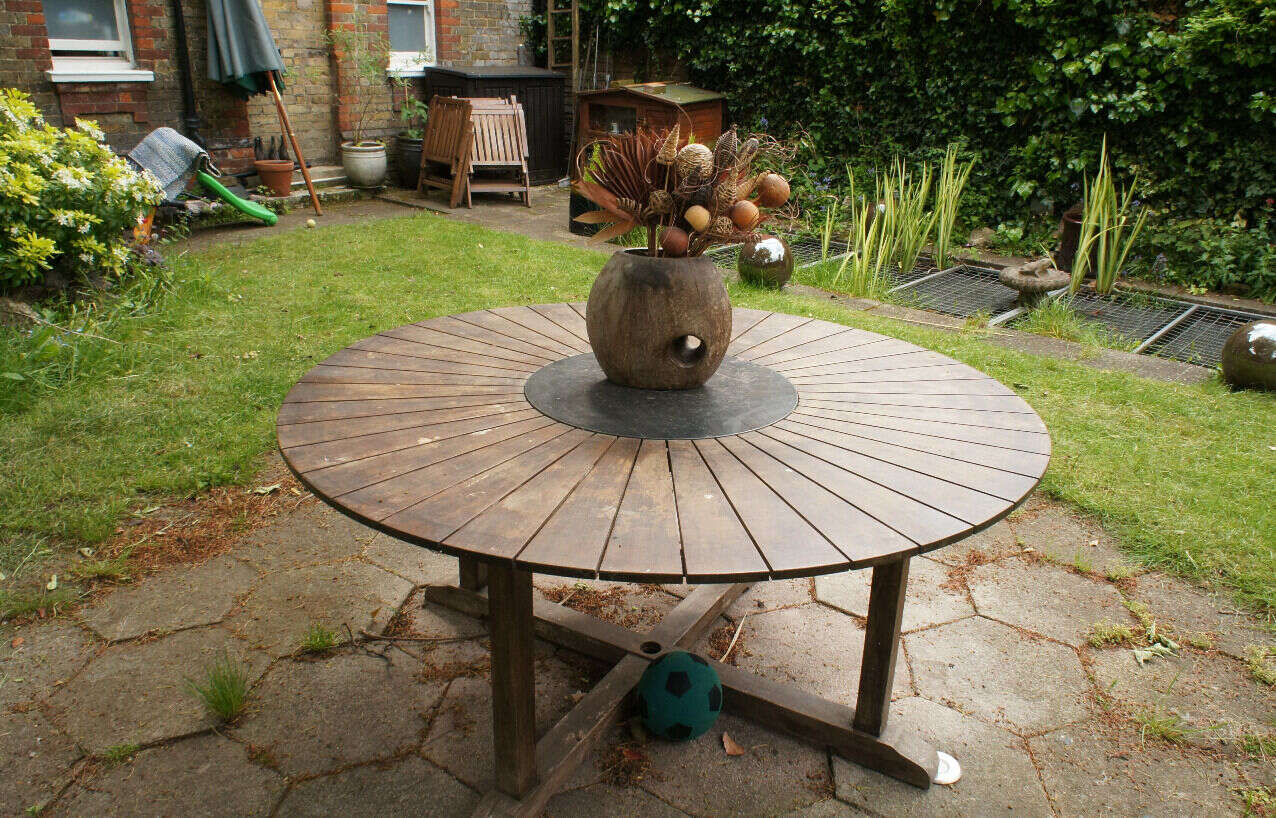} &
        \includegraphics[height=\heightmipoutdoor]{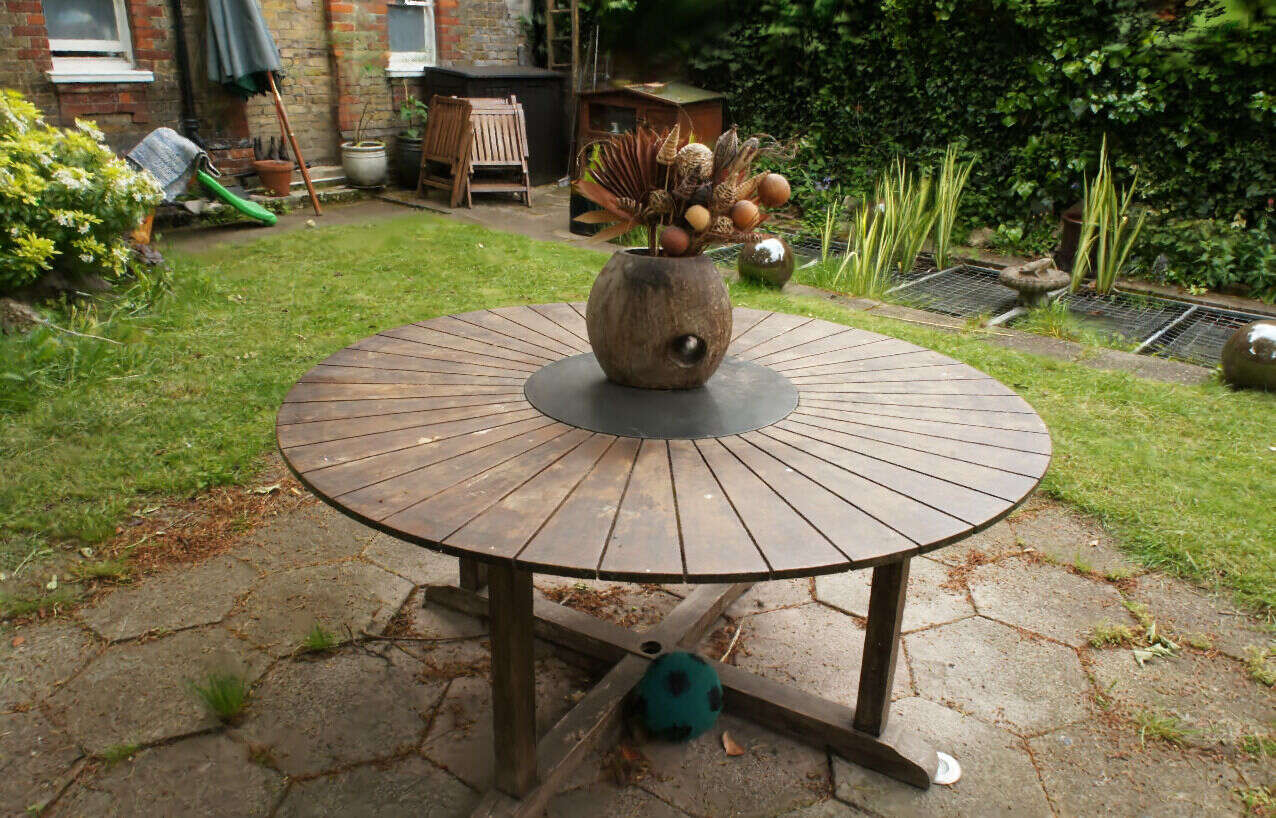} &
        \includegraphics[height=\heightmipoutdoor]{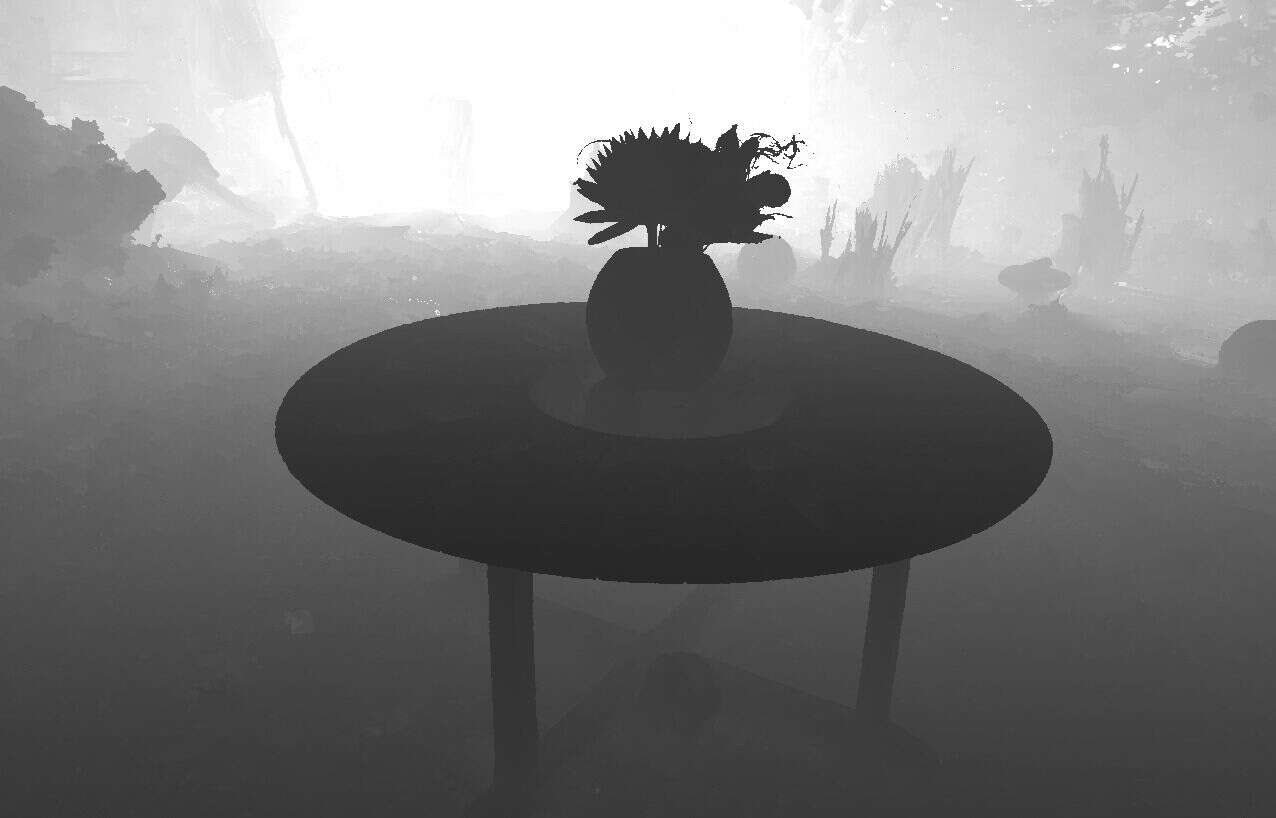} \\
        \includegraphics[height=\heightmipoutdoor]{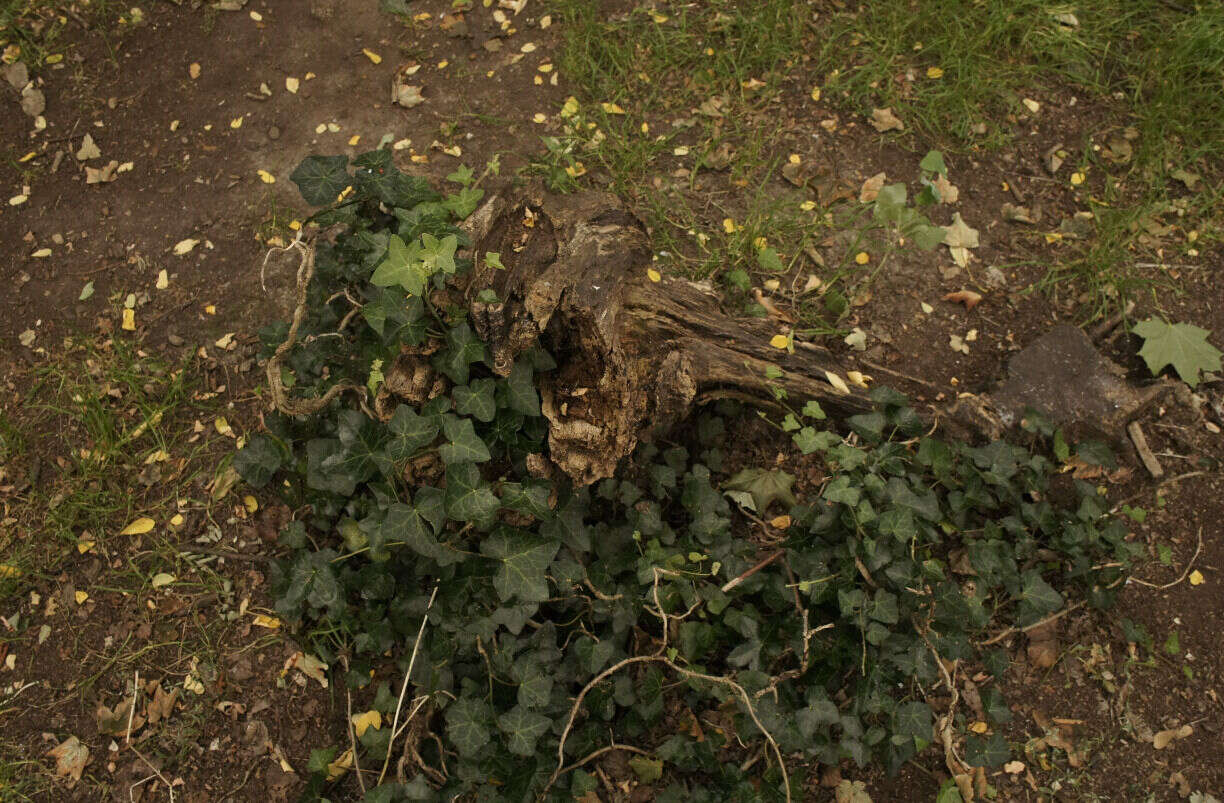} &
        \includegraphics[height=\heightmipoutdoor]{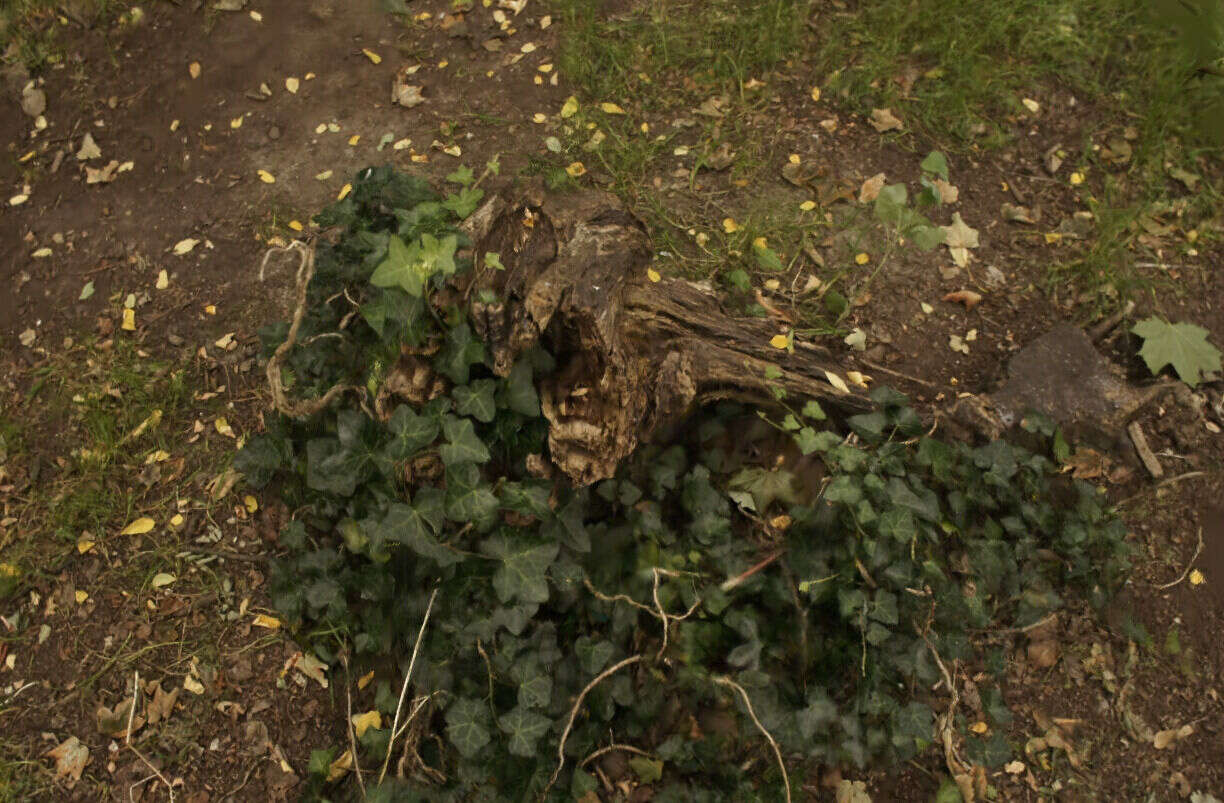} &
        \includegraphics[height=\heightmipoutdoor]{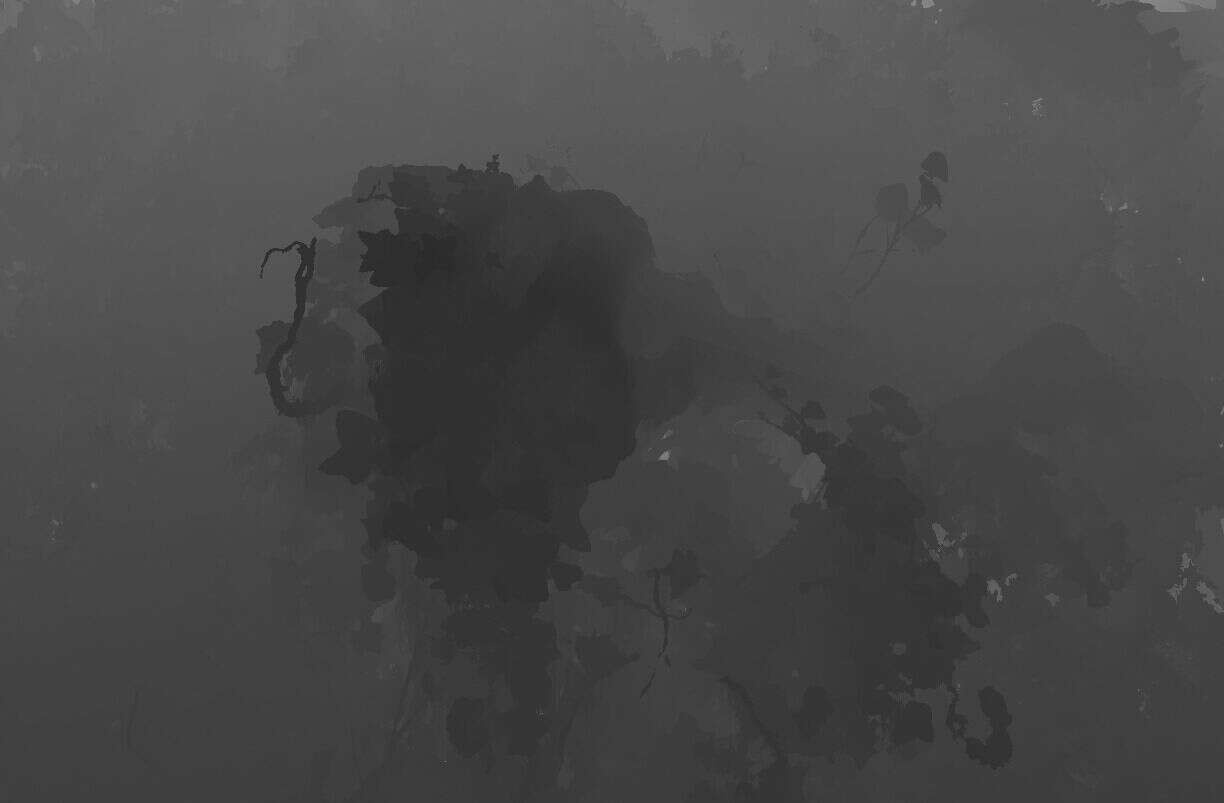} \\
        \includegraphics[height=\heightmipoutdoor]{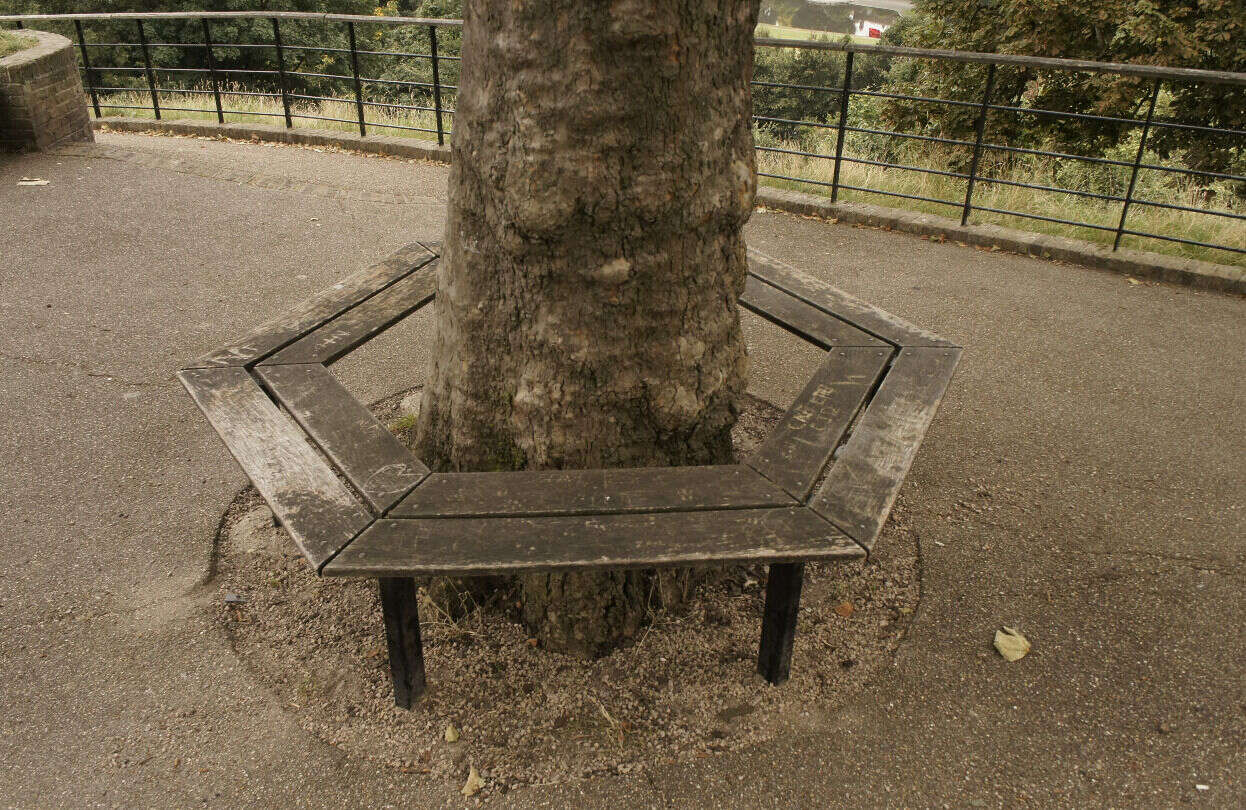} &
        \includegraphics[height=\heightmipoutdoor]{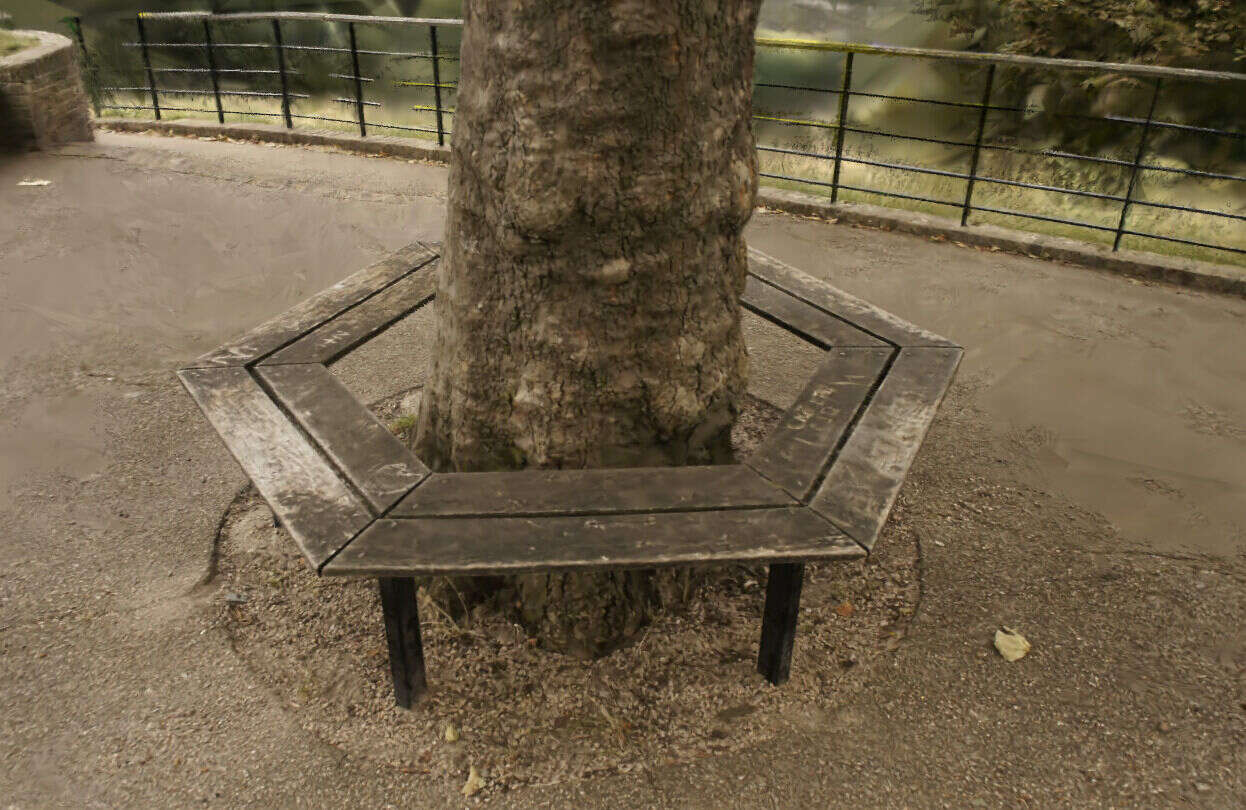} &
        \includegraphics[height=\heightmipoutdoor]{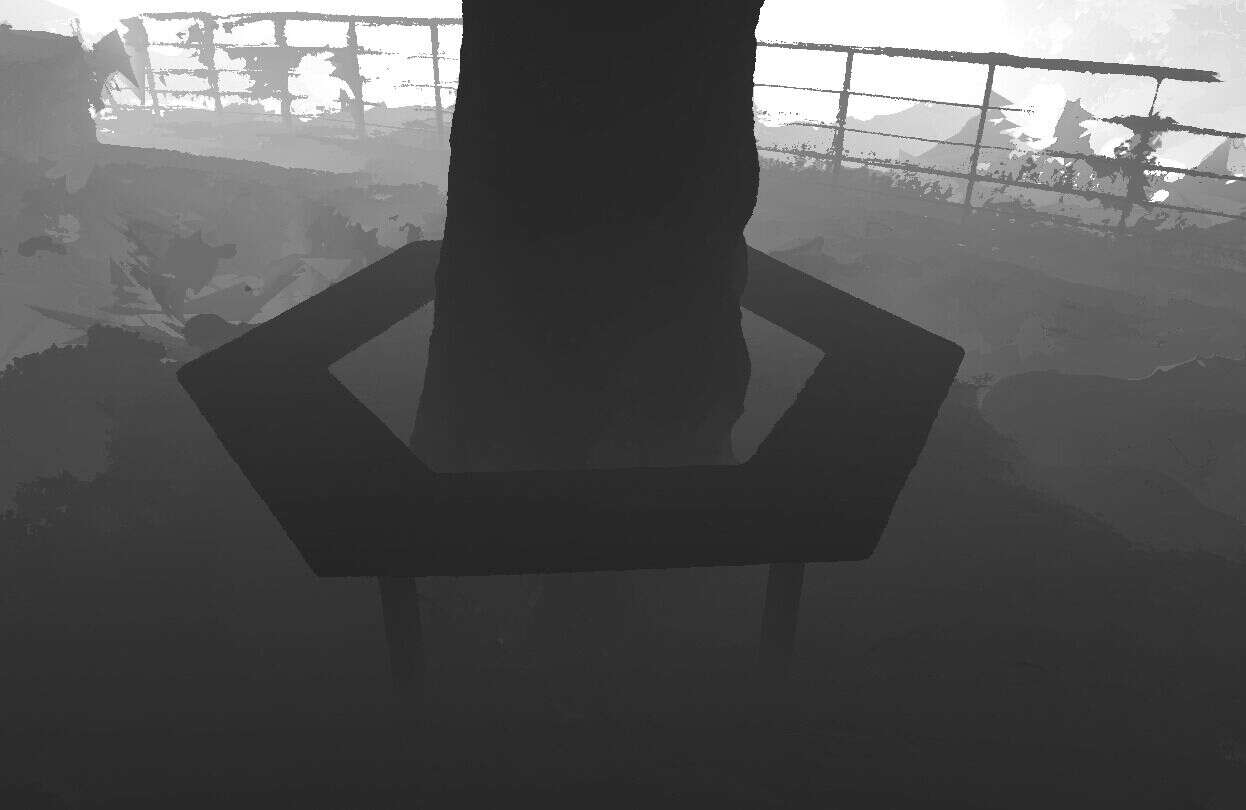} \\
        
        (a) GT Image & (b) Rendered Image & (c) Rendered Depth \\
    \end{tabular}
    }
    \caption{Visualizations of all outdoor scenes, \textit{(top-down) [bicycle, flowers, garden, stump, treehill]}, from the Mip-NeRF 360 dataset \citep{barron_2022_mip}. In the top-right section of the rendered image of treehill, \network{} imprecisely approximates the appearance of the background foliage, while GS-based algorithms typically blur this region.}
    \label{fig:mipnerf_360_outdoor}
\end{figure}

\end{document}